\PassOptionsToPackage{table}{xcolor}
\documentclass[onecolumn,letterpaper]{openhelix}

\usepackage{amsmath,amsfonts,bm}

\def\eqref#1{equation~\ref{#1}}
\def\1{\bm{1}}

\DeclareMathAlphabet{\mathsfit}{\encodingdefault}{\sfdefault}{m}{sl}
\SetMathAlphabet{\mathsfit}{bold}{\encodingdefault}{\sfdefault}{bx}{n}

\usepackage{array}
\usepackage{url}
\usepackage{wrapfig}
\usepackage{longtable}
\usepackage{adjustbox}
\usepackage{pifont}
\usepackage{makecell}
\usepackage{float}
\usepackage{listings}

\newcolumntype{M}{>{\centering\arraybackslash}m{0.85cm}}
\newcommand{\method}{EmbodiedSmith}


\hypersetup{
  pdftitle={EmbodiedSmith: Scaling Embodied Data through Recursive Self-Improvement Flywheel in Simulation},
  pdfauthor={Yikai Qin, Mingjian Liang, Yifei Deng, Wenxuan Song, ZePeng Lin, Zhiyi Jiang, Jiajun Fu, Qiao Sun, Huashuo Lei, Xicheng Gong, Jiayi Chen, Han Zhao, Shuanghao Bai, Pengxiang Ding, Pengwei Wang, Haoang Li},
  pdfkeywords={Generative AI, Robotic Data}
}

\title{EmbodiedSmith: Scaling Embodied Data through Recursive Self-Improvement Flywheel in Simulation}
\author[1, *]{Yikai Qin}
\author[1, *]{Yifei Deng}
\author[1, *]{Mingjian Liang}
\author[1, 2, \dagger, *]{Wenxuan Song}
\author[1, *]{Zepeng Lin}
\author[3, *]{Zhiyi Jiang}
\author[4, 7]{Jiajun Fu}
\author[2]{Qiao Sun}
\author[1]{Huashuo Lei}
\author[8]{Xicheng Gong}
\author[1]{Jiayi Chen}
\author[5, 6]{Han Zhao}
\author[7, 9]{Shuanghao Bai}
\author[5, 6, 7]{Pengxiang Ding}
\author[10]{Pengwei Wang}
\author[1, \ddagger]{Haoang Li}

\affiliation[1]{The Hong Kong University of Science and Technology (Guangzhou)}
\affiliation[2]{Xiaomi Robotics}
\affiliation[3]{Southeastern University}
\affiliation[4]{Tsinghua University}
\affiliation[5]{Zhejiang University}
\affiliation[6]{Westlake University}
\affiliation[7]{Symbiosis Robotics}
\affiliation[8]{Peking University}
\affiliation[9]{Xi'an Jiaotong University}
\affiliation[10]{Beijing Academy of Artificial Intelligence}

\contribution[*]{Equal Contribution}
\contribution[\dagger]{Project Lead}
\contribution[\ddagger]{Corresponding Author}
\metadata[Keywords]{Generative AI, Robotic RSI}

\abstract{
Scaling robotic foundation models requires diverse training data and reliable evaluation environments. Simulation offers a scalable solution, yet existing generation pipelines remain constrained by predefined assets and skills, a disconnect between scene generation and task generation, and limited support for complex embodiments and physics. We introduce EmbodiedSmith, a framework for scalable embodied data generation through recursive self-improvement (RSI). EmbodiedSmith unifies asset, scene, and task generation in a pipeline that supports autonomous creation and language-driven customization. Its core is an agentic refinement loop: scene generation anticipates downstream task requirements, while task generation guides targeted scene edits, allowing scenes and tasks to iteratively improve one another. This joint refinement improves task generation success, including for long-horizon tasks. The framework further supports mobile manipulators, humanoids, and dexterous hands, as well as interactions involving deformable objects and fluids, broadening the range of behaviors and physical phenomena represented in generated data. Together, these capabilities provide a flexible simulation engine for both robot pretraining and evaluation. Extensive experiments validate the quality, diversity, and generation efficiency of the resulting data, while downstream policy experiments demonstrate that increased data diversity improves generalization.
}

\begin{document}
\maketitle

\begin{figure*}[t]
    \centering    
    \includegraphics[width=\linewidth]{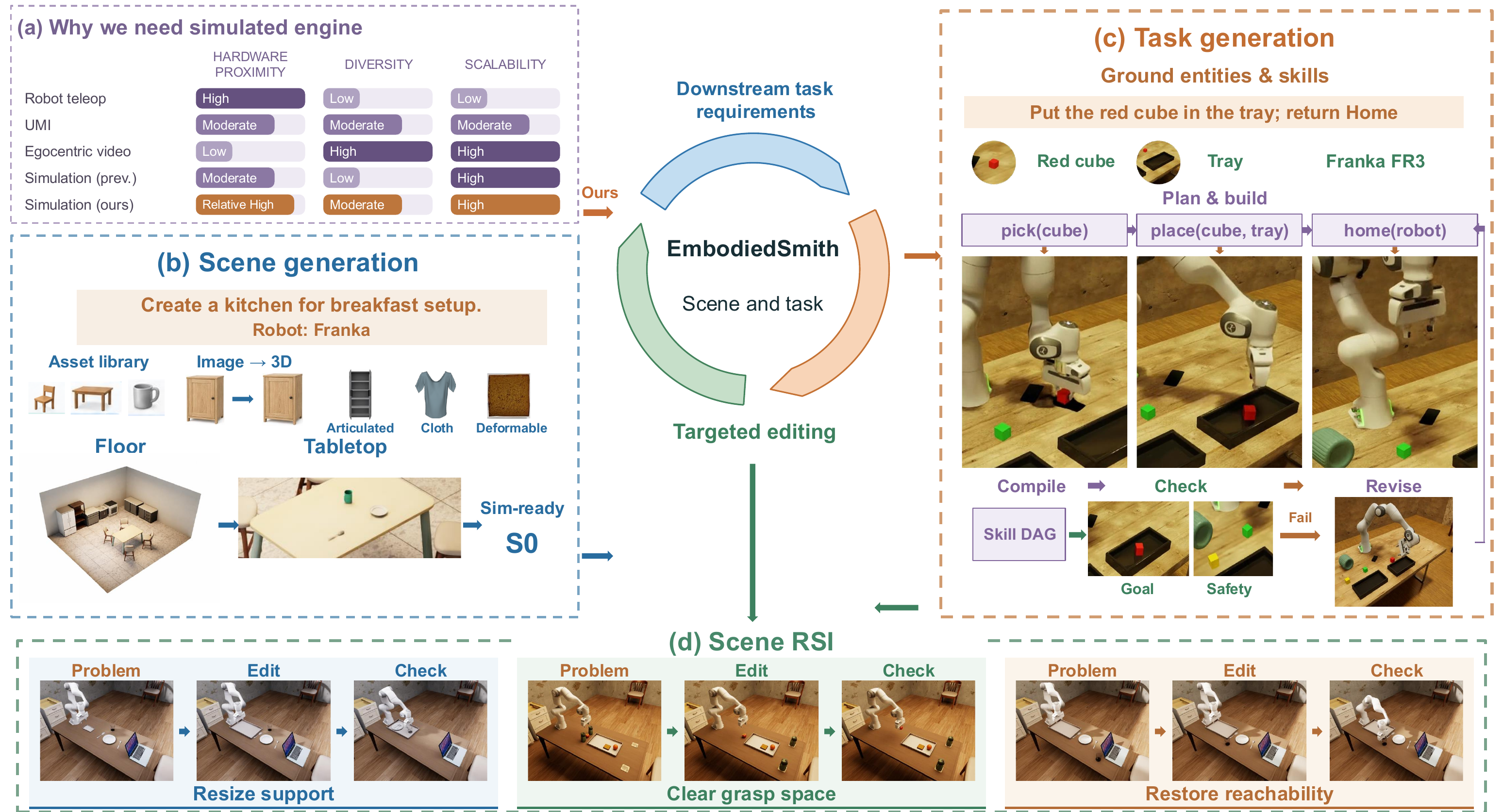}
    % \vspace{-6mm}
    \caption{\textbf{Overview of our \method. }Conditioned on a task and robot embodiment, EmbodiedSmith couples scene generation with task generation and execution, using feedback to iteratively refine scenes and replan.}
    \label{fig:pipeline}
    % \vspace{-5mm}
\end{figure*}

% \begin{wrapfigure}{b}{0.5\textwidth}
%     \centering
%     \includegraphics[width=\linewidth]{fig/teaser.png}
%     \caption{Comparison of different data sources.
%     We evaluate data along three dimensions:
%     \textit{hardware proximity}, alignment with the target
%     robot's embodiment and interfaces;
%     \textit{diversity}, coverage of tasks, scenes, objects,
%     and behaviors; and
%     \textit{scalability}, collection cost and speed.}
%     \label{fig:teaser}
% \end{wrapfigure}

\section{Introduction}
%wenxuan， mingjian
%yifei 已加入cite
%Large-scale pretrained models have been validated as an effective path towards robot foundation model.  A growing body of work has focused on collecting large-scale real-world robot datasets and training high-capacity robotic foundation models. real-world robot data has a high hardward proximity with low diversity and scalability；egocentric data和umi data则反之。得益于scalability强，high diversity and hardward proximity，仿真数据作为一种重要的数据源加入到机器人预训练中~\cite{interndata, robocasa 365, genesis}。
Large-scale pretrained models have been validated as an effective path towards robot generalists that can perform everyday tasks across diverse environments~\citep{kim2024openvla, bai2026embodied, song2026reconvla, song2025pd, yang20264d, chen2026uniwam}.  
A growing body of work has focused on collecting large-scale robot datasets and training high-capacity robotic foundation models.
These models have demonstrated meaningful generalization to novel objects, environments, and tasks, showing great promise towards developing broadly capable policies.

%仿真引擎在推动Large-scale pretrained models上发挥重要作用。
Simulation plays an important role in advancing large-scale pretrained models.
First, as shown in \Cref{fig:pipeline} (a), real-world robot data closely matches the target hardware but offers limited diversity and scalability, whereas egocentric and UMI data exhibit the opposite trade-off~\citep{ye2026data}. Simulation \textbf{data} combines scalability and diversity with close alignment to the target hardware, making it an important data source for robot pretraining~\citep{tian2026interndata,nasiriany2026robocasa365,Genesis}.
%成本低，收集速度
Second, simulation environments enable low-cost, fast, and reproducible large-scale \textbf{evaluation}~\citep{nasiriany2026robocasa365} and provide an ideal evolving environment for robotic agents. 
Thus, as recursive self-improvement (RSI) techniques advance~\citep{weng2026harness, xiao2026enpire}, simulation environments are becoming an essential component of robotics infrastructure~\citep{liu2026passive, Genesis}.

%然而，现有的仿真数据生成工作仍然存在明显局限： 1. limited diversity of scenes/asset/layout/skill，都是预设好或者规则生成的，无法自定义 2. task plan的成功率低，因为场景生成和机器人任务规划是解耦的，无法保证生成的机器人场景可以执行真实动作，这个问题随着任务变复杂/长程而扩大 3. 只能生成简单构型的机器人，距离人类的灵巧程度差距大，具体来说表现在对移动操纵支持差，对灵巧手和humanoid支持差, 这些构型的机器人的任务规划难度很大 4. 距离真实的物理世界远，因为对柔性和流体的支持差，而这些是物理世界的重要组成部分
However, existing simulation generation methods still face several limitations:
(1) limited diversity and customizability of scenes, assets, and skills, where the scenes and assets are already available in the dataset and skills are typically predefined or rule-based~\citep{nasiriany2026robocasa365};
(2) limited physical realism~\citep{wang2026gauge} due to inadequate support for deformable objects and fluids, both of which are essential components of the physical world; (3) low task execution success rates, as scene generation is decoupled from robot task generation, leaving the feasibility of planned actions unverified. This issue becomes more pronounced as tasks grow more complex or longer-horizon;
(4) limited support for complex embodiments, including mobile manipulators, dexterous hands, and humanoids. Task generation for these embodiments is challenging~\citep{li2026momagen}, while missing these embodiments limits robots' ability to acquire human-like dexterous behaviors.

%我们提出embodiedsmith，通过rsi的方式实现规模化的具身数据生成。具体来说，1. 我们级连了不同的生成能力，object, layout, scene, task都可以自主生成，并且支持language driven。2. 我们在场景生成的过程中加入对下一步任务生成的预先考虑，在任务生成时又赋予其反向进行场景编辑的能力，二者可以反复配合进行多轮调优，从而实现场景generation和task generation的Recursive self-improvement生成，通过agentic的配合实现场景和task生成中的互相考虑和裨益。这使得任务规划成功率大幅提升。这进一步使得我们的仿真引擎在long-horizon task中仍然保持较高的成功率。 3. 我们将robot task plan的支持构型扩充到移动操纵、humanoid和灵巧手，从而使得\method可以生成human-level dexterous data 4. 在资产和任务类型上，我们增加了对更复杂物理的样本（柔性、流体）的支持，填补了数据多样性的空白，并且使得仿真离真实物理世界更近一步。
We introduce \method, a framework for scalable embodied data generation through RSI. It is structured around four features:
\begin{itemize}
    \item \textbf{High Data Diversity:} We design strong 3D generation methods to generate any novel object, layout, and scene.
Besides, we construct effective agents that autonomously create and compose new skills to accomplish diverse task generation. These generation capabilities, spanning multiple levels of abstraction, are integrated into a unified pipeline that supports autonomous generation and language-driven customization. 

    \item \textbf{Rich Physics Simulation:} We support assets and tasks involving more complex physics, including deformable objects and fluids, addressing gaps in existing data diversity and narrowing the gap between simulation and the real physical world.
    \item \textbf{RSI Data Generation:} Scene generation anticipates downstream task requirements, while task generation can revise the scene through targeted editing. This agentic loop enables scene and task generation to recursively refine one another. This process substantially improves task generation success rates, allowing our simulation engine to maintain high success rates even for long-horizon tasks.
    \item \textbf{Human-level Dexterity}: We extend robot task generation to mobile manipulators, humanoids, and dexterous hands, enabling \method~to generate data involving human-like dexterity.
\end{itemize}
% (1) 
% (2)  (3)  (4) 
% We expand the diversity of assets and task types by supporting samples involving complex physical phenomena, such as deformable objects and fluids, thereby narrowing the gap between simulation and the real physical world.
%
%重要问题：数据集的贡献是否放在本文里 【目前认为不放，因为确实也没生成好，没法统计数据】，但pretrain data和benchmark之间的关系怎么说？

% By integrating these elements, our \method~为机器人基础模型提供了一个effective，flexible、comprehensive的仿真数据引擎。他可以作为预训练的有效数据源，并分析任务、环境和构型多样性对预训练模型的影响。他也可以作为有效的评估器为基础模型提供指标和反馈，从而赋能基础模型进化。我们进行了大量实验，验证了生成数据的质量、多样性、效率；进一步与policy结合的实验验证了更多样数据带来的模型泛化能力的提升。
By integrating these components, our \method~provides an effective, flexible, and comprehensive simulation data engine for robotic foundation models. It serves as an efficient source of pre-training data and enables systematic analysis of how task, environment, and embodiment diversity affect pre-trained models. It can also function as an effective evaluator, providing metrics and feedback that facilitate the iterative advancement of foundation models. Through extensive experiments, we validate the quality, diversity, and generation efficiency of the produced environments and demonstrations. Furthermore, experiments integrating the data with downstream policies demonstrate that greater data diversity improves model generalization.

%robocasa table1 @lmj

\section{Related Work}

\noindent\textbf{3D Scene Generation.}
% @mingjian
3D scene generation has been studied through learned generative models, procedural and foundation-model-guided pipelines, and agentic systems. Data-driven methods learn object distributions and spatial relations from 3D scene datasets: DiffuScene and Mixed Diffusion model indoor layouts using diffusion processes~\citep{tang2024diffuscene,hu2024mixed}, while PhyScene introduces collision, accessibility, and interaction constraints to improve physical plausibility~\citep{yang2024physcene}. Procedural methods instead assemble environments using predefined grammars and geometric rules, as demonstrated by ProcTHOR and Infinigen Indoors~\citep{deitke2022procthor,raistrick2024infinigen}. To enable open-vocabulary control, foundation-model-guided methods translate user instructions into structured scene layouts through direct coordinate prediction~\citep{feng2023layoutgpt}, language-guided asset retrieval~\citep{yang2024holodeck}, vision-language-based layout optimization~\citep{sun2025layoutvlm}, hierarchical scene composition~\citep{pun2025hsm}, or explicit constraint planning~\citep{sun2025roomplanner}. Agentic methods introduce iterative reasoning and refinement, including self-reflective tool use~\citep{yang2025sceneweaver} and coordinated semantic and physical correction~\citep{gao2025discolayout}. SAGE, SceneSmith, and EmbodiedGen V2 extend this paradigm toward embodied simulation by coupling multi-agent generation with task-aware composition, physical validation, interactive assets, and simulator-ready deployment~\citep{xia2026sage,pfaff2026scenesmith,wang2025embodiedgengenerative3dworld}.

\noindent\textbf{Sim-ready 3D Assets Generation.}
% @mingjian, zhiyi
3D asset generation is increasingly moving beyond visual fidelity toward interactive and physically grounded representations required by simulation. To make static assets interactive, articulated generation models part structure and kinematics from visual or geometric inputs, as explored in PAct~\citep{liu2026pact} and SIMART~\citep{zhang2026simart}. Agentic approaches further make this process iterative and verifiable, with articulation refined through closed-loop visual feedback in Articulate-Anything~\citep{le2024articulate} and constructed through program synthesis with executable validation in Articraft~\citep{zhou2026articraft}. Beyond kinematics, physical generation seeks to model how assets behave during interaction. Joint reasoning over geometry, articulation, and physical properties is introduced in PhysX-Anything~\citep{physxanything}, while functional and kinematic constraints guide asset synthesis in PhysForge~\citep{yang2026physforge}. Recent work moves toward broader physical grounding through intrinsic property modeling in UniPhysGen~\citep{li2026uniphysgen}, unified rigid, articulated, and deformable generation in PhysX-Omni~\citep{cao2026physxomni}, and simulation-diagnosed deformable refinement in DiagGen~\citep{chen2026diaggen}. At the system level, automated validation and simulator deployment are integrated with asset generation in EmbodiedGen V2~\citep{wang2025embodiedgengenerative3dworld}.

% @yikai, yifei, zepeng, sunqiao, jiajun
% zepeng(0902)
%yifei
%一类方法通过迁移与重组少量人类示范，扩大机器人训练数据的规模。MimicGen 将源示范中的操作片段适配到新的物体配置。DexMimicGen 将其扩展到双臂灵巧操作，SkillMimicGen 则结合运动规划连接不同操作技能。RoboCasa 将示范扩增应用于多样化的厨房环境和日常任务。这些方法降低了重复示教的成本，但仍依赖源示范及其对应的操作结构，为新行为生成数据时通常需要相应的示范或任务适配。
% 另一类方法利用仿真状态、抓取预测和运动规划直接构造机器人操作轨迹，减少对人类示范的依赖。SAGE 将任务条件下的场景生成与动作生成相结合，在生成环境中合成抓放和移动操作示范；SceneSmith 则生成可物理交互的室内场景，并将其用于自动机器人策略评估。这些工作推动了生成场景向机器人学习环境的转化，也使研究进一步关注：如何在变化的场景中，将任务要求落实为符合机器人能力与物理约束的执行方案。
% EmbodiedSmith 面向新实例化的场景与任务组合，将环境适配、任务规划和示范采集连接起来。首先，将输入场景中的资产、物体布局、支撑关系和物理属性规范化为仿真配置，并建立可供任务规划使用的对象与能力描述。随后，根据自然语言任务确定操作对象、目标关系和子任务顺序，从已有技能库中编排操作技能；结合机器人能力选择执行机械臂与工作位姿，并通过几何、碰撞和运动规划检查，将语义计划编译为可执行配置。最后，在物理仿真中执行技能序列，记录视觉观测、机器人状态与动作，并依据任务终态和数据完整性检查筛选有效示范。对于失败实例，系统根据结构化执行证据，在限定次数内调整工作位姿、场景布局或技能参数并重新验证。由此，系统在已有资产和技能支持范围内，无需为每个新任务实例采集源示范，即可将新的场景—任务组合转化为机器人训练数据。
% [1] Human-demo-based synthesis
%     MimicGen
%     → DexMimicGen / SkillMimicGen
%     → RoboCasa
%     → limitation:
%       requires source demonstrations / predefined task decomposition

% [2] Simulator-/planner-based expert generation
%     motion planning / privileged simulator state
%     → SAGE
%     → RoboCasa
%     → SceneSmith robot interaction
%     → limitation:
%       usually tied to predefined tasks/scenes/planners
% [3] Ours
%     scene + task + demonstration generation are coupled
%     → generate executable data for newly instantiated task environments
%     → jointly scale environment/task/data diversity
%jiajun
\noindent\textbf{Simulated Robot Demonstration Generation.}
Humanoid simulation benchmarks increasingly cover locomotion and object interaction within the same embodiment. HumanoidBench \citep{sferrazza2024humanoidbench} provides a standardized suite of whole-body locomotion and dexterous manipulation tasks, while SIMPLE \citep{wei2026simple} combines contact-rich simulation and photorealistic rendering with motion-planning and teleoperation pipelines for demonstration collection. More directly, HumanoidGen \citep{jing2025humanoidgen} generates language-conditioned bimanual tasks and demonstrations through spatial constraints, and HumanoidMimicGen \citep{lin2026humanoidmimicgen} adapts segmented source demonstrations to new configurations using whole-body planning. These works establish important benchmark and synthesis capabilities, but their reported settings center on curated task suites, annotated manipulation primitives, or source-demonstration-dependent skill sequences. Mapping newly instantiated scene–task combinations to executable, feasibility-checked whole-body demonstrations through a reusable skill interface remains less systematically examined.

\section{Method}

% Given a task specification and a target robot embodiment, \method~ generates validated robot demonstrations through a task-conditioned simulation pipeline with RSI, as illustrated in \Cref{fig:pipeline}. \Cref{sec:3.1} constructs a simulation-ready scene from validated
% assets and provides scene edit for targeted revision. \Cref{sec:3.2} grounds the task into executable skill plans and closes the RSI loop by coupling planning and execution with feedback-driven scene and plan refinement. \Cref{sec:3.3} extends the shared planning, execution, and validation interface to humanoid and dexterous embodiments through embodiment-specific execution adapters.
Given a task specification and robot embodiment, \method~generates validated
robot demonstrations through the RSI pipeline in \Cref{fig:pipeline}.
\Cref{sec:3.1} constructs task-conditioned scenes through asset generation, scene generation, and scene edit.
\Cref{sec:3.2} grounds tasks
into executable skill plans and refines them through generation and execution
feedback.
\Cref{sec:3.3} extends execution to humanoid and dexterous
embodiments through embodiment-specific adapters.

\begin{figure*}[t]
    \centering    
    \includegraphics[width=1.0\linewidth]{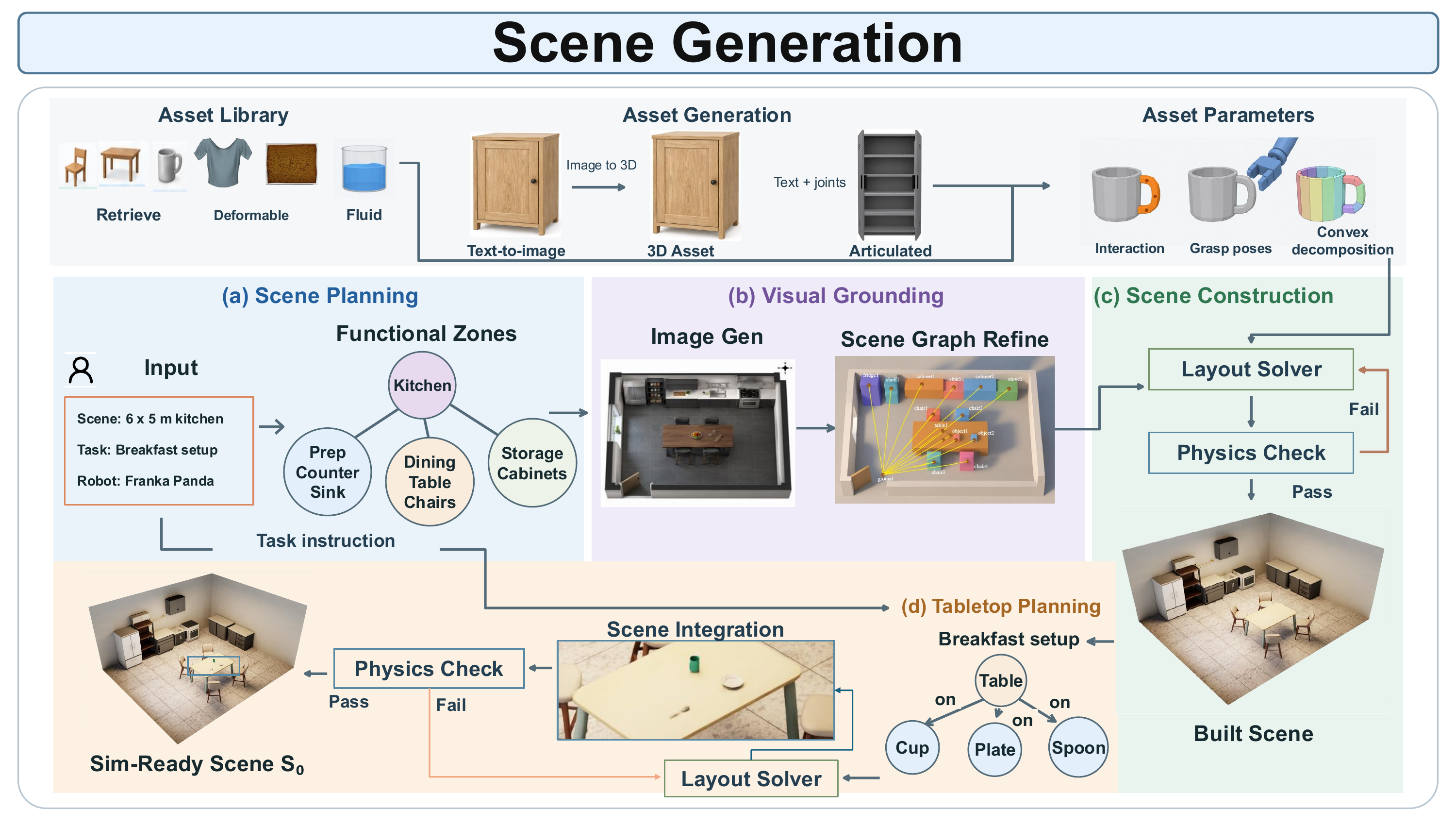}
    \caption{\textbf{Scene overview. }Our system transforms scene specifications, task instructions, and robot configurations into simulation-ready 3D scenes through functional zone planning, visual grounding, asset retrieval and generation, and physics-validated scene and tabletop layout optimization.}
    \label{fig:scene_generation}
\end{figure*}

\subsection{Environment Construction}
\label{sec:3.1}
% 1. 场景和资产的生成和标注@mingjian
% \method 通过四个相互关联的模块构建仿真环境：通用资产生成、物理资产生成、任务条件场景生成和场景编辑。给定任务规范，场景生成模块确定所需实体及其空间关系，资产流水线则提供经过验证的刚性、关节、柔性和流体资产。这些资产被组合为候选场景，并在仿真中接受几何可行性和物理稳定性验证。场景编辑模块通过局部修改生成不同的任务配置，最终环境则包含时间同步的观测数据、标注信息和仿真状态。
% \method~constructs task-conditioned environments through four complementary modules: common asset generation, physical asset generation, scene generation (\Cref{fig:scene_generation}), and scene editing. Together, these modules construct assets with validated geometry and physics, assemble scenes under task and embodiment constraints, and adapt existing environments to new tasks through targeted edits.

% \method~constructs a task-conditioned simulation scene through Asset Generation, Scene Generation and Scene Edit. The asset modules provide
% validated simulation-ready assets, which Scene Generation assembles under task
% and embodiment constraints (\Cref{fig:scene_generation}). 

\method~constructs task-conditioned simulation scenes through Asset Generation,
Scene Generation, and feedback-driven Scene Edit. 
\noindent\textbf{Asset Generation.}
% Existing asset libraries vary in semantics, coordinate conventions, scale, and geometry, while generated assets may have unreliable collision or physical properties. We therefore maintain an asset pool combining validated open-source and generated assets. Existing assets are normalized by a VLM from multi-view renderings and metadata and checked for semantics, geometry, scale, and scene compatibility. If retrieval fails, a scene-consistent reference image is refined and reconstructed by SAM3D~\citep{sam3d}, then scaled, repaired, and equipped with collision geometry through adaptive convex decomposition. A VLM assigns physical parameters, while resting, contact, and perturbation tests validate the asset. Failed assets are repaired or regenerated, whereas validated assets are retained for reuse across scenes.
Existing asset libraries provide broad object coverage but often differ in semantic labels, coordinate conventions, scale, and geometric quality, while on-demand 3D generation extends open-category coverage at the cost of less reliable geometry and physical properties. We therefore maintain a continuously expandable asset pool that combines validated open-source assets with scene-conditioned generated assets. Existing assets are canonicalized by a VLM using multi-view renderings and metadata to align semantic categories, orientations, coordinate systems, and real-world scales, and are subsequently checked for semantic correctness, geometric integrity, and scene compatibility. When retrieval cannot provide a suitable asset, we first refine a scene-consistent reference image in image space to control its dimensions, appearance, and coarse structure, and then reconstruct it into a textured 3D asset using SAM3D~\citep{sam3d}. The resulting asset is scaled, geometrically repaired when necessary, and equipped with collision geometry through adaptive convex decomposition. A VLM further assigns category-consistent physical parameters, while simulation-based resting, contact, and perturbation tests verify geometric and physical validity. Both retrieved and generated assets pass through the same validation pipeline: failed candidates are repaired or regenerated, whereas validated assets are stored in the persistent pool for reuse across scenes.

\noindent\textbf{Physical Asset Generation.}
Physical asset generation extends the common asset pipeline through three stages of construction, physics validation, and task validation, aiming to produce assets that are simulator-compatible, exhibit the intended physical behavior, and support downstream robotic interactions. In the construction stage, articulated assets follow an Articraft-based~\citep{zhou2026articraft} loop in which an LLM iteratively writes and revises construction programs against a domain-specific SDK using compiler and validation feedback. Generation further incorporates downstream robotic constraints, including metric scale, simulator coordinate-frame conventions, gripper compatibility, and modality-specific motion ranges. Deformable generation supports part-aware volumetric representations with regional material properties, while the evaluated implementation adopts a simplified procedural representation with uniform material properties, subsequently calibrated for interaction stability. Fluid assets similarly couple geometry with solver-specific physical parameters to support interactions such as pouring.

In the physics validation stage, constructed assets are evaluated for their intended behavior in simulation. Articulated assets are checked for inertial consistency, functional joint motion, grasp keypoints, and enclosure geometry, while deformable and fluid assets undergo corresponding deformation- and flow-based probes. For the evaluated deformable configuration, incomplete post-release recovery is documented as a known limitation rather than an acceptance criterion. Failed gated checks trigger targeted revisions, with non-improving changes reverted. In the task validation stage, CuRobo~\citep{sundaralingam2023curobo} first assesses collision-free motion feasibility before simulator execution. If manipulation fails, an agent role distinct from execution analyzes telemetry to identify whether task configurations, typed execution skills, or derived simulator assets require revision. Repairs are then applied and revalidated against the failed criterion, with unsuccessful changes reverted. Revisions to derived assets follow reproducible procedures that preserve the original source assets. Only episodes passing scene inspection, motion planning, and data export are retained as keepers, with validation records tied to the corresponding robot configuration. Integration into the scene generator remains downstream work.

\noindent\textbf{Scene Generation.}
Given a natural language description $\mathcal{T}$, task context, and robot embodiment configuration $\mathcal{E}$, our scene generator synthesizes an initial scene hypothesis $\mathcal{S}_0 = (\mathcal{G}, \mathcal{O}, \mathcal{X}, \mathcal{Q})$, where $\mathcal{G}$ denotes room boundaries and architectural openings, $\mathcal{O}$ represents object instances, $\mathcal{X}$ specifies 6-DoF poses, and $\mathcal{Q}$ designates support regions and robot interaction constraints. As illustrated in \Cref{fig:scene_generation}, the process decouples open-ended semantic structuring from deterministic geometric solving. A VLM first parses the input into a typed relational graph encoding functional zones, static fixtures, orientations, support surfaces, and robot approach directions. Dedicated asset modules then instantiate its nodes with retrieved or generated assets.

Scene construction subsequently solves the layout hierarchically: ground fixtures are organized into local compositions and joint-optimized via Beam-DFS search, followed by wall fixture placement and parent-relative task-object positioning. The optimization enforces boundary limits, collision avoidance, clearances, connectivity, and workspace bounds, with reachability validated by CuRobo. The resulting $\mathcal{S}_0$ is therefore an initial geometry- and reachability-validated scene rather than a guaranteed executable environment. Downstream robot planning and execution further evaluate task feasibility and return feedback to the RSI loop (\Cref{fig:pipeline}). Environment-induced failures then trigger targeted Scene Edit, followed by revalidation, replanning, and re-execution.

\noindent\textbf{Scene Edit.}
\label{sec:scenedit}
Scene Edit adapts validated environments to new tasks within
the RSI loop without repeated scene synthesis. Given the scene, task, robot embodiment, and generation or execution feedback, the scene agent applies schema-constrained edits while preserving unrelated scene elements. Supported operations include task-object addition, replacement, rescaling, and reachability-aware repositioning within or across support surfaces. Asset replacement updates geometry, physical properties, grasp annotations, and simulator bindings according to the placement policy. Each edit can be previewed before application and produces a traceable revision identifying the physical, reachability, and export checks requiring revalidation before replanning.
%@zepeng
%\noindent\textbf{场景编辑。}
% 许多具身任务共享相同的功能环境，仅在任务相关物体与局部空间需求上有所差异。我们通过针对性的场景编辑复用已验证的环境，减少重复的场景合成、资产实例化与验证成本。给定任务指令，场景智能体结合当前场景与机器人本体确定必要修改，同时保留与任务无关的几何结构、固定设施和资产。

% 编辑通过受 schema 约束的接口执行，支持任务物体的新增与替换、支撑面内或跨支撑面的移动、尺寸调整，以及面向机器人可达性的放置优化。资产替换遵循指定的放置策略，并同步更新相关几何、物理属性、抓取标注与仿真绑定。每项编辑均可在执行前预览；执行后生成可追踪的场景版本，并标记需要重新验证的物理、可达性与导出检查项。

% Many embodied tasks share a functional environment but differ in task-relevant objects and local spatial requirements. We reuse validated environments through targeted scene edits to reduce repeated synthesis, asset instantiation, and validation costs. Given a task instruction, the scene agent identifies necessary changes based on the current scene and robot embodiment while preserving unrelated geometry, fixtures, and assets.

% Edits are applied through a schema-constrained interface supporting task-object addition and replacement, repositioning within or across support surfaces, rescaling, and placement adjustment for robot reachability. Asset replacement follows a specified placement policy and updates the associated geometry, physical properties, grasp annotations, and simulator bindings. Each edit can be previewed before execution and, once applied, produces a traceable scene revision identifying the physical, reachability, and export checks that require revalidation.

\subsection{Robot Trajectory Generation}
\label{sec:trajec_gen}

With the task-conditioned scene $S_0$ from \Cref{sec:3.1}, the task generation grounds the task to scene entities and atomic skills, forming an executable Skill DAG that is validated through compilation, checking, and simulation. The robot task generation system consists of three core components.
The \textbf{Task Plan Agent} interprets the task, makes the main generation decisions, and proposes repairs.
The \textbf{Harness} manages the workflow, initiates independent reviews, and serves as the driver and central scheduler of the overall framework.
The \textbf{\texttt{simengine}} carries out simulation-based validation and records the execution.

% Using the task-conditioned scene $S_0$ from \Cref{sec:3.1}, the task planning grounds the task to scene entities and atomic skills, forming an executable Skill DAG that is validated through compilation, checking, and simulation, as shown in Figs.~1(c--d) and~3. The system comprises three components. The \textbf{Task Plan Agent} makes planning and repair decisions, the \textbf{Harness} coordinates validation and workflow, and the \textbf{\texttt{simengine}} compiles, executes, and records each attempt. Both the Agent and the Harness invoke VLMs for semantic and visual reasoning.

\begin{figure}[p]
    \centering
    \includegraphics[width=\linewidth]{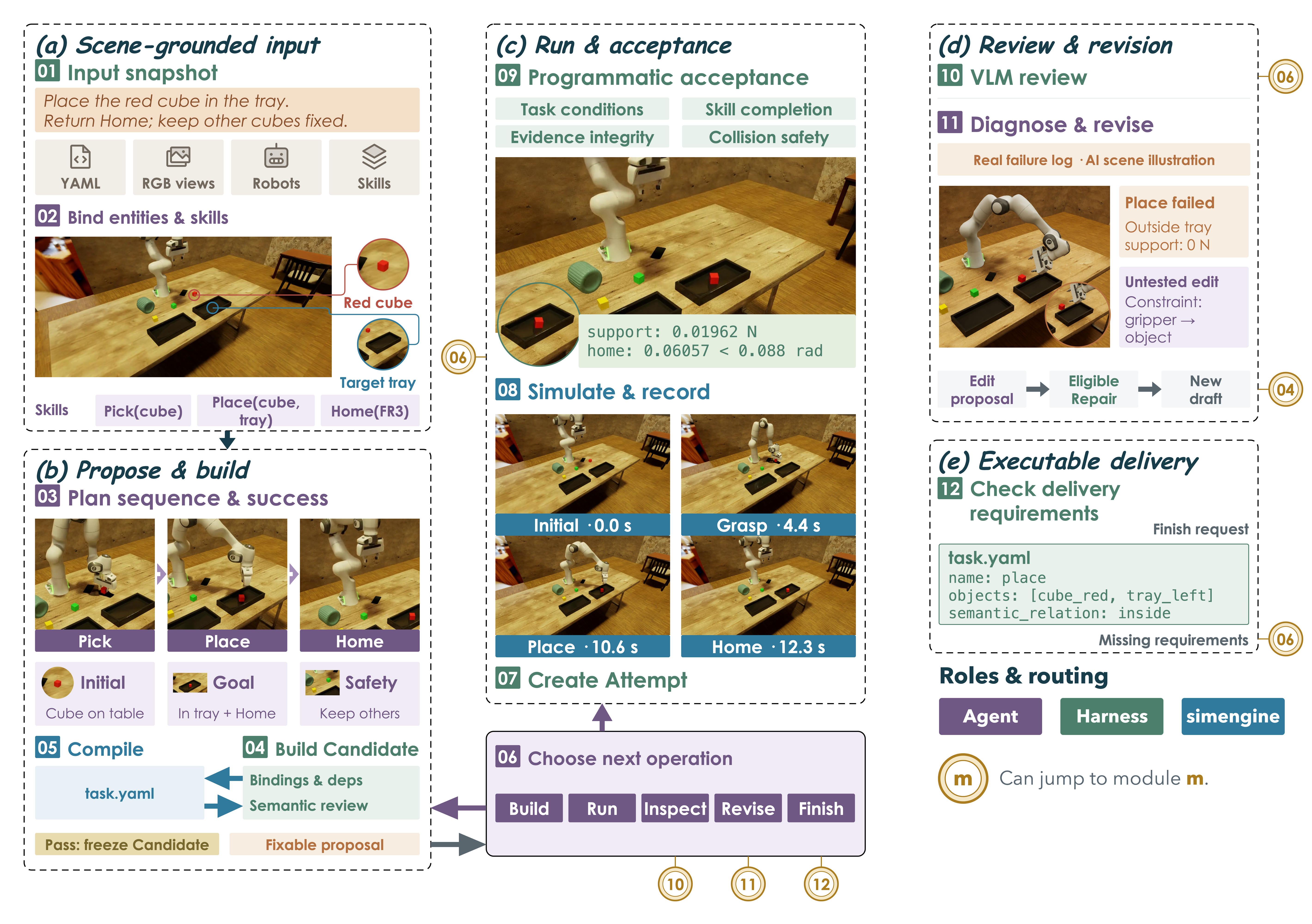}
    \caption{
\textbf{Closed-loop robot task generation and validation.}
(a) Instructions are grounded in the scene, robot configuration,
and available skills.
(b) The Agent proposes a skill sequence and success conditions,
which are used to build and compile a candidate task configuration.
(c) The candidate is executed in simulation and evaluated through
programmatic acceptance checks.
(d) Independent VLM review complements these checks, while failures
trigger diagnosis, revision, and revalidation.
(e) The executable configuration is delivered upon final acceptance.
% The Agent plans and revises, the Harness coordinates and validates,
% and \texttt{simengine} compiles, simulates, and records.
}
    \label{fig:ta}
\par\vspace{8pt}
\begingroup
\captionsetup{type=table}
\centering
\caption{\textbf{Agent Tools.} We summarize the five task-level tools exposed by the harness for planning, execution, feedback-driven repair, and final validation.}
\label{tab:agent_tools}
\footnotesize
\setlength{\tabcolsep}{3pt}
\renewcommand{\arraystretch}{1.15}
\begin{tabular}{@{}p{0.20\linewidth}p{\dimexpr0.80\linewidth-2\tabcolsep\relax}@{}}
\toprule
\textbf{Name} & \textbf{Description} \tabularnewline
\midrule
\texttt{inspect} & Provide object identities, poses, support relations, interaction regions, and visual observations so the agent can ground targets and select an initial robot placement. After execution, invoke an independent VLM review of the candidate and its rollout evidence, and return the reviewer's feedback before finalization. \tabularnewline
\texttt{build} & Organize grounded interactions into a skill DAG with target entities, execution resources, goal parameters, and dependencies. Define success criteria for terminal states, required events, safety constraints, and prohibited events. Before execution, the harness checks object references, skill parameters, robot compatibility, dependency structure, and consistency with the task instruction. \tabularnewline
\texttt{run} & Execute the skill DAG in physics simulation with embodiment-specific motions. The harness returns skill outcomes, object and attachment states, safety events, and task checks. \tabularnewline
\texttt{revise} & Use failed-rollout evidence to revise affected object bindings, dependencies, robot working poses, grasp poses, approach directions, or motion parameters. The revised plan is revalidated while task-success criteria remain fixed. \tabularnewline
\texttt{finish} & Submit a successful rollout for final checks of terminal state, required events, runtime safety, and data integrity. The harness retains accepted trajectories as demonstrations. \tabularnewline
\bottomrule
\end{tabular}
\endgroup

\end{figure}

% 进行任务规划时，我们可利用的资源和已知的信息包括描述任务生成的prompt，场景生成agent生成的场景和相关配置文件，机器人类型和相关资产与配置（从prompt提取得到或者配置文件指定），可用的原子动作技能列表和一个对场景的全局视角的观察图片。这些将作为原始依据提供给harness检查一致性并初步冻结，以保证后续规划与验证的依据始终一致，并在完成验证后交付给agent完成对应和动作skill的映射和绑定。agent会根据绑定结果，决定机器人站位、控制器、动作参数和依赖顺序，形成紧凑的原子动作Skill DAG，同时把任务要求转成可检查的成功条件，并建立“任务要求—动作节点—验证证据”的对应关系。例如，“放到托盘上”既需要 Place 动作，也需要最终物体与托盘的关系检查，并输出一份完整的任务提案。
\textbf{Task Grounding and Planning.}
We first input the task prompt, the scene configuration files produced by the scene generation agent, the embodiment configuration files, a list of available atomic Skills, and an image of the scene from a global viewpoint to the Harness.
As shown in \Cref{fig:ta}(a), the Harness first checks these inputs for consistency and freezes them as an input snapshot. Then it passes the input to the Agent, which maps task references to scene entities and binds the required atomic Skills.
Based on these bindings, the Agent determines the robot placement, action parameters, and skill execution order, forming a directed acyclic graph of atomic skills \textbf{(Skill DAG}).
After that, it translates task requirements into checkable success conditions and establishes an explicit correspondence among task requirements, Skill nodes, and validation evidence.
For example, ``place the cube in the tray'' requires checking both the \texttt{Place} action and the final spatial relation between the cube and the tray.
Together, these elements form a complete task proposal, as illustrated in step~03 of \Cref{fig:ta}(b).

% 如\Cref{fig:ta}中06所示，在我们的系统中，task plan agent可调用的tool（可调用的高层交互Skill）一共有5个，分别为build、run、inspect、finish、revise。所有agent的操作都依赖于这5个高层工具的调用，这五个工具按任务生命周期划分接口，每个接口通过结构化参数表达具体请求，由底层组件完成相应处理；Agent再根据返回结果组合、重复调用这些工具，从而完成不同任务。如\Cref{fig:ta}中04所示，agent在得到提案后调用build，提交提案给harness进行实体引用、参数、机器人绑定和任务要求覆盖等一系列检查，并形成实际任务配置文件task.yaml。这份配置文件会提交给仿真引擎做真实解析任务和验证，随后编译器返回节点信息（如果成功编译）或者失败日志（如果失败）给harness继续对配置文件进行几何检查和语义审查等，直到通过后才冻结配置为Candidate并反馈回agent，第一次的build也就此结束。
\textbf{Task Construction.}
As shown in step~06 of \Cref{fig:ta}, the Task Plan Agent has access to five high-level interaction tools: \textbf{\texttt{build}}, \textbf{\texttt{run}}, \textbf{\texttt{inspect}}, \textbf{\texttt{finish}}, and \textbf{\texttt{revise}}. \Cref{tab:agent_tools} summarizes their interfaces and validation responsibilities. For example, after forming a proposal, the Agent calls \texttt{build} to submit it to the Harness, as illustrated in step~04 of \Cref{fig:ta}(b).
The Harness checks entity references, parameters, robot bindings, and task-requirement coverage, among other aspects, and produces the task configuration file. This configuration is then passed to \texttt{simengine} for parsing and compilation. The compiler returns node information if compilation succeeds, or failure logs otherwise.
The Harness uses this feedback in its subsequent validation, which also includes geometric checks and semantic review. Only after all checks pass is the configuration frozen as a Candidate and returned to the Agent, completing the first successful \texttt{build}.

% 当第一次build结束，Agent会进行一个持续决策过程，不断地读取工具返回的检查结果、Candidate信息或执行反馈，选择下一项操作。\Cref{fig:ta}中06的5个系统tool就是决策可用的操作。
% After the first \texttt{build}, the Agent continues in a decision loop, repeatedly reading the check results, Candidate information, or execution feedback and choosing its next operation.

% 例如当agent进行判断后认为当前配置可以使用，会调用run，请求harness指定Candidate、验证阶段和seed并准备运行环境等来启动真实实验。实验时会同步记录物体与机器人状态、Skill事件、安全信息、观测、动作和视频。任务结束或失败后，将这些证据交给Harness检查初始条件、必需动作与事件、最终状态和安全条件，同时核验证据是否完整、是否属于本次Candidate和Attempt，以及测量与结果是否一致等。例如在\Cref{fig:ta}中09：Place已结束，还需要检查方块是否满足目标关系；Home已调用，还需要检查机械臂是否到达规定状态。一系列校验结果（以及一些如果有的话的原始失败证据如失败类别、日志与图像）将会再次返回给agent进行判定接下来调用哪种tool（通常在一次run结束之后，不会再次调用run）。

\label{sec:3.2}

\textbf{Execution and Validation.}
% After the configuration is ready, the agent calls \texttt{run}.
% The Harness then prepares the environment and launches a simulation for a specified Candidate, validation phase, and random seed, creating a new \textbf{Attempt} as shown in step~07 of \Cref{fig:ta}(c).
% During execution, the system records object and robot states, Skill events, safety information, observations, actions, and video.
% Once the task finishes, these evidences are passed to the Harness for \textbf{programmatic acceptance}.Then the check results, together with any available raw failure evidences are passed to the Agent to inform its next tool call.
Once configured, the agent calls \texttt{run}. The Harness prepares the environment and launches a simulation for a specified Candidate, validation phase, and random seed, creating a new Attempt as shown in step~07 of \Cref{fig:ta}(c). During execution, the system records object and robot states, Skill events, safety information, observations, actions, and video. Once the task finishes, this evidence is passed to the Harness for programmatic acceptance. Then, the check results and available raw failure evidence are passed to the Agent to inform its next tool call.

% 如果幸运地第一次就判定通过，那么agent也不能直接请求finish，通常会至少一次地调用inspect工具来启用独立vlm reviewer再次审核并由agent接收反馈；直到agent根据补充审核完全地断定通过后才会调用finish来获得最终可用的任务配置。这样的复核审议制度能大大提高判断的准确性和最终产物的实际可用性。
\textbf{Independent Review and Finalization.}
If the first run passes these checks, the Agent does not immediately request \texttt{finish}. Instead, it invokes \texttt{inspect}, asking the Harness to initiate an independent VLM review and return the reviewer's feedback, as illustrated in step~10 of \Cref{fig:ta}(d).
Only after considering this additional review and confirming
task success does the Agent invoke \texttt{finish}, triggering
the Harness's final delivery checks. If these checks pass, the Harness returns the final executable task configuration, as shown in \Cref{fig:ta}(e).

% 如果判定不通过，agent会调用revise，根据日志、测量和图像分析原因，提出具体修改，例如调整合法站位、动作参数、依赖关系，或提出有证据支持的抓取位姿修正。然后agent提出新的提案由Harness检查修复是否允许。允许后再从原Candidate或失败草稿复制出新草稿执行受控修改。新草稿再次由agent调用build重新编译、检查和冻结，随后重新验证。已有Candidate及其执行记录保留，新版本不能直接继承旧版本的通过结论。
\textbf{Repair and Revalidation.}
If the task does not pass, the Agent invokes \texttt{revise}
to propose changes, such as adjusting robot placement, tuning action parameters, or correcting grasp poses.
As shown in step~11 of \Cref{fig:ta}(d), the Harness first checks whether the proposed repair is allowed.
Upon approval, it copies the original Candidate or failed draft into a new draft and applies the approved changes. The Agent then invokes \texttt{build} to recompile and check the revised draft. If these checks pass, the Harness
freezes it as a new Candidate, which must undergo validation again.

% Existing Candidates and their execution records are preserved, and \textbf{a new version cannot inherit a previous version's passing status}.
% The Task Plan Agent and the Scene Agent can interact during the repair stage. If the Task Plan Agent identifies fundamental scene-level issues that cannot be resolved through task-plan repair alone, such as poor reachability of task-critical objects from the robot's initial position or existing tabletop assets obstructing the robot's motion paths, it can also invoke the scene-editing capabilities described in \Cref{sec:trajectory_generation} to refine the scene. These capabilities include adding or replacing task objects, moving objects within or across support surfaces, resizing objects, and optimizing object placement for robot reachability.

\subsection{Embodiments for Human-level Dexterity}
% 5. @jiajun，桥哥
\label{sec:3.3}

\noindent\textbf{Humanoid Robots.}
% 为了在 Unitree G1 人形机器人上执行动作，我们在共享技能图下方引入了专属执行适配器，在无需修改通用技能的前提下实现控制解耦：位移目标被转换为平面速度与航向角以驱动步态平衡策略，手臂动作则由 G1 运动学模型和 cuRobo 求解。在 50Hz 周期下，适配器将 15 自由度下肢与 14 自由度手臂指令融合成 29 自由度的全身动作下发，手部单独控制。这种将硬件映射完全隔离在适配层的架构，确保了上层任务技能跨本体的完全复用。
To execute the generated task plan on the Unitree G1 humanoid \citep{unitree2024g1}, an execution adapter sits beneath the shared skill DAG and decouples control without altering atomic skills: locomotion targets are translated into planar velocity and heading references for learned balance-walking policies \citep{nvidia2025grootwbc}, while arm goals are resolved via the G1 kinematics and cuRobo \citep{sundaralingam2023curobo}. At 50~Hz, the adapter merges 15-DoF lower-body and 14-DoF arm commands into a 29-DoF body action, dispatching hands separately. Confining embodiment-specific mappings entirely to the adapter keeps upper-level task skills fully reusable across morphologies \citep{he2024hover,li2025amo}.

\noindent\textbf{Dexterous-hand Robots.}
% 针对灵巧操作，我们在 Split ALOHA 上搭载 Sharpa Wave 灵巧手（单侧 6 轴机械臂加 22 自由度手）。专用的接触求解器将交互需求转换为无碰撞的手部姿势与指关节构型。运行中由 cuRobo 规划机械臂路径，直接控制手部各关节，并利用 PhysX 采集逐连杆接触遥测数据而非单一夹爪数据。任务执行采用严格的故障即闭（fail-closed）契约判定。该机制接入统一的五工具智能体循环（\texttt{inspect}、\texttt{build}、\texttt{run}、\texttt{repair}、\texttt{finish}），无需任何人类示范即可生成高精度的接触丰富型轨迹。
For dexterous manipulation, we deploy Sharpa Wave hands on the Split ALOHA platform (6-DoF arm and 22-DoF hand per side). A task-specific contact solver maps object interaction contracts into collision-checked hand poses and finger configurations. During execution, cuRobo plans arm trajectories while finger joints are commanded explicitly, with PhysX recording per-link contact telemetry rather than aggregated gripper readings. Each task enforces a fail-closed contact contract (e.g., designated index-pad contact and phase-wise force gates). Integrated with the five-tool agent loop above, this pipeline enables automatic contact-rich trajectory generation.

\section{Experiments}
We design our experiments to answer three questions:
%我们的任务导向的环境生成管线是否更好的满足perceptual和geometric上的要求？
%我们的环境生成+机器人轨迹生成的rsi框架是否使得模型更可靠的生成更困难、更多样的任务？
%由我们的方法生成的多样化环境是否能带来更强的模型性能？
\textbf{(Q1)} Does our task-oriented environment generation pipeline better satisfy perceptual and geometric requirements? (\Cref{sec:4.1})
\textbf{(Q2)} Does our RSI framework enable more reliable generation of challenging and diverse robot tasks? (\Cref{sec:4.2})
% \textbf{(Q3)} Do the diverse demonstrations generated by our \method~improve model performance? (\Cref{sec:4.3})
\textbf{(Q3)} Do our generated diverse demonstrations improve model performance? (\Cref{sec:4.3})

% 1. 模型训练无关的：场景质量、多样性、生成速度、标注的质量
\subsection{Environment Generation} %(EGV2 3.1, scenesmith 4.2) %user study; mllm rating; qualitative comparison; ablation; speed and cost; qualitaty of annotation
% mingjian：
% 对比基线：SceneSmith, Embodied gen v2, Holodesk, Layout VLM, SAGE
% 场景质量实验首先通过 Content & Function 相关指标验证生成场景是否具备完整、合理的物体组成与功能布局，证明场景不仅“看起来像”，而是真正满足日常功能需求。其次，通过 Spatial & Physics 指标评估支撑关系、可达性、可导航性、越界、碰撞和稳定性，证明生成结果能够直接进入物理仿真并支持后续机器人任务执行。最后，通过 MLLM Rating 与 Human Study 评价真实感、功能合理性、布局质量和文本忠实度，验证自动指标之外的整体视觉质量与人类主观偏好。最后还有定性展示哈！
% 具体实验量：20个场景，每个场景每个模型跑3次
% ddl: 9.9

\begin{table*}[t]
\centering
\caption{
\textbf{Quantitative comparison with prior scene-generation methods.}
Entries report mean $\pm$ sample standard deviation; $\uparrow$/$\downarrow$
denote higher/lower is better, \textbf{bold} denotes the best result.
}
\label{tab:historical_mean_sd}
\small
\setlength{\tabcolsep}{2pt}
\renewcommand{\arraystretch}{0.7}
\resizebox{\textwidth}{!}{%
\begin{tabular}{@{}lccccccccc@{}}
\toprule
\textbf{Method}
& \textbf{CF}$\uparrow$
& \textbf{SL}$\uparrow$
& \textbf{PP}$\uparrow$
& \textbf{TF}$\uparrow$
& \textbf{VQ}$\uparrow$
& \textbf{COL}$\downarrow$
& \textbf{NAV}$\uparrow$
& \textbf{OOB}$\downarrow$
& \textbf{OPN}$\uparrow$ \\
\midrule

Holodeck
& 5.56$\pm$1.72
& 5.47$\pm$1.50
& 6.02$\pm$1.09
& 7.32$\pm$3.15
& 5.64$\pm$0.98
& 0.62$\pm$6.24
& 0.9967$\pm$0.0073
& 0.37$\pm$0.84
& 0.5997$\pm$0.2951 \\

LayoutVLM
& 5.27$\pm$1.84
& 5.32$\pm$1.78
& 6.00$\pm$1.19
& 7.05$\pm$3.66
& 5.50$\pm$0.79
& 0.43$\pm$0.89
& 0.9954$\pm$0.0135
& 0.52$\pm$0.70
& 0.5417$\pm$0.4147 \\

SAGE
& 5.20$\pm$1.98
& 5.13$\pm$1.60
& 5.92$\pm$1.08
& 7.07$\pm$3.31
& 5.15$\pm$0.78
& 0.35$\pm$1.09
& 0.9976$\pm$0.0081
& 0.03$\pm$0.18
& 0.8500$\pm$0.2311 \\

EmbodiedGen V2
& 5.57$\pm$1.68
& 5.87$\pm$1.65
& 5.43$\pm$1.39
& 6.75$\pm$3.80
& 5.95$\pm$0.98
& 2.37$\pm$1.45
& \textbf{0.9988$\pm$0.0052}
& 0.19$\pm$0.48
& 0.9145$\pm$0.1363 \\

SceneSmith
& 5.34$\pm$1.44
& 6.18$\pm$1.38
& 6.59$\pm$1.01
& \textbf{8.92$\pm$2.33}
& 5.52$\pm$0.96
& 2.05$\pm$1.87
& 0.9916$\pm$0.0217
& \textbf{0.00$\pm$0.00}
& 0.9621$\pm$0.0742  \\

\midrule

\textbf{Ours}
& \textbf{7.13$\pm$1.48}
& \textbf{7.27$\pm$1.41}
& \textbf{7.68$\pm$0.83}
& 8.33$\pm$2.88
& \textbf{7.32$\pm$0.79}
& \textbf{0.29$\pm$0.35}
& 0.9984$\pm$0.0130
& \textbf{0.00$\pm$0.00}
& \textbf{0.9875$\pm$0.0481} \\
\bottomrule
\end{tabular}%
}
\end{table*}
\noindent\textbf{Evaluation setup.}
\label{sec:4.1}
We compare our method with five representative scene generation baselines: Holodeck~\cite{yang2024holodeck}, LayoutVLM~\cite{sun2025layoutvlm}, SAGE~\cite{xia2026sage}, EmbodiedGen V2~\cite{wang2026embodiedgenv2}, and SceneSmith~\cite{pfaff2026scenesmith}. All methods are evaluated on the same set of 20 indoor scene specifications spanning diverse room types and functional requirements, with three independent generations per scene for each method. 
We evaluate the generated environments from both perceptual and geometric perspectives. GPT-5.5 \citep{openai_gpt55} is provided with four rendered views of each scene and scores five perceptual metrics on a 1--10 scale: Content / Function (CF), Spatial Layout (SL), Physical Plausibility (PP), Text Fidelity (TF), and Visual Quality (VQ). We additionally report four geometry-based metrics: Collision (COL), Navigability (NAV), Out of Bound (OOB), and Opening Clearance / Openness (OPN). Detailed definitions of all metrics and used assets are provided in Appendix~\ref{app:assets} and~\ref{app:metrics}.

\noindent\textbf{Quantitative Results.}
\Cref{tab:historical_mean_sd} demonstrates that our method achieves SOTA performance on CF, SL, PP, VQ, COL, OOB, and OPN. 
Specifically, considering perceptual metrics, CF reaches $7.13\pm1.48$, compared to $5.57\pm1.68$ for the strongest baseline, representing a 28.0\% relative improvement. SL and PP reach $7.27\pm1.41$ and $7.68\pm0.83$, outperforming the strongest baselines by 17.6\% and 16.5\%, respectively, while VQ reaches $7.32\pm0.79$, improving over EmbodiedGen V2 ($5.95\pm0.98$) by 23.0\%. SceneSmith obtains the highest TF score of $8.92\pm2.33$, whereas our method yields a competitive score of $8.33\pm2.88$. 
Crucially, these gains in functional completeness and spatial organization do not occur at the expense of geometric usability. Our method achieves the lowest COL score of $0.29\pm0.35$, produces no out-of-bounds objects, maintains a NAV score of $0.9984\pm0.0130$ (comparable to the top baseline result of $0.9988\pm0.0052$), and achieves the highest OPN score of $0.9875\pm0.0481$. This combination is particularly crucial for embodied applications, as richer and more functionally complete scenes frequently introduce collisions, blocked openings, or fragmented traversable space. In contrast, our approach enhances scene content and organization while preserving navigability, valid boundaries, and access to openings.

\begin{figure}[t]
  \centering

  % 拼图子文件：只包含图片拼接，不包含浮动环境
  % 本文件只生成左侧拼图。
% 不包含 figure、wrapfigure、caption 或 label。

% 列宽变量
\ifdefined\methodcol\else
  \newlength{\methodcol}
\fi
\ifdefined\scenecol\else
  \newlength{\scenecol}
\fi

% 插入图片；找不到文件时显示占位框
\providecommand{\panel}[2][0 0 0 0]{%
  \IfFileExists{#2}{%
    \includegraphics[
      width=\linewidth,
      trim={#1},
      clip
    ]{#2}%
  }{%
    \fbox{%
      \parbox[c][0.6\linewidth][c]{%
        \dimexpr\linewidth-2\fboxsep-2\fboxrule\relax
      }{%
        \centering\tiny Missing PDF%
      }%
    }%
  }%
}

% 旋转的方法名称
\providecommand{\name}[1]{}
\renewcommand{\name}[1]{%
  \rotatebox[origin=c]{90}{\scriptsize\strut #1}%
}

% 图片下方的提示词
\providecommand{\prompt}[1]{%
  \parbox[t]{\linewidth}{%
    \centering\scriptsize\itshape ``#1''%
  }%
}

\centering
\setlength{\tabcolsep}{1pt}
\renewcommand{\arraystretch}{0.96}

% 此时 \linewidth 已经是父文件中左侧 minipage 的宽度。
% 直接使用，不要再套一个半宽 wrapfigure。
\setlength{\methodcol}{0.065\linewidth}
\setlength{\scenecol}{%
  \dimexpr(\linewidth-\methodcol-8\tabcolsep)/4\relax
}

\begin{tabular}{%
  @{}
  >{\centering\arraybackslash}m{\methodcol}
  *{4}{>{\centering\arraybackslash}m{\scenecol}}
  @{}
}
  \name{Holodeck}
    & \panel[114 32 113 34]{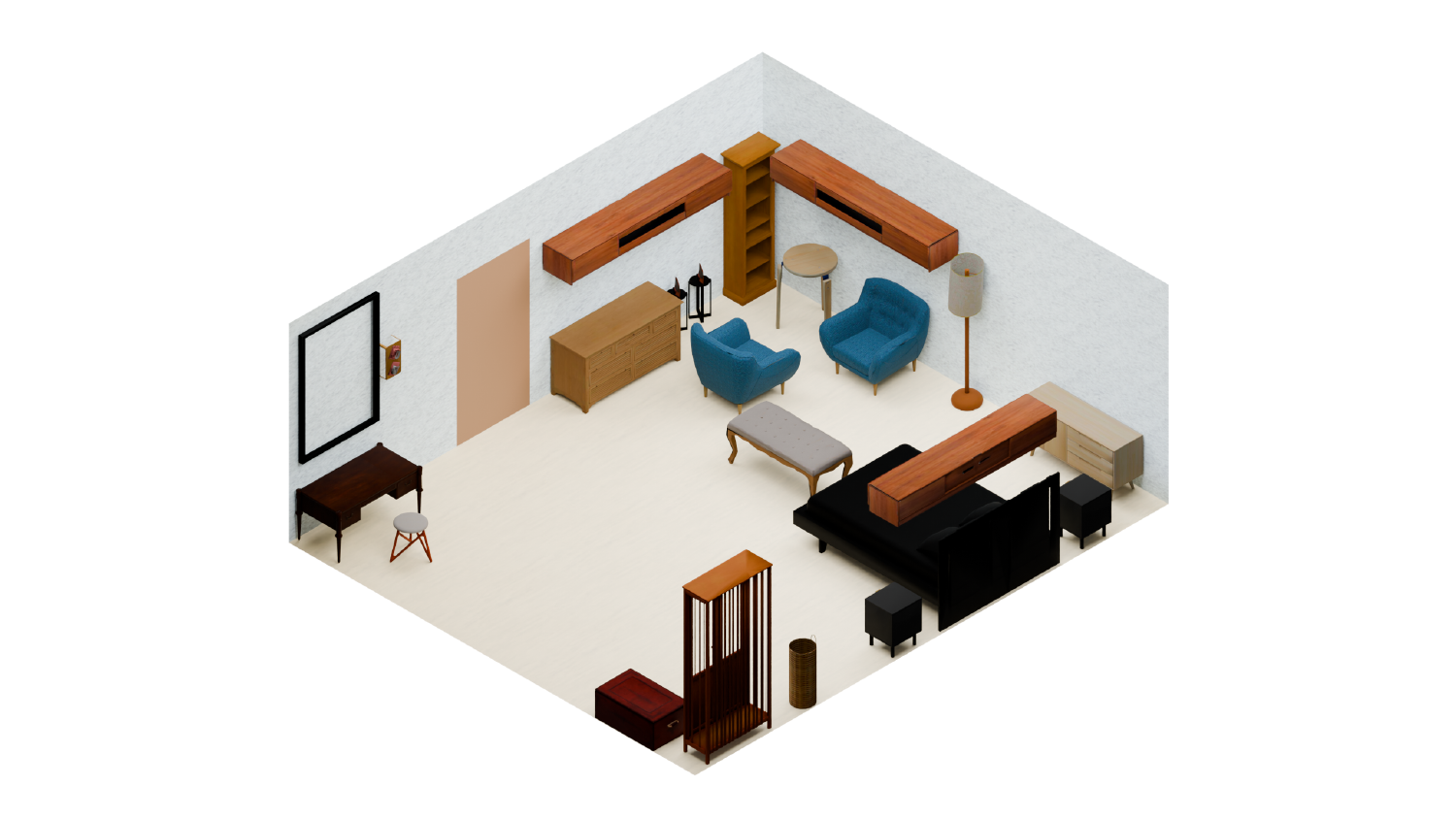}
    & \panel[114 32 113 34]{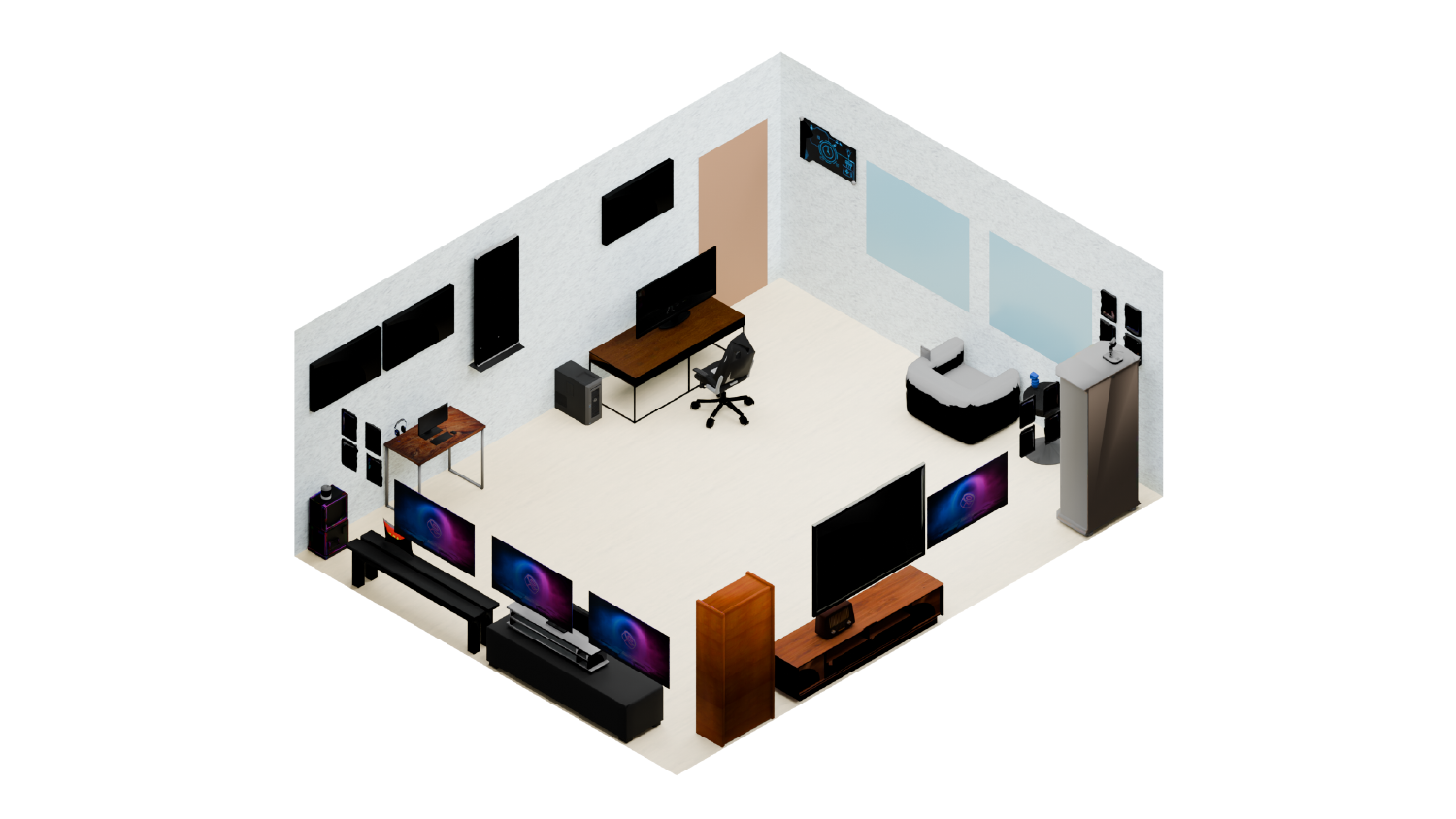}
    & \panel[114 32 113 34]{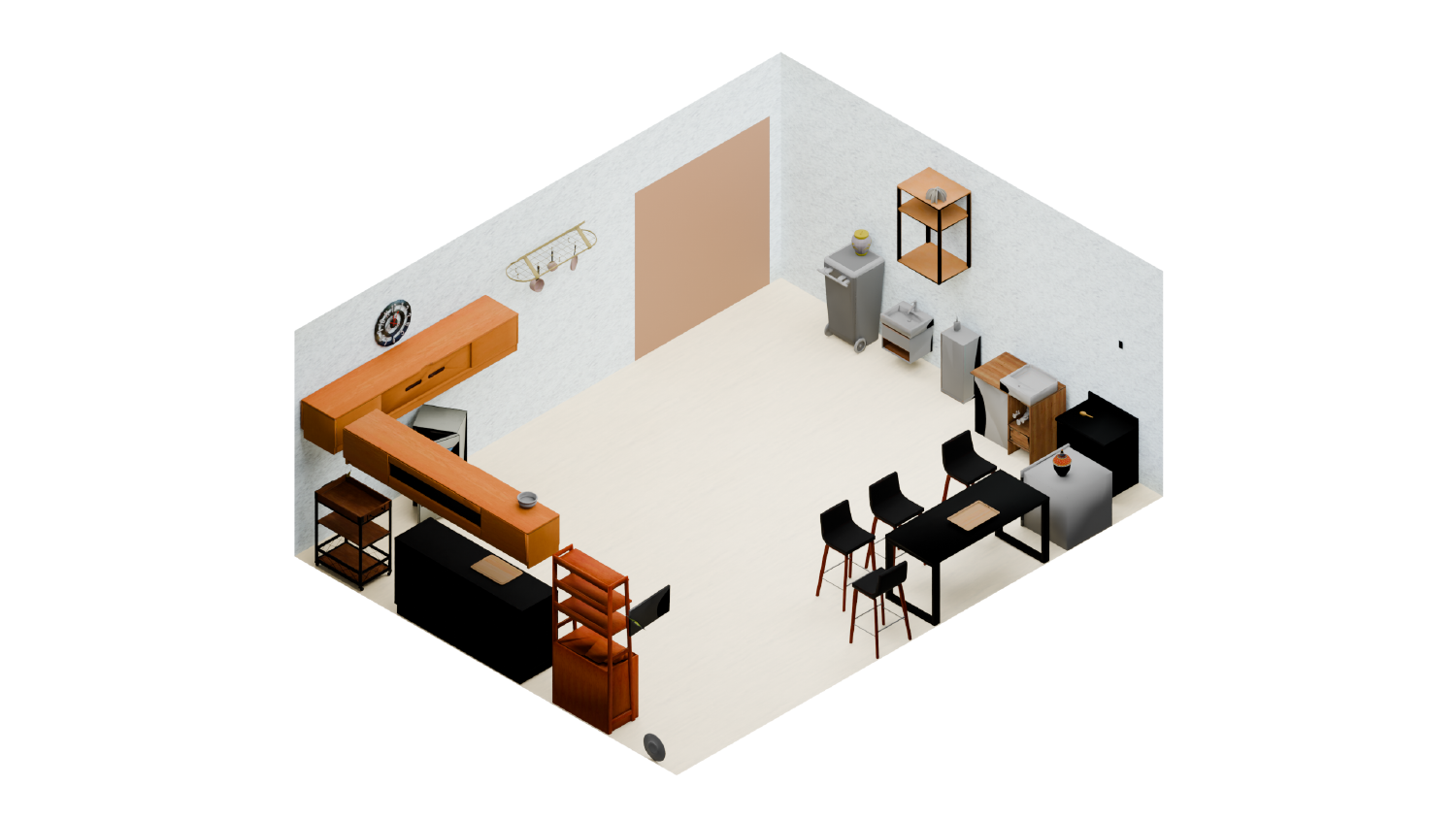}
    & \panel[114 32 113 34]{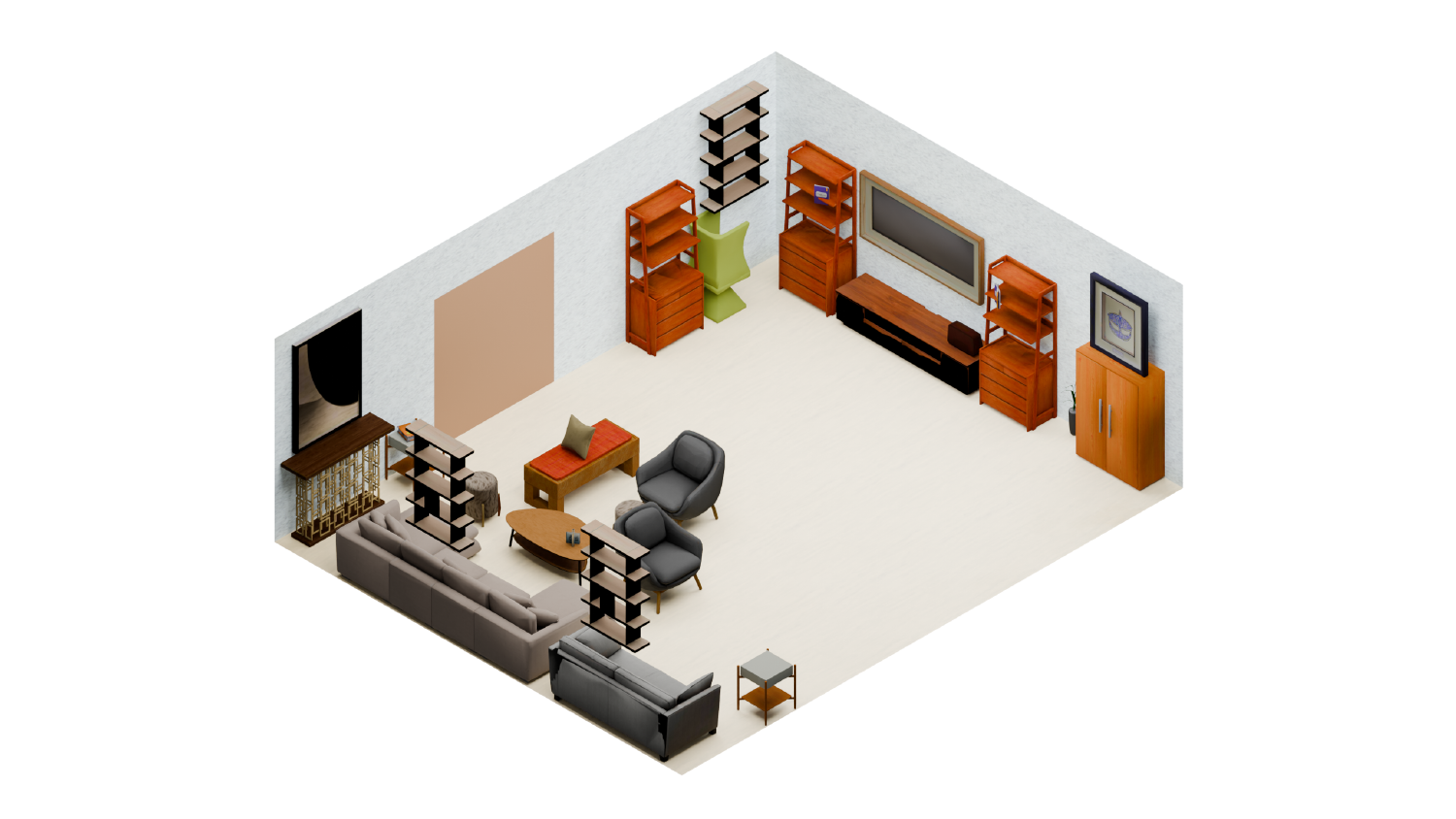}
    \\[1.5pt]

  \name{SAGE}
    & \panel[114 32 113 34]{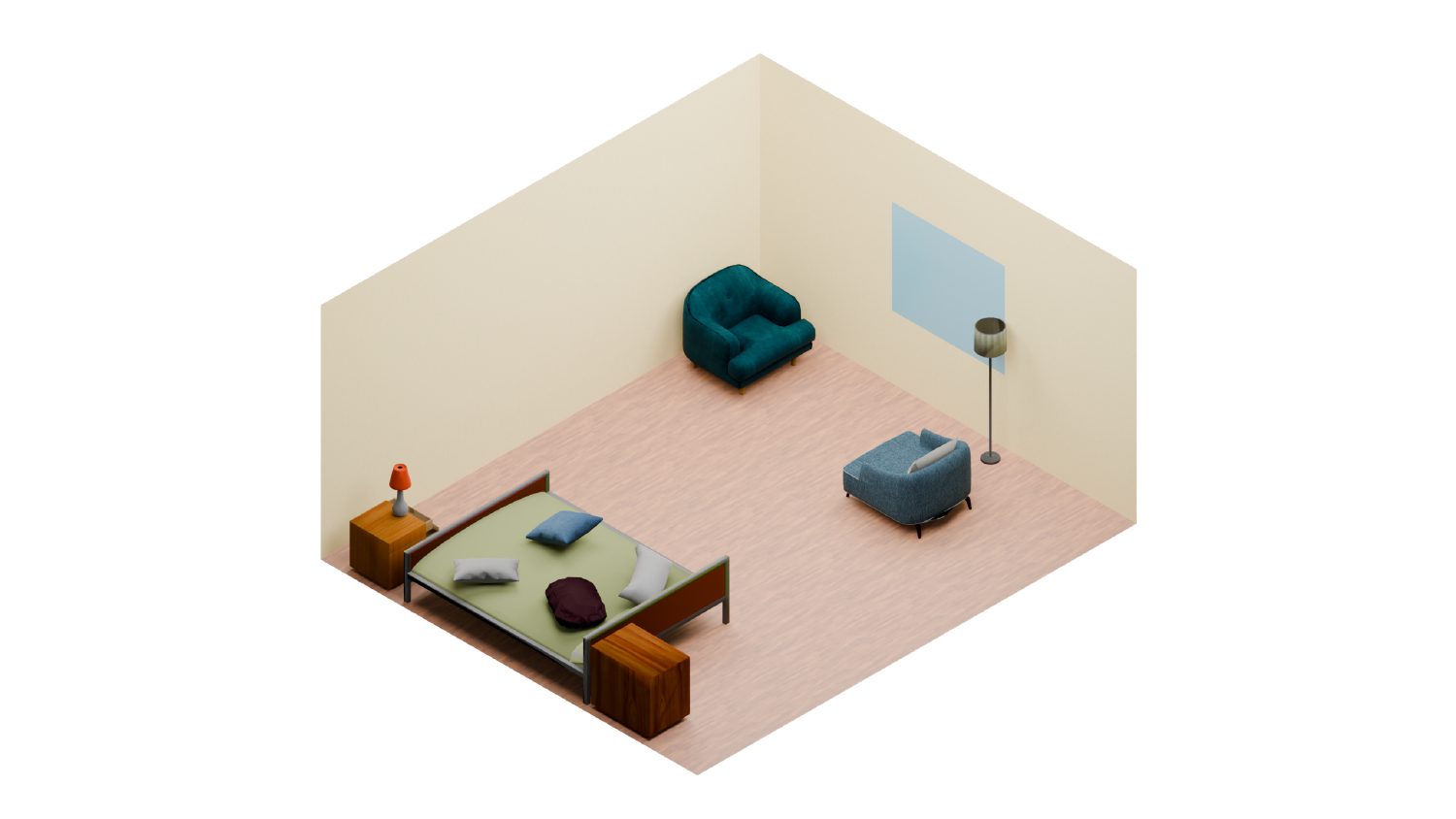}
    & \panel[114 32 113 34]{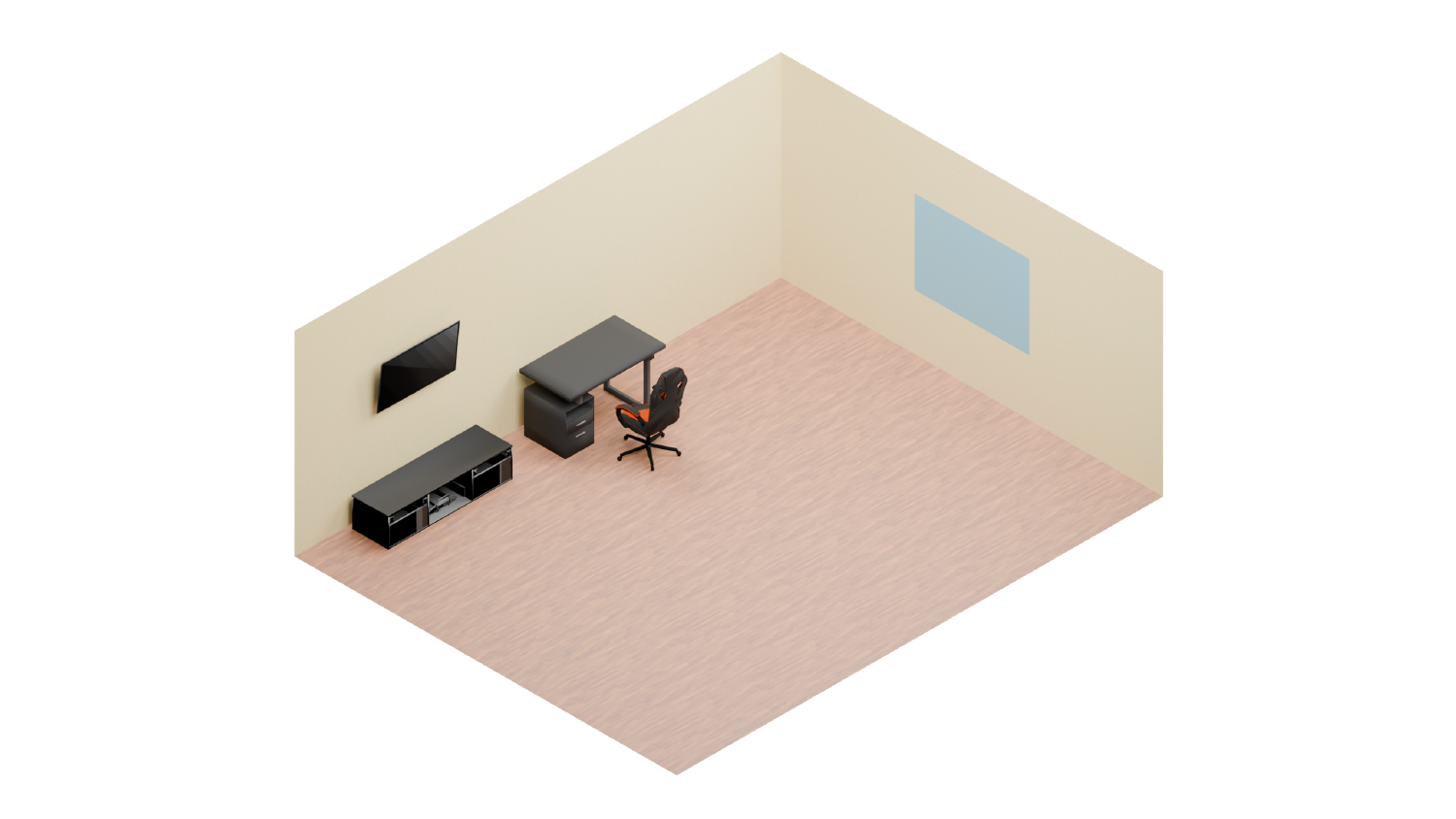}
    & \panel[114 32 113 34]{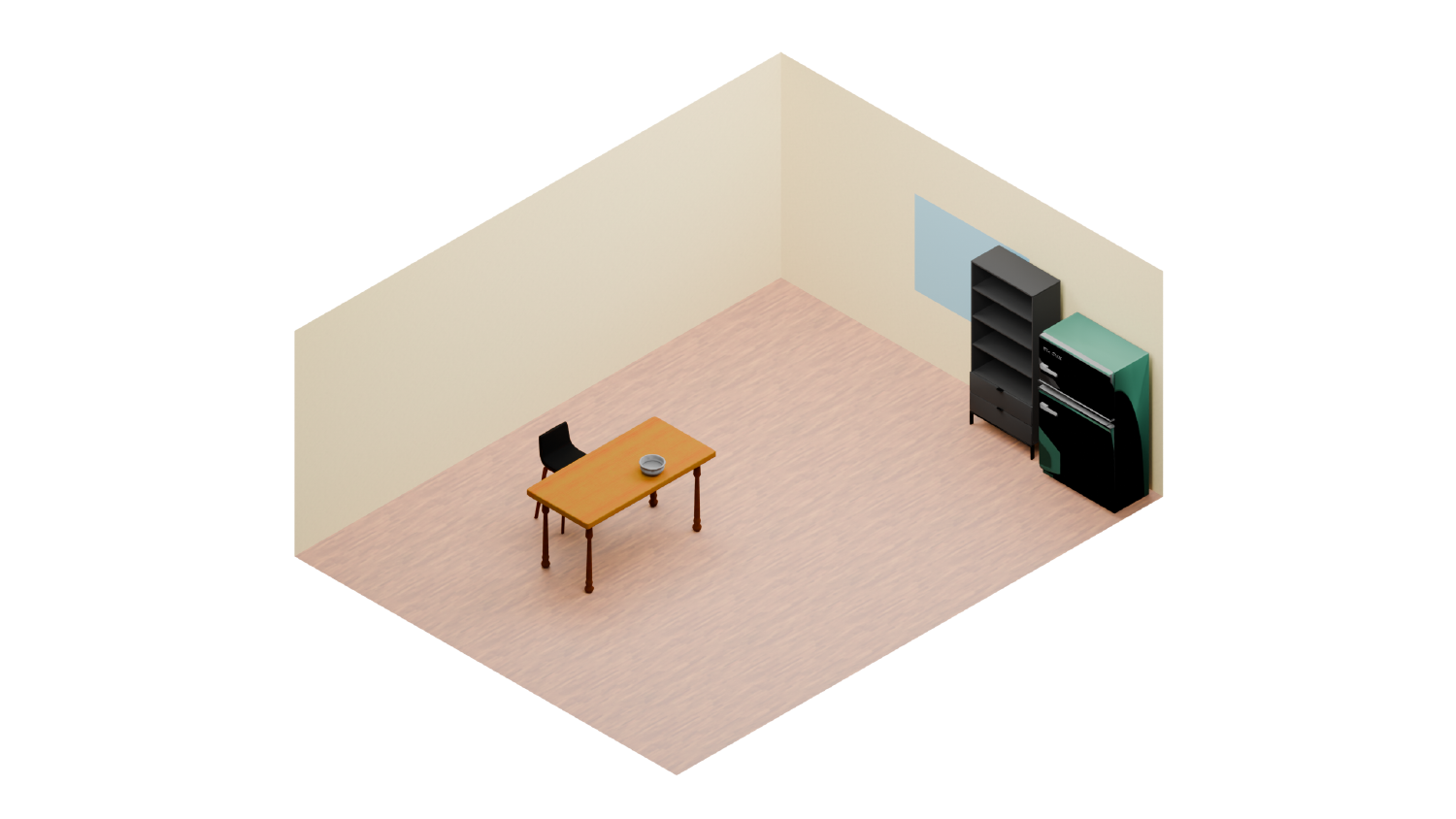}
    & \panel[114 32 113 34]{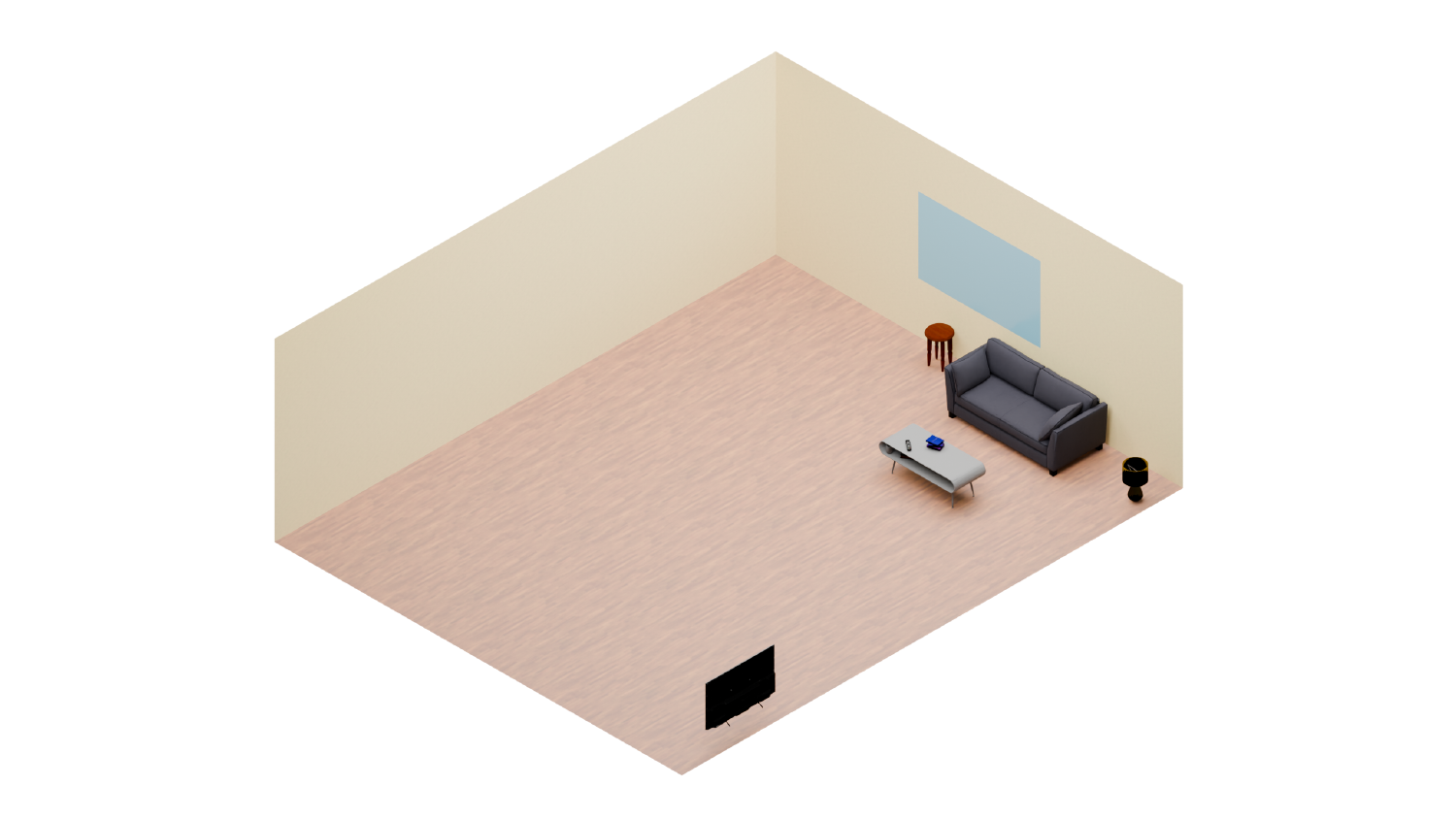}
    \\[1.5pt]

  \name{SceneSmith}
    & \panel[114 32 113 34]{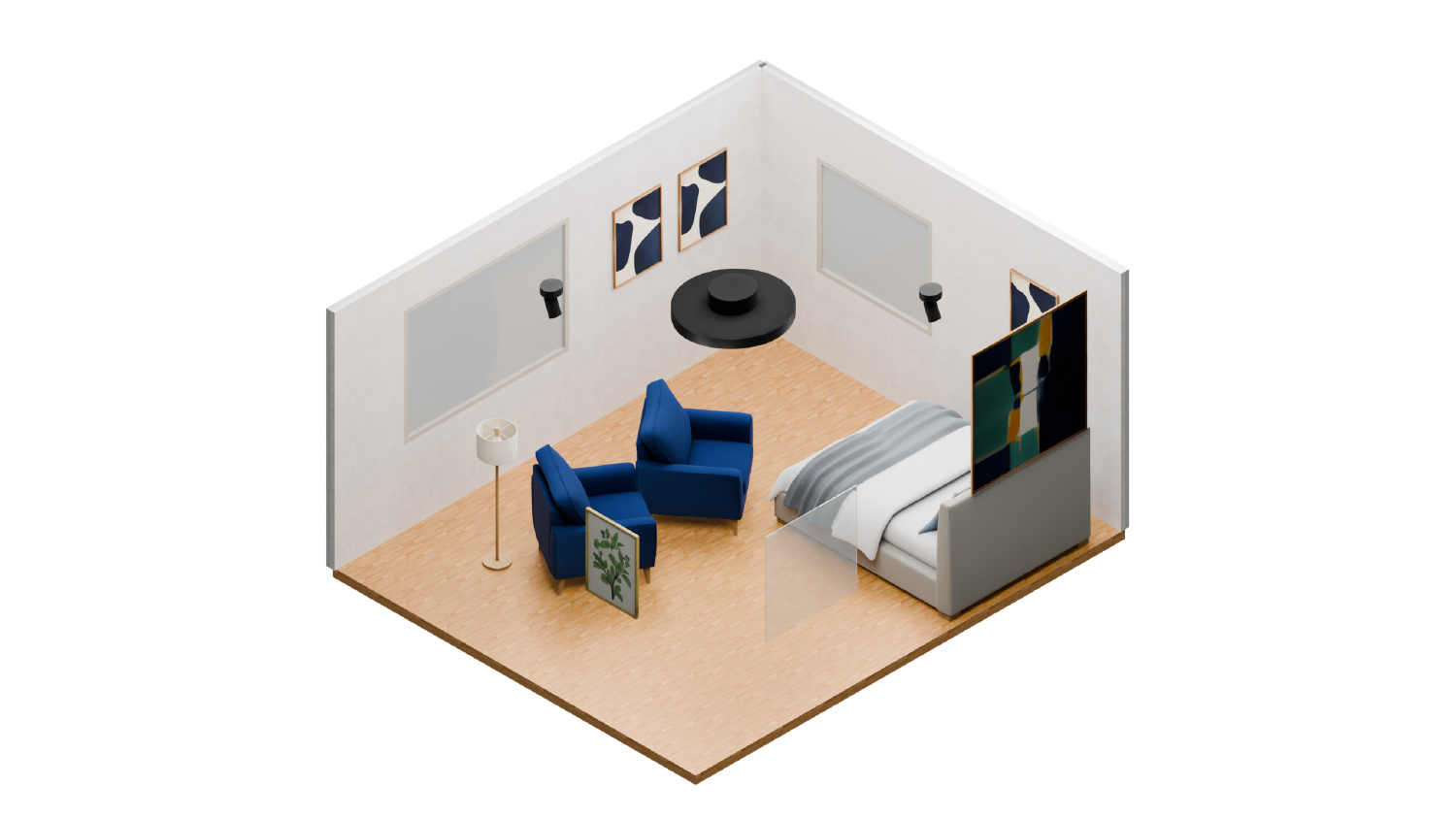}
    & \panel[114 32 113 34]{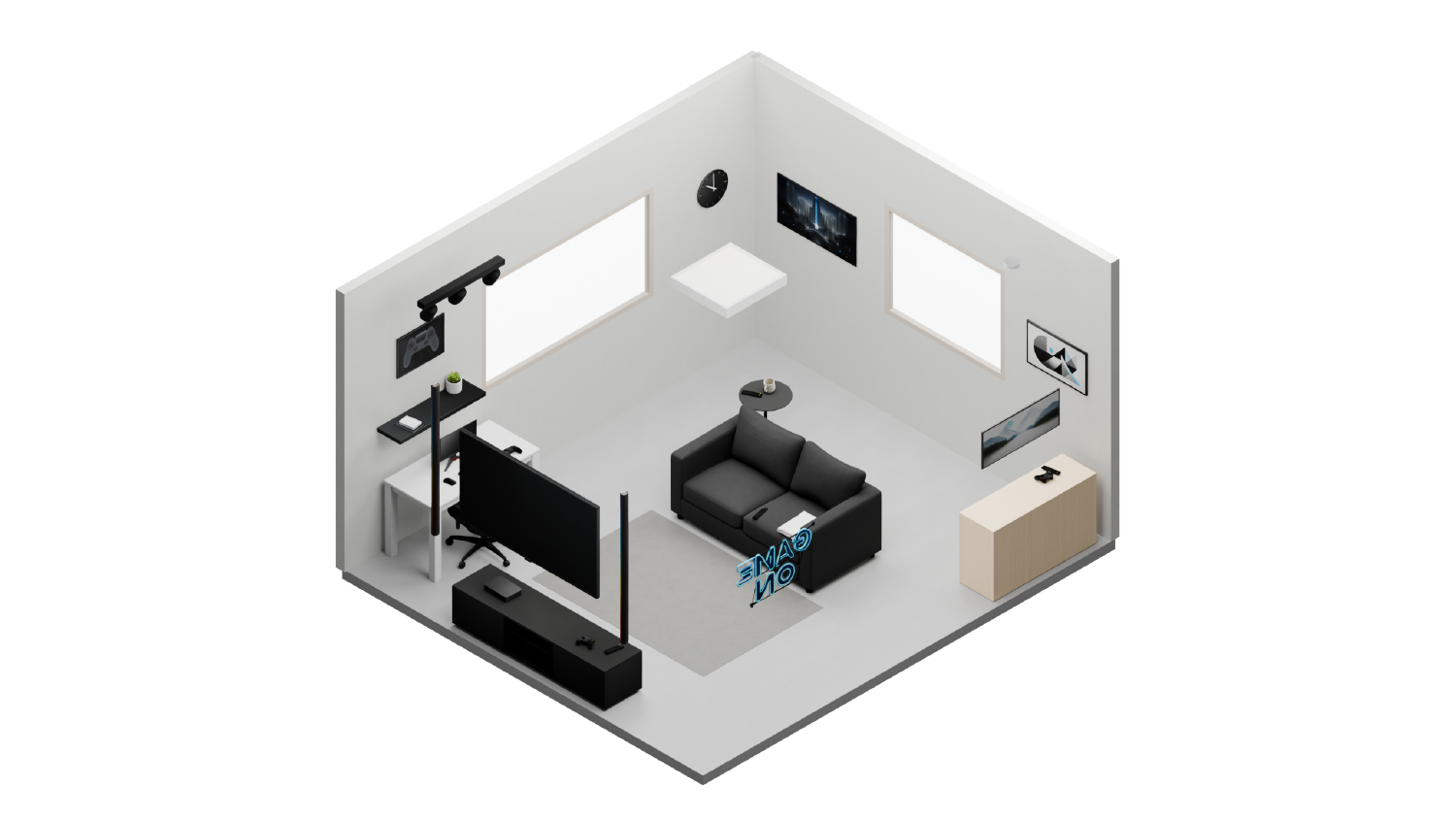}
    & \panel[114 32 113 34]{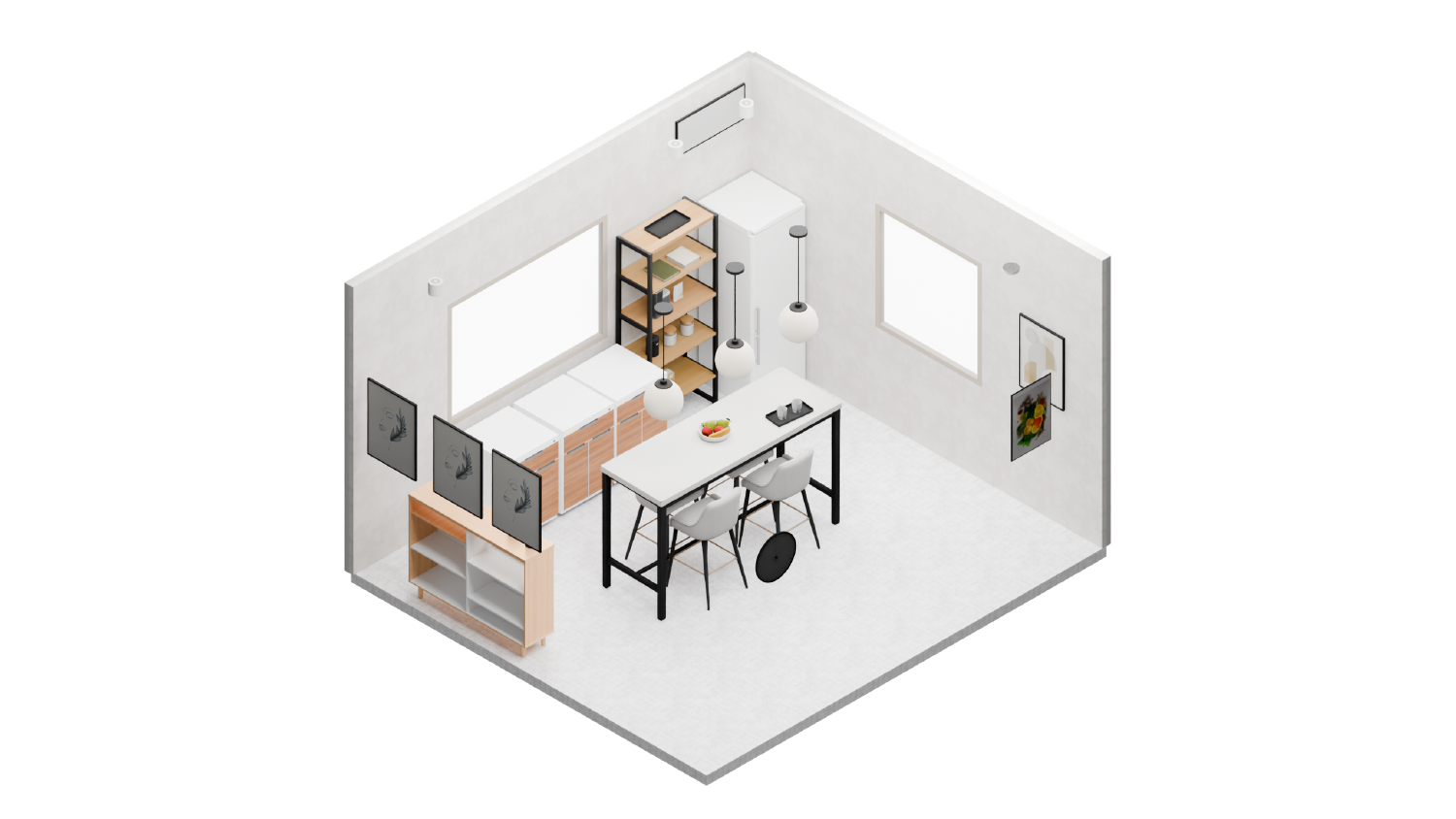}
    & \panel[114 32 113 34]{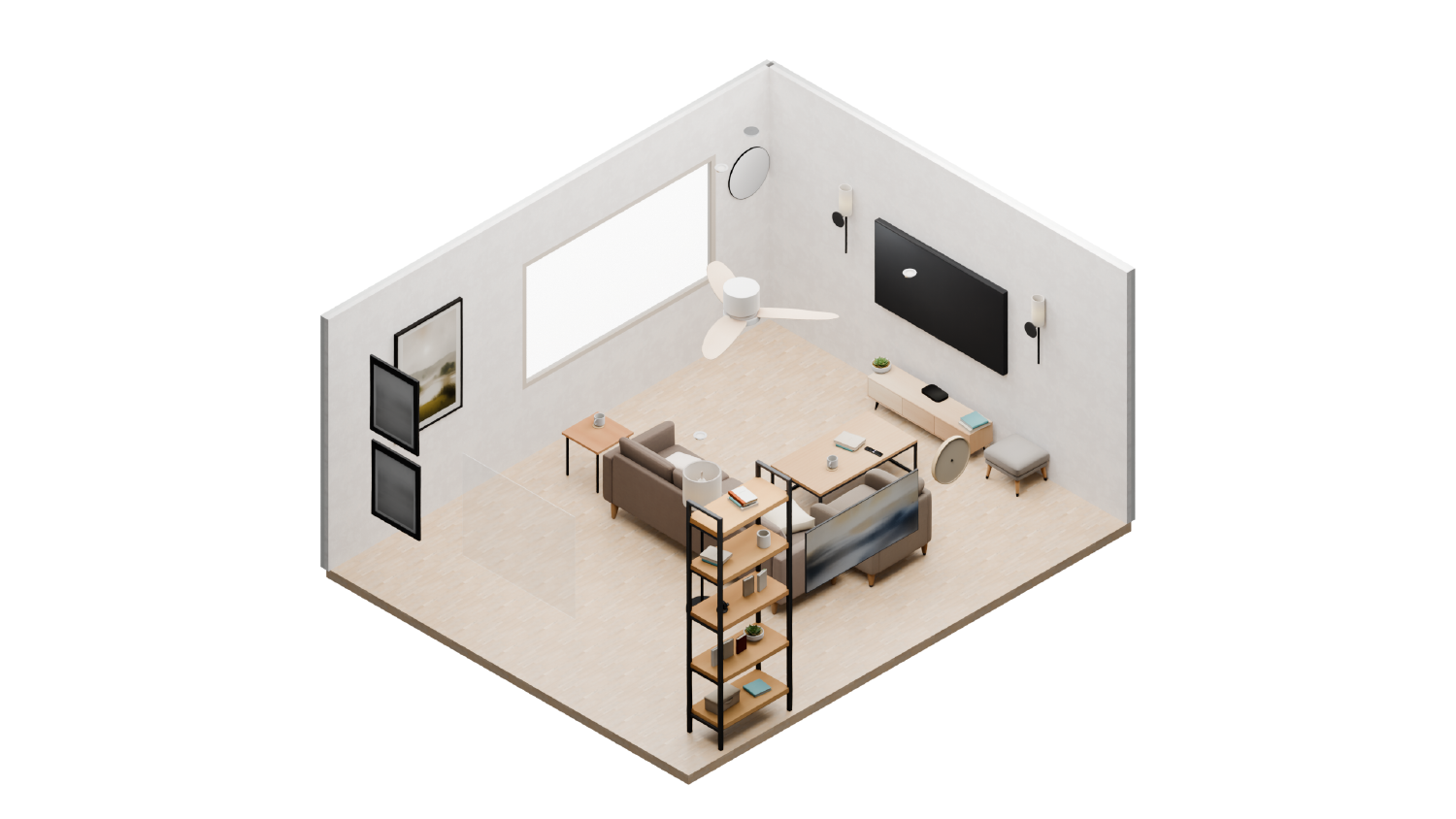}
    \\[1pt]

  \name{Ours}
    & \panel[114 32 113 34]{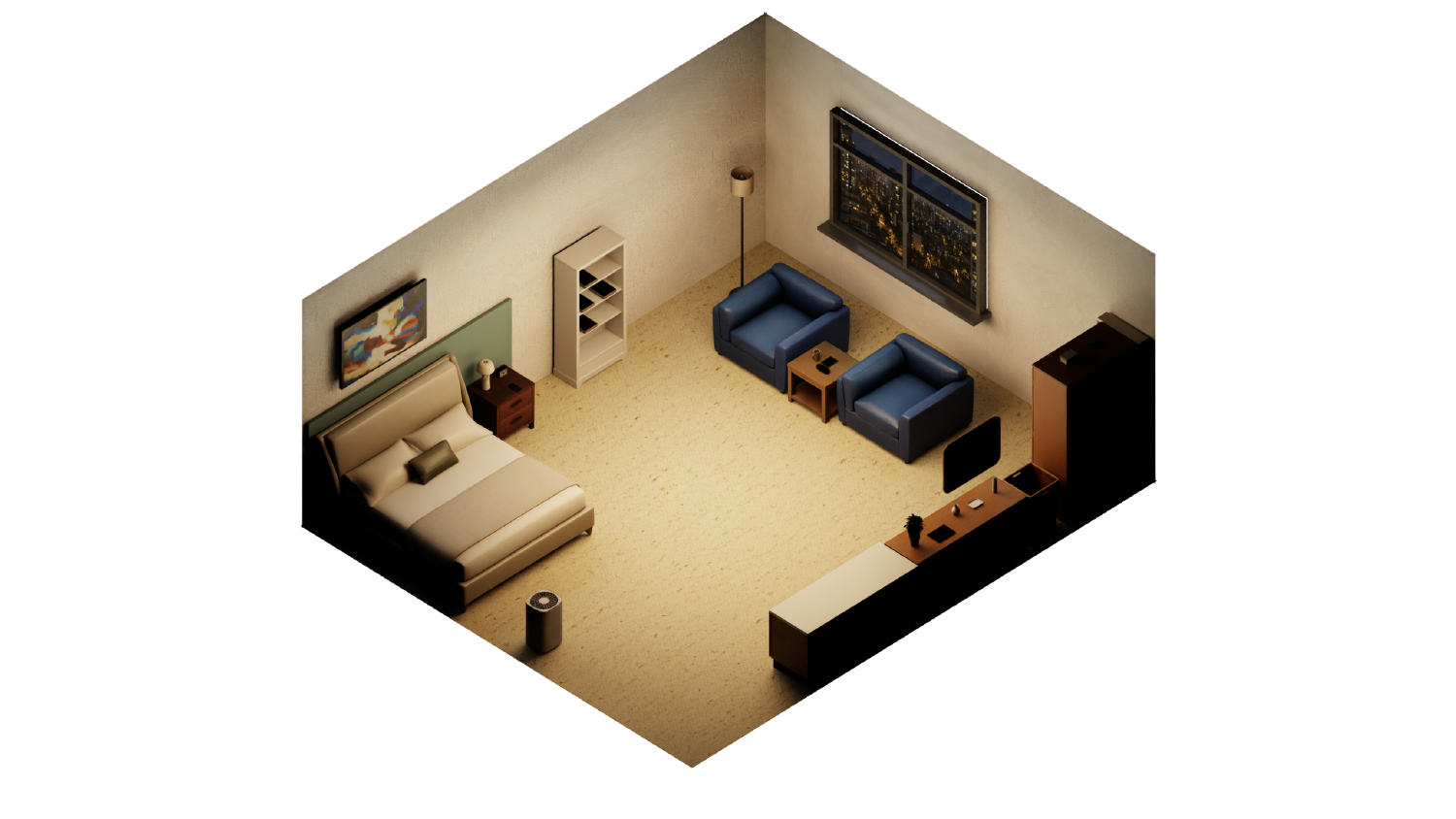}
    & \panel[114 32 113 34]{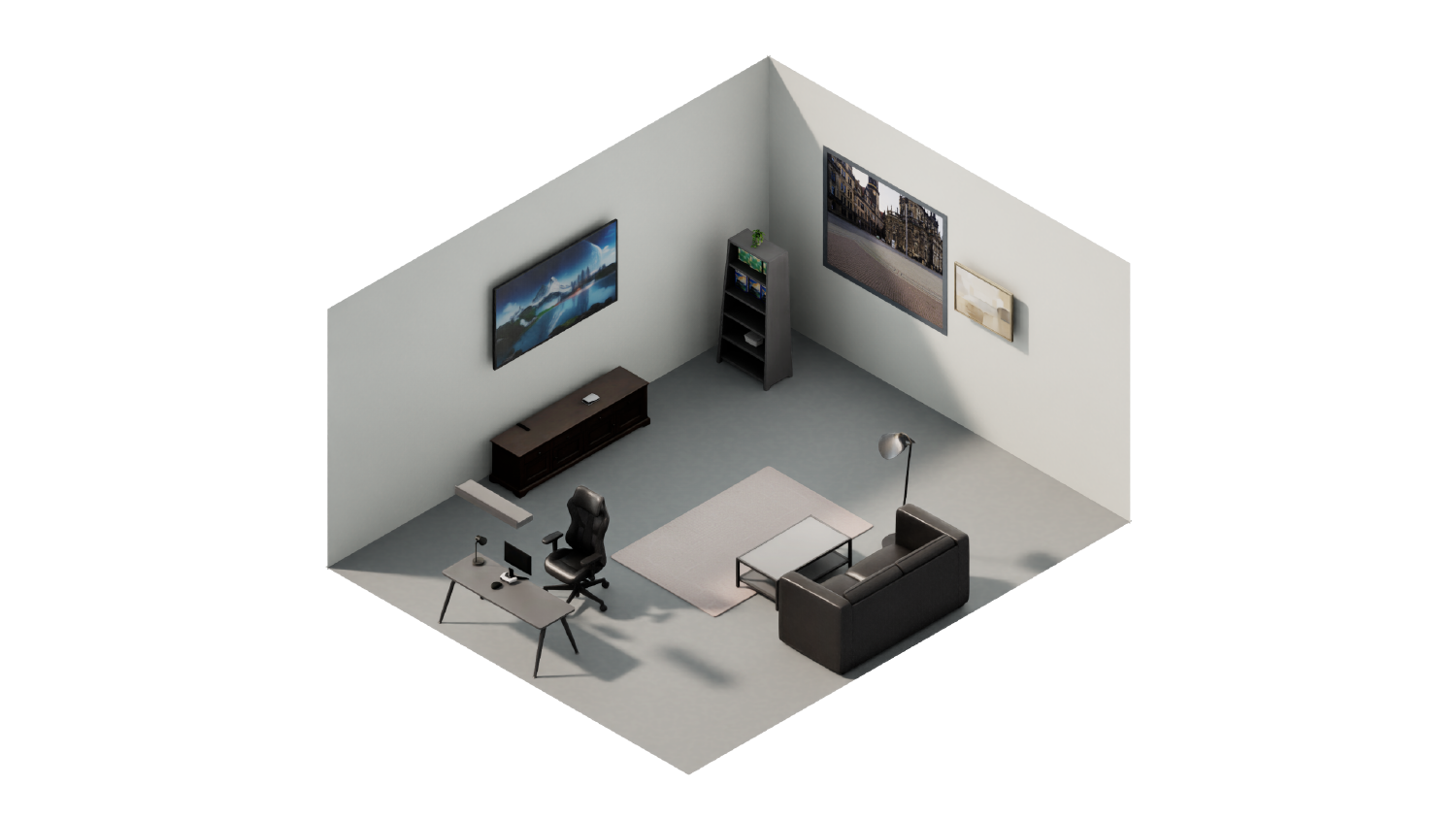}
    & \panel[114 32 113 34]{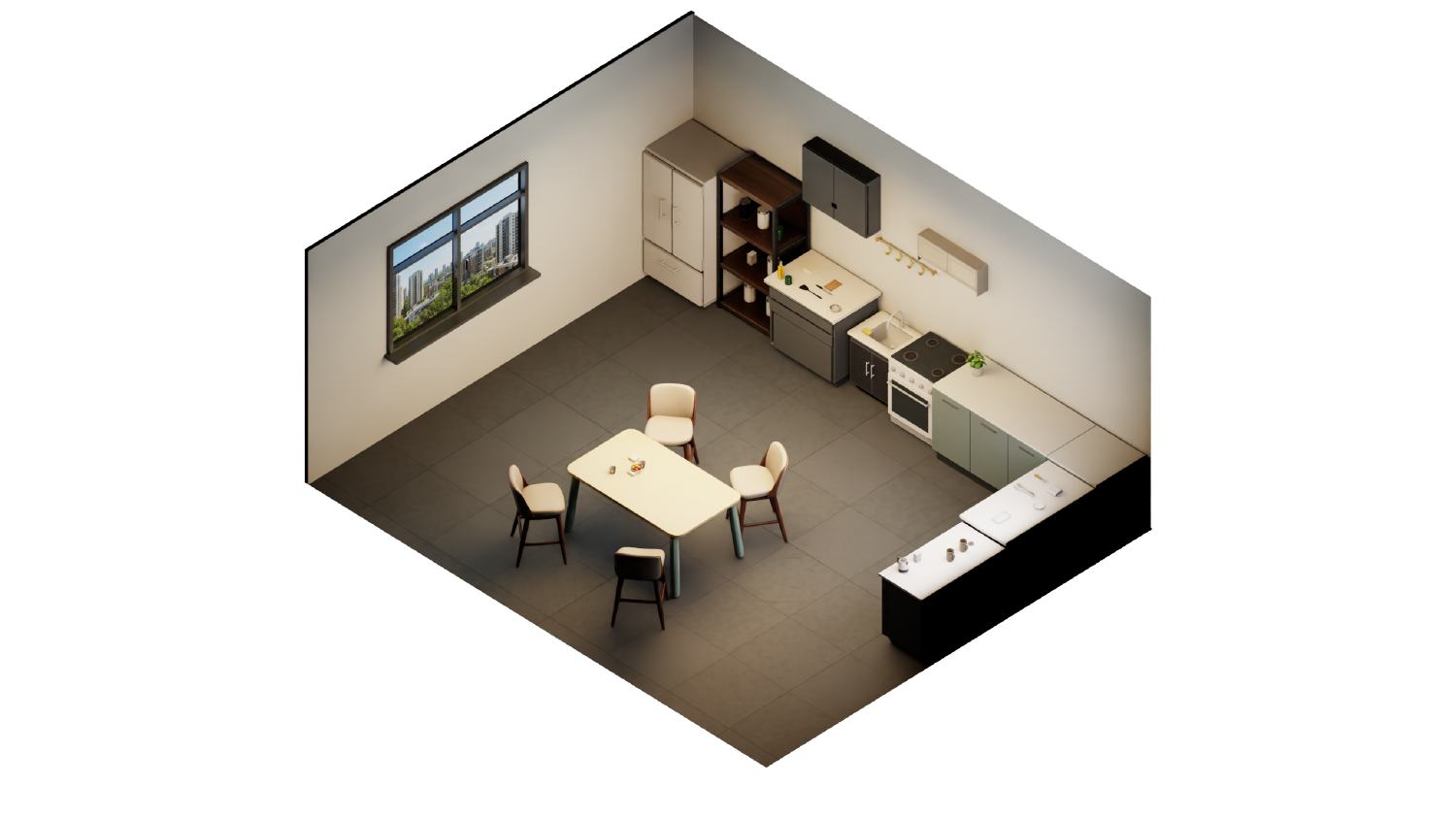}
    & \panel[114 32 113 34]{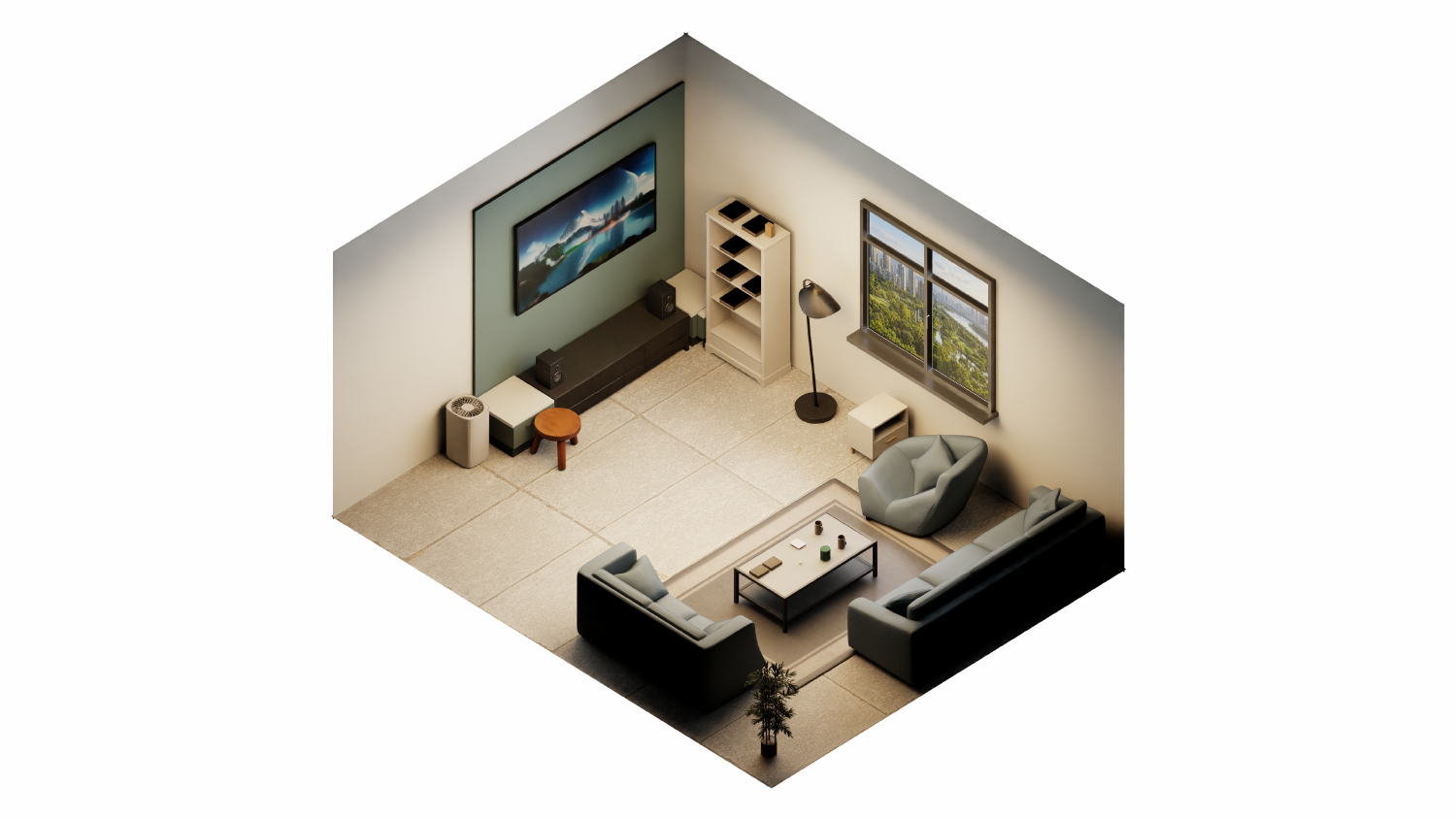}
    \\[2pt]

  \name{Prompt}
    & \prompt{Bedroom.}
    & \prompt{Gaming workspace.}
    & \prompt{Kitchen.}
    & \prompt{Living room.}
\end{tabular}
\par
  \par

  \caption{
    \textbf{Qualitative comparison with indoor scene generation baselines; our method yields more realistic layouts and more complete content.}
  }
  \label{fig:qualitative-comparison}
\end{figure}

\noindent\textbf{Qualitative Results.}
\Cref{fig:qualitative-comparison} further illustrates this distinction: Holodeck and SAGE often generate relatively sparse scenes, LayoutVLM frequently leaves large unused areas or weakly structured layouts, and EmbodiedGen V2 and SceneSmith generally create richer scenes but still exhibit locally inconsistent arrangements. Conversely, our method produces denser yet structured environments with clear functional regions and preserved circulation space. 

% Overall, these results demonstrate that our approach yields environments that are not only visually richer, but also functionally complete, spatially coherent, and geometrically suitable for downstream embodied interaction.
% @mingjian, zhiyi
% ddl: 9.9
\begin{figure}[t]
  \centering

  \includegraphics[width=\linewidth]{
    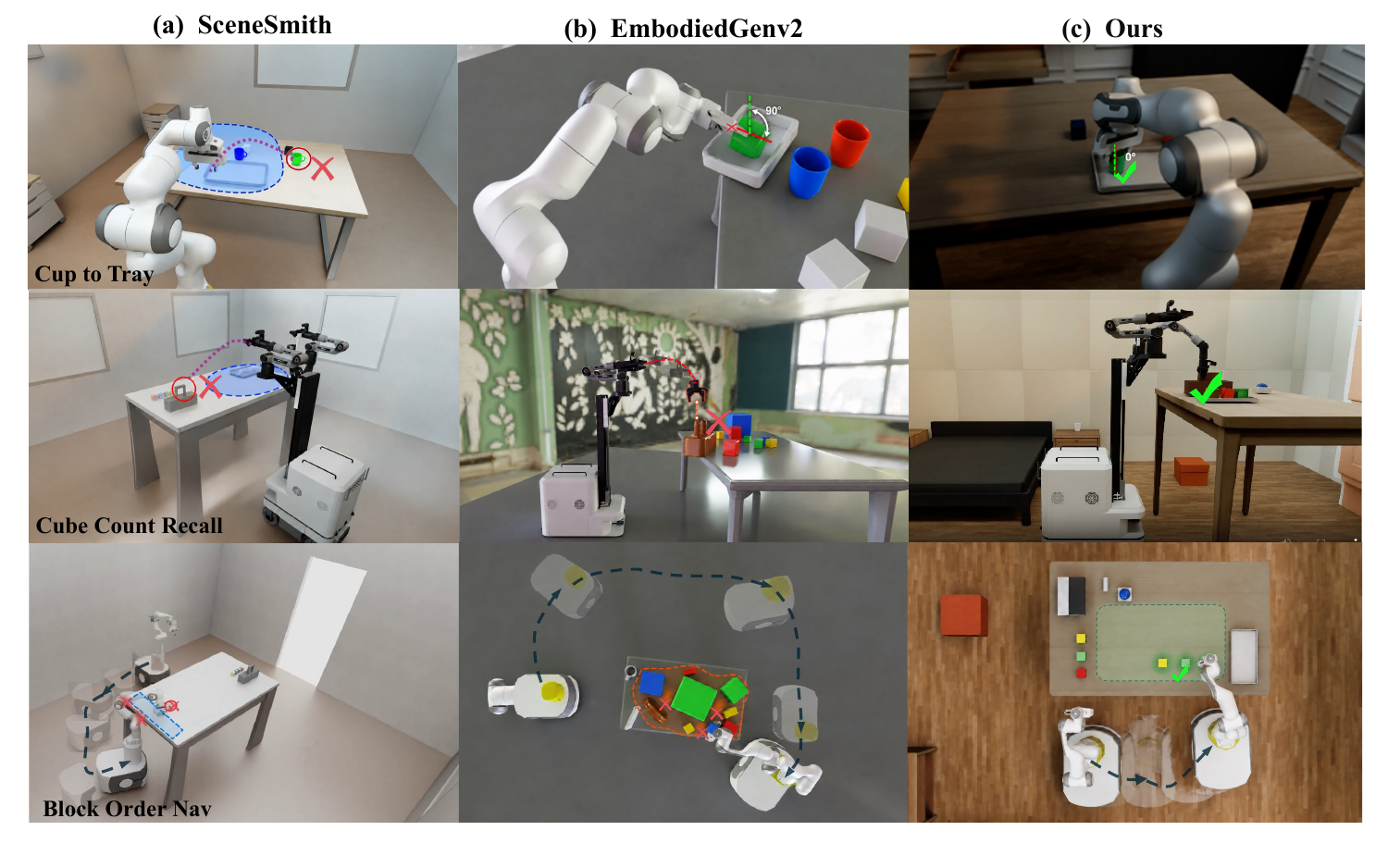
  }
  \par

  \caption{
    \textbf{Qualitative comparison.}
    Blue regions indicate reachable workspaces.
    Red regions indicate execution difficulties or failures.
    Green check marks indicate successful interactions;
    red crosses indicate failed interactions.
  }
  \label{fig:embodiment_rollouts}
\end{figure}
\subsection{Robotic Demonstration Generation}
\label{sec:4.2}
% 我们围绕两个问题评估端到端示范生成：(1) 相比基线，我们的场景生成方法是否更有利于下游示范生成？(2) 智能体的执行反馈闭环与机器人站位适配如何影响示范生成的成功率与成本？
We evaluate robotic trajectory generation through two questions: \textbf{Does our scene generation method better support downstream demonstration generation than the baselines?}
\textbf{How do execution-feedback-driven refinement and robot placement adaptation in our agent affect the success rate and cost of demonstration generation?}

\noindent\textbf{Experimental setup.}
% 主定量评测包含固定底座操纵与移动操纵共 16 个任务，固定底座操纵包含 5 个短程任务和 5 个长程任务而移动操纵包含 3个短程任务和 3个长程任务。场景生成对比采用相同的下游智能体，而智能体消融使用相同的初始场景；各组对比均保持任务描述、机器人配置和底层控制接口一致。成功示范须在规定的执行预算内满足预先定义的任务完成判据，并通过轨迹数据完整性检查；
% 规定的执行预算,完整任务列表、长短程划分和验收条件(轨迹数据完整性检查和任务完成判据)见附录。
Our main quantitative benchmark comprises 16 tasks: 10 for fixed-base manipulation and 6 for mobile manipulation, evenly split between short- and long-horizon tasks within each category.  A demonstration is accepted only if it satisfies predefined task completion predicates and passes trajectory data-integrity checks. The complete task list, horizon assignments, and acceptance criteria are provided in the Appendix~\ref{app:demonstration}. 

% \textbf{Scene generation comparisons} use the same downstream agent, while \textbf{agent ablations} share the same initial scenes; task specifications, robot configurations, and low-level control interfaces are held constant within each comparison.

\begin{table*}[t]
\centering
\caption{\textbf{Per-task demonstration generation performance on fixed-base and mobile manipulation.}
Short- and long-horizon tasks are distinguished by background color, with long-horizon tasks shaded in orange. Darker cells indicate the best result for each metric. \ding{55} indicates that EmbodiedGen V2 does not support generating the articulated assets required by the task. \textbf{m.} denotes memory tasks.}
\label{tab:task_specific_comparison}

\footnotesize
\setlength{\tabcolsep}{3pt}
\renewcommand{\arraystretch}{0.6}
\resizebox{\linewidth}{!}{
\begin{tabular}{ll*{9}{c}}
\toprule

\multirow{2}{*}{\textbf{Category}}
& \multirow{2}{*}{\textbf{Task}}
& \multicolumn{3}{c}{%
    \makecell[cc]{\textbf{SceneSmith}\\\textbf{+ Planner}}}
& \multicolumn{3}{c}{%
    \makecell[cc]{\textbf{EmbodiedGen V2}\\\textbf{+ Planner}}}
& \multicolumn{3}{c}{%
    \makecell[cc]{\textbf{\method}}} \\

\cmidrule(lr){3-5}
\cmidrule(lr){6-8}
\cmidrule(lr){9-11}

&
& \textbf{ASR}$\uparrow$
& \textbf{Top-5}$\uparrow$
& \textbf{Token}$\downarrow$
& \textbf{ASR}$\uparrow$
& \textbf{Top-5}$\uparrow$
& \textbf{Token}$\downarrow$
& \textbf{ASR}$\uparrow$
& \textbf{Top-5}$\uparrow$
& \textbf{Token}$\downarrow$ \\

\midrule

% =========================================================
% Fixed-base manipulation: Short-horizon
% =========================================================

\multirow{10}{*}{\shortstack[l]{Fixed-base\\manipulation}}
& Open Drawer
& $32\%$ & $40\%$
& \cellcolor{blue!15}$\mathbf{8782}$
& \ding{55} & \ding{55} & \ding{55}
& \cellcolor{blue!15}$\mathbf{78\%}$
& \cellcolor{blue!15}$\mathbf{80\%}$
& $10885$ \\

& Cup to Tray
& $19\%$ & $24\%$ & $9672$
& $35\%$ & $40\%$ & $8721$
& \cellcolor{blue!15}$\mathbf{64\%}$
& \cellcolor{blue!15}$\mathbf{72\%}$
& \cellcolor{blue!15}$\mathbf{4288}$ \\

& Click Button
& $37\%$ & $44\%$ & $69814$
& $89\%$
& \cellcolor{blue!15}$\mathbf{100\%}$
& \cellcolor{blue!15}$\mathbf{9235}$
& \cellcolor{blue!15}$\mathbf{100\%}$
& \cellcolor{blue!15}$\mathbf{100\%}$
& $16193$ \\

& Basket to Tray
& $69\%$ & $78\%$ & $43879$
& $0\%$ & $0\%$
& \cellcolor{blue!15}$\mathbf{13838}$
& \cellcolor{blue!15}$\mathbf{70\%}$
& \cellcolor{blue!15}$\mathbf{82\%}$
& $23221$ \\

& Open Cabinet
& $0\%$ & $0\%$ & $713248$
& \ding{55} & \ding{55} & \ding{55}
& \cellcolor{blue!15}$\mathbf{35\%}$
& \cellcolor{blue!15}$\mathbf{42\%}$
& \cellcolor{blue!15}$\mathbf{20429}$ \\

\cmidrule(lr){2-11}

% =========================================================
% Fixed-base manipulation: Long-horizon
% =========================================================

& \cellcolor{orange!12}Cup Pairing (\textbf{m.})
& $0\%$ & $0\%$
& \cellcolor{blue!15}$\mathbf{62518}$
& $12\%$ & $20\%$ & $138551$
& \cellcolor{blue!15}$\mathbf{51\%}$
& \cellcolor{blue!15}$\mathbf{58\%}$
& $120071$ \\

& \cellcolor{orange!12}Block Order (\textbf{m.})
& $7\%$ & $10\%$ & $201026$
& $0\%$ & $0\%$ & $87644$
& \cellcolor{blue!15}$\mathbf{83\%}$
& \cellcolor{blue!15}$\mathbf{88\%}$
& \cellcolor{blue!15}$\mathbf{80386}$ \\

& \cellcolor{orange!12}Cube Count (\textbf{m.})
& $3\%$ & $6\%$ & $186377$
& $0\%$ & $0\%$ & $177621$
& \cellcolor{blue!15}$\mathbf{79\%}$
& \cellcolor{blue!15}$\mathbf{84\%}$
& \cellcolor{blue!15}$\mathbf{90316}$ \\

& \cellcolor{orange!12}Button History (\textbf{m.})
& $15\%$ & $26\%$ & $67662$
& $71\%$ & $80\%$ & $93251$
& \cellcolor{blue!15}$\mathbf{83\%}$
& \cellcolor{blue!15}$\mathbf{92\%}$
& \cellcolor{blue!15}$\mathbf{27611}$ \\

& \cellcolor{orange!12}Object Gesture (\textbf{m.})
& $59\%$
& \cellcolor{blue!15}$\mathbf{82\%}$
& \cellcolor{blue!15}$\mathbf{57626}$
& $0\%$ & $0\%$ & $88343$
& \cellcolor{blue!15}$\mathbf{62\%}$
& $74\%$
& $72532$ \\

\midrule

% =========================================================
% Mobile manipulation: Short-horizon
% =========================================================

\multirow{6}{*}{\shortstack[l]{Mobile\\manipulation}}
& Open Drawer
& $0\%$ & $0\%$ & $161643$
& \ding{55} & \ding{55} & \ding{55}
& \cellcolor{blue!15}$\mathbf{43\%}$
& \cellcolor{blue!15}$\mathbf{62\%}$
& \cellcolor{blue!15}$\mathbf{16389}$ \\

& Cup to Tray
& $22\%$ & $44\%$ & $118139$
& $0\%$ & $0\%$ & $12315$
& \cellcolor{blue!15}$\mathbf{79\%}$
& \cellcolor{blue!15}$\mathbf{96\%}$
& \cellcolor{blue!15}$\mathbf{5396}$ \\

& Lift Cover
& $7\%$ & $14\%$ & $140514$
& $40\%$ & $80\%$ & $22415$
& \cellcolor{blue!15}$\mathbf{88\%}$
& \cellcolor{blue!15}$\mathbf{100\%}$
& \cellcolor{blue!15}$\mathbf{4979}$ \\

% & Open Microwave+Nav
% & \tbd & \tbd & \tbd
% & \ding{55} & \ding{55} & \ding{55}
% & $1\%$ & $2\%$ & $21008$ \\

% & Open Cabinet+Nav
% & $0\%$ & $0\%$ & $85134$
% & \ding{55} & \ding{55} & \ding{55}
% & \tbd & \tbd & \tbd \\

\cmidrule(lr){2-11}

% =========================================================
% Mobile manipulation: Long-horizon
% =========================================================

& \cellcolor{orange!12}Book Stow(\textbf{m.})
& $0\%$ & $0\%$ & $90168$
& \ding{55} & \ding{55} & \ding{55}
& \cellcolor{blue!15}$\mathbf{17\%}$
& \cellcolor{blue!15}$\mathbf{34\%}$
& \cellcolor{blue!15}$\mathbf{26385}$ \\

& \cellcolor{orange!12}Cup Pairing(\textbf{m.})
& $0\%$ & $0\%$ & $137257$
& $0\%$ & $0\%$ & $221043$
& \cellcolor{blue!15}$\mathbf{52\%}$
& \cellcolor{blue!15}$\mathbf{62\%}$
&  \cellcolor{blue!15}$\mathbf{73892}$ \\

& \cellcolor{orange!12}Block Order(\textbf{m.})
& $0\%$ & $0\%$ & $124671$
& $0\%$ & $0\%$ & $219762$
& \cellcolor{blue!15}$\mathbf{93\%}$
& \cellcolor{blue!15}$\mathbf{100\%}$
& \cellcolor{blue!15}$\mathbf{50283}$ \\

% & Microwave Storage
% & \tbd & \tbd & \tbd
% & \ding{55} & \ding{55} & \ding{55}
% & \tbd & \tbd & $118864$ \\

% & Box Store \& Sort
% & \tbd & \tbd & \tbd
% & \ding{55} & \ding{55} & \ding{55}
% & \tbd & \tbd & \tbd \\

\midrule

\multicolumn{2}{l}{\textbf{\textit{Average}}}
& $17\%$ & $23\%$ & $137062$
& $21\%$ & $27\%$ & $91062$
& \cellcolor{blue!15}$\mathbf{67\%}$
& \cellcolor{blue!15}$\mathbf{77\%}$
& \cellcolor{blue!15}$\mathbf{40204}$ \\

\bottomrule
\end{tabular}}
\end{table*}

% \input{fig/Experiment/qualitative}
% \noindent\textbf{演示生成对比。}
% 我们将本方法与 SceneSmith 和 EmbodiedGen V2 进行比较，各场景生成方法均接入相同的下游示范生成智能体。对于每个任务，各方法根据相同的任务描述和机器人配置生成场景，随后固定场景资产与物体布局。下游智能体采用相同的模型、控制接口、机器人站位适配能力和最多 30 轮的执行反馈预算，以评估不同生成场景对示范生成的支持程度。
% 在每个生成场景中，我们运行 10 个独立的智能体会话，每次会话均从相同的初始场景状态开始，生成一个候选任务计划及其对应的机器人初始站位。会话结束后，任务计划与站位均被冻结，并在 10 个共享的、未用于生成或修订的仿真随机种子下进行评估。我们报告所有候选与任务上的平均成功率（ASR）、每个任务中表现最好的五个候选的平均成功率（Top-5 ASR），以及每次会话累计的平均 LLM 输出 token 数。未能产生可执行任务计划的会话按零成功率计入，并保留其 token 开销。主表分别报告固定底座与移动操纵中短程和长程任务的结果。
% We compare our method with SceneSmith and EmbodiedGen V2. All three scene generators are paired with the same downstream agent workflow for demonstration generation. For each task, each method generates a scene using the same task specification and robot configuration.
% The scene assets and object layout are then held fixed
% throughout downstream planning and refinement.
\noindent\textbf{Demonstration Generation Comparison.}
To evaluate how reliably each generated scene supports
demonstration generation, we run 10 independent agent sessions
from the same initial scene state to produce 10 simulation-ready task configurations as described in \Cref{sec:trajec_gen} for each task. We then evaluate each configuration using the same set of 10 simulation seeds, which
are not used during generation or refinement.

We report \textbf{ASR} (mean success rate across all candidates and tasks), \textbf{Top-5 ASR} (mean success rate of the five best-performing candidates per task) and \textbf{Token} (mean VLM output tokens per session). Sessions that produce no executable task plan are assigned zero success and included in the token accounting. As shown in \Cref{tab:task_specific_comparison},
\method~outperforms each baseline in ASR on its supported tasks.
For our method, the mobile setting performs better on
\textit{Cup to Tray} but worse on \textit{Open Drawer},
suggesting that navigation may improve reachability,
while grasp-feasible placement may still be unsuitable
for joint-constrained opening motions.

% 通过定性分析,我们发现SceneSmith的场景由于在场景生成时没有考虑到机器人构型和任务空间,通常存在生成的场景机器人不可达的情况,即使agent会综合考虑操作起点与目标位置、机器人运动学约束和碰撞间隙，选择机器人的初始底座位置与朝向，使尽可能多的任务所需交互处于可达范围内。但是一般还是无解从而导致数据生成成功率大幅度下降,而EmbodiedGenv2相比而言生成的场景不具备良好的物体大小一致性(比如方块大小不一),而且生成的场景十分杂乱不利于机器人操作进行数据搜集,相比之下我们的方法xxx
To understand these performance differences, we qualitatively examine the generated scenes in \Cref{fig:embodiment_rollouts}. Some SceneSmith scenes have unsuitable support-surface heights or widely separated interaction locations, limiting reachability despite robot placement adaptation. Scenes generated by EmbodiedGen V2 exhibit inconsistent object
scales and cluttered layouts that hinder manipulation. In contrast, our method better aligns scene layouts with task requirements and the robot's workspace, facilitating successful demonstration generation.

% 我们在相同的初始场景上进行两项消融：\textit{w/o Feedback} 保留初始站位选择与任务规划，但关闭执行反馈驱动的修订；\textit{w/o Placement Validation} 仅依据三视图与物体世界坐标选择机器人站位，不考虑运动学与碰撞约束，也不验证站位可行性，同时保留程序反馈修订与正常导航。其余设置与评估协议保持一致，并采用 ASR 衡量各组件对生成成功率的贡献。

\noindent\textbf{Ablations of RSI Design.}
% 我们在相同的初始场景上进行两项消融：\textit{w/o Feedback} 保留初始站位选择与任务规划，但关闭执行反馈驱动的修订；\textit{w/o Placement Validation} 仅依据三视图与物体世界坐标选择机器人站位，不考虑运动学与碰撞约束，也不验证站位可行性，同时保留程序反馈修订与正常导航。其余设置与评估协议保持一致，并采用 ASR 衡量各组件对生成成功率的贡献。
We evaluate two ablations on the same initial scenes. \textit{w/o Feedback} retains initial robot placement selection and task plan but disables revision based on execution feedback. \textit{w/o Placement Validation} selects robot placement solely from three-view observations and object world coordinates, without considering kinematic or collision constraints or validating placement feasibility, but retains execution feedback for program revision. All other settings and the evaluation protocol follow those used in the demonstration generation comparison above.

% As shown in Table \ref{tab:ablation},我们发现没有feedback 对所有任务的影响都很大balbala,
As shown in \Cref{tab:ablation}, removing execution
feedback reduces ASR to almost $0\%$ across all task categories,
despite lower token consumption.
Removing placement validation particularly affects
long-horizon fixed-base tasks; smaller drops on mobile
tasks suggest that navigation may partially compensate
for suboptimal initial placement.

% \noindent\textbf{跨本体示范生成。}
% 除上述定量评测外，我们进一步展示单臂、双臂、人形和灵巧手配置下的专家示范。如图~\ref{fig:embodiment_rollouts} 所示，各组关键帧呈现了从初始状态、关键交互到任务完成的执行过程。平台在执行过程中同步记录视觉观测、机器人状态与控制动作，将成功轨迹转化为可用于策略学习的示范数据。这些案例展示了示范生成与采集流程对不同机器人配置的支持。

\begin{table}[t]
\centering
\caption{\textbf{Agent Workflow Ablation.}
We evaluate the effects of execution feedback and robot placement
validation across fixed-base and mobile manipulation tasks.
Token denotes the mean cumulative VLM output tokens per session.}
\label{tab:ablation}

\footnotesize
\setlength{\tabcolsep}{3pt}
\renewcommand{\arraystretch}{0.7}
\resizebox{\linewidth}{!}{
\begin{tabular}{l*{12}{c}}
\toprule
& \multicolumn{3}{c}{\textbf{Fixed-Short}}
& \multicolumn{3}{c}{\textbf{Fixed-Long}}
& \multicolumn{3}{c}{\textbf{Mobile-Short}}
& \multicolumn{3}{c}{\textbf{Mobile-Long}} \\
\cmidrule(lr){2-4}
\cmidrule(lr){5-7}
\cmidrule(lr){8-10}
\cmidrule(lr){11-13}
\textbf{Method}
& \textbf{ASR}$\uparrow$
& \textbf{Top-5}$\uparrow$
& \textbf{Token}$\downarrow$
& \textbf{ASR}$\uparrow$
& \textbf{Top-5}$\uparrow$
& \textbf{Token}$\downarrow$
& \textbf{ASR}$\uparrow$
& \textbf{Top-5}$\uparrow$
& \textbf{Token}$\downarrow$
& \textbf{ASR}$\uparrow$
& \textbf{Top-5}$\uparrow$
& \textbf{Token}$\downarrow$ \\
\midrule

w/o Feedback
& $4.0\%$ & $8.0\%$ & 3721
& $0.0\%$ & $0.0\%$ & 6529
& $0.0\%$ & $0.0\%$ & 2979
& $0.0\%$ & $0.0\%$ & 5448 \\

w/o Placement Validation
& $48.6\%$ & $56.8\%$ & $63819$
& $13.2\%$ & $22.4\%$ & $208316$
& $67.3\%$ & $79.6\%$ & $13294$
& $47.6\%$ & $64.7\%$ & $40898$ \\
\midrule

\textbf{Ours}
& \cellcolor{blue!15}\textbf{69.4}\% & \cellcolor{blue!15}\textbf{75.2}\% & $15003$
& \cellcolor{blue!15}\textbf{71.6}\% & \cellcolor{blue!15}\textbf{79.2}\% & $78183$
& \cellcolor{blue!15}\textbf{70.0}\% & \cellcolor{blue!15}\textbf{86.0}\% & $8921$
& \cellcolor{blue!15}\textbf{54.0}\% & \cellcolor{blue!15}\textbf{65.3}\% & $50187$ \\
\bottomrule
\end{tabular}}
\end{table}

%指标：ASR, Top-5 AS, Token这三个指标（参考robotwin），
%任务类型：并且为静态操纵（五个简单五个长程）、移动操纵（五个简单五个长程)分别提供成功率，这部分有一个大表格就行，和scene smith/embodied gen v2 + task plan做成功率定量+可视化定性对比
%对比实验：1. 消融实验（1. 和我们没有rsi/agent的版本（没有回环、迭代和反馈）对比 2. 消融skill）一个表格
% 定量rollout展示: 展示我们的数据搜集本体支持,rollout专家展示,灵巧手,人形机器人只进行定量展示,体现我们的平台能够支持这两种本体,4个rollout 视频帧平铺, 分别是单臂双臂人形灵巧手,进行画图,正文call一下就行
% @yifei，zepeng，yikai
% 具体实验量：对于每类任务，每类10个任务上评估，每个任务跑10次；
% 1.base line场景+agent规划
% 2.我们自己的场景+ agent规划
% 3.我们自己的场景+ agent规划+消融
% ddl: 9.9

%yifei Embodied Gen V2 不带铰链资产生成，有铰链资产的场景可以直接标注 -

% yifei 

% ddl: 9.9
\begin{figure*}[htbp]
    \centering    
    \includegraphics[width=\linewidth]{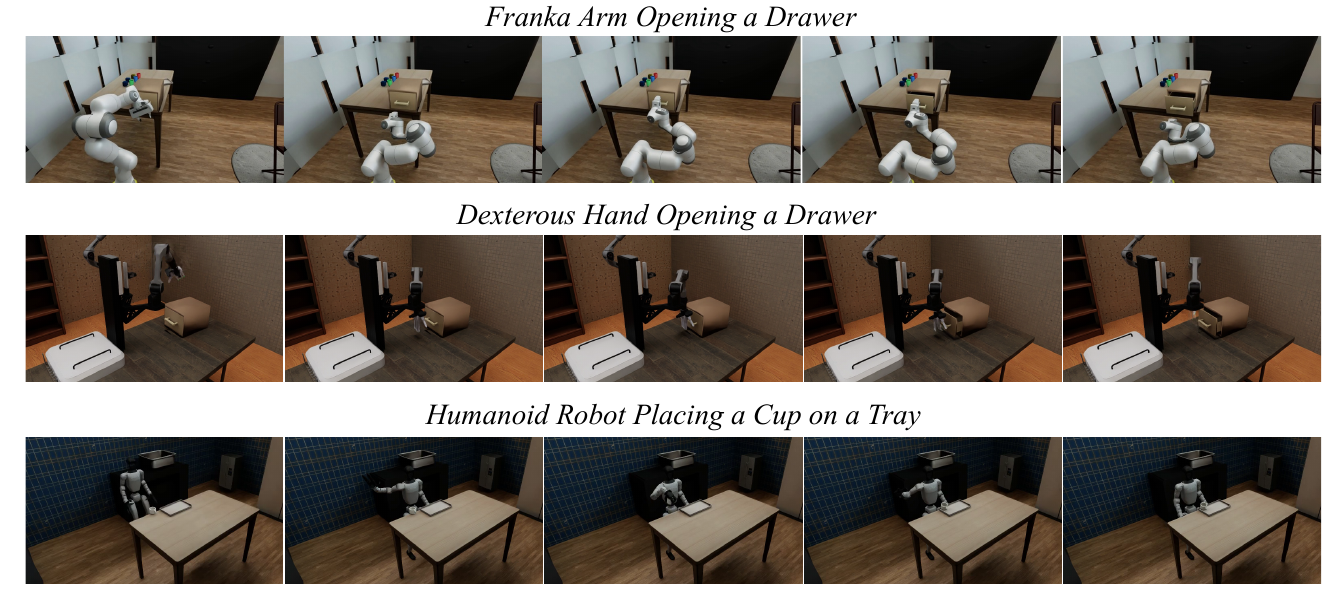}
    \caption{\textbf{Visualization of generated robot demonstrations of diverse embodiments: mobile manipulator with a single arm, mobile manipulator with dexterous hands, and humanoid.} Specifically, the humanoid robot first moves to the table and then manipulates the target.}
    \label{fig:embodiment}
\end{figure*}
\noindent\textbf{Demonstrations across embodiments.}
Beyond the quantitative evaluations above, we present expert demonstrations using single-arm, dual-arm, humanoid, and dexterous-hand configurations. \Cref{fig:embodiment} shows keyframes spanning the initial state, intermediate interactions, and task completion. During execution, EmbodiedSmith records synchronized visual observations, robot states, and control actions, converting successful trajectories into demonstration data for policy learning. 

% %灵巧手机器人成功完成了对抽屉把手的手内抓握并且打开了抽屉。
% The humanoid robot first moves to the table and then manipulates the target.
% %这些demonstration证明了我们的method具备完成human-level dexterous tasks的能力。因此能够为通用机器人基础模型生成不同构型的数据，将机器人基础模型的能力提升至人类水平。
The robot equipped with dexterous hands successfully performs an in-hand grasp of the drawer handle and opens the drawer.
The humanoid robot first moves to the table and then manipulates the target.
These demonstrations show that our method can perform dexterous tasks requiring human-level dexterity. The method can thus generate data across diverse robot embodiments for training general-purpose robot foundation models, advancing their capabilities toward human-level performance.

% 2. 模型训练有关的：展示数据能用来训练模型
% \subsection{Model Pre-training} % 说明我们的数据可以用来进行预训练，规模不用太大
% \subsection{Sim2Real} % 说明我们的数据可以sim2real
\subsection{Generated Data for Policy Training}
\label{sec:4.3}
%机器人实验：搞20小时的后训练数据（对应10个任务，有不同的难度等级，每个任务2小时），然后跑两个model (pi05和fastwam)，划分一个evaluation set（完全不同的空间、背景和物体，但是任务相同）；想得到的结论是：我们以这种方式生成的数据可用，diverse的数据训练出的模型具有泛化性。并且泛化的效果随着数据的多样性程度和数量提升而提升，对应两个scale维度--多样性和数量。具体来说数据可以有20小时是场景相对单一的，10小时是场景和整个的泛化非常多的，方便我们做对比多样性的实验。
%两个模型够吗？是不是要再多加几个模型
%考虑再加上pi0，openvla，openvla oft，lingbotvla
%@yikai
%ddl: 9.9
%我们在书房和厨房两个最具有代表性的家庭场景展开我们的实验和测试。我们一共设计了10个不同的任务，每个任务总时长约2小时（场景和任务描述见图和附录）。针对每个模型提供的预训练点，我们使用我们的embodysmith端到端生成的数据来进行微调训练，并分别对比和验证了：1.随着训练持续进行模型性能和表现的对比，验证我们的数据是确实有效的（5000和10000 step）；2.在与训练数据完全不同场景下（包括背景空间和具体物品）进行推理实验，验证我们的数据训练出来的模型具有一定的泛化性；3.使用我们的数据，即使是混杂不同的任务训练模型，相比于使用单一量的数据，也对任务成功率有帮助，证明我们diverist的数据对模型是有好处的，有预训练的潜力（单一任务ckpt和所有任务ckpt对比，训练步数两者保持一样还是后者是前者的n倍？）
%我们的训练是在a模型，x场景下的5个任务，分别训练5个的单一任务的ckpt和一个使用所有任务的ckpt，然后进行上面三个分别对比实验？
%放一个图表展示一下数据构成
%放一个表展示一下每个模型：1.使用原始预训练权重的成功率 2.使用我们数据训练之后的成功率 3.使用我们数据训练，更换验证场景成功率 
%针对预训练潜力的实验，挑选两个任务，对比仅使用对应任务数据和使用全量数据对比。

We design experiments to verify that the generated robot trajectories can effectively improve the performance and generalization ability of trained models, thereby demonstrating the capability of our engine to provide large-scale training data. 

% \subsubsection{Automated Task Plan and Data Generation}

%我们选取书房和厨房两个目标场景来从零开始进行场景规划，并且分别为其设计了5个合乎场景的任务，覆盖了常见的pick，place，open等原子技能，任务的长度从3个到6个原子技能之间。值得一提的是，我们在确定好任务之后，直接使用Prompt 生成了所有的任务规划，并且能够直接得到运行，完全没有人工调试介入。详细的生成任务的prompt描述见附录。

%放图，2*5小图拼大图
%放表，每个场景每个任务的数据时间，episode数，收集成功率。
\noindent\textbf{Training Data Generation.}
We selected two target scenes, a study room and a kitchen, and planned each scene from scratch. For each scene, we designed five scene-appropriate tasks covering common atomic skills such as pick, place, and open, with task lengths ranging from three to six atomic skills. It is worth noting that, after determining the tasks, we directly used prompts to generate all task plans, which could be executed successfully without any manual debugging or intervention. Detailed task-generation prompts are provided in Appendix~\ref{app:eval}.

% 两批数据分别是在固定场景布局中不开启随机化和在开启随机化不固定桌面场景的条件下收集的，其中前者约20小时，后者约10小时。我们使用两批数据构造了两个训练数据集：20小时单一数据构造为clean数据集，10小时的单一数据和10小时的混合数据构造为mixed数据集。然后我们用这两个数据集在官方提供的原始checkpoint训练了π0.5和fastwam。
% We collect two batches of demonstrations: approximately 20 hours in fixed scene layouts without randomization, and approximately 10 hours with randomization enabled and varying tabletop layouts.
% From these, we construct two training datasets of equal duration.
% The \textbf{clean} dataset contains 20 hours of fixed-layout data, while the \textbf{mixed} dataset combines 10 hours of fixed-layout data with 10 hours of randomized data.
% We then fine-tune models on each dataset, starting from their officially released pretrained checkpoints.
% After verification, when randomization is disabled, the robot trajectories in the data generated directly using the configurations produced by the Task Plan Agent are not completely identical across episodes.
We construct two equal-duration datasets: \textbf{clean}, comprising 20 hours of fixed-layout demonstrations without randomization, and \textbf{mixed}, combining 10 hours of these demonstrations with 10 hours of randomized demonstrations featuring varying tabletop layouts.

Our randomization mechanism organizes variations at three levels: asset composition, spatial layout, and observation and initial conditions. The enabled dimensions and sampling ranges are configured per task, with layout sampling constrained by object clearance, support regions, and reserved manipulation space. Appendix~\ref{app: Task Randomization} details the hierarchical sampling procedure and execution-feedback-driven resampling used during collection.

%放一个表格

% \subsubsection{Evaluation of Diversity Data Scaling Law}

%\textbf{Experimental Setup.}为了评估我们的数据效果，我们选取了VLA和WAM中最具代表性的两个模型：π0.5和fastwam，分别在书房（t1-t5)和厨房（t6-t10）收集的的数据上进行了训练，并且记录在两种不同泛化程度下的场景中的评估表现。每个场景数据集的都由5个不同的任务组成，每个任务的平均时长，原子动作数和详细描述见附录。两批数据分别是在固定场景布局中不开启随机化和在开启随机化不固定桌面场景的条件下收集的，其中前者约20小时，后者约10小时。我们使用两批数据构造了两个训练数据集：20小时单一数据构造为clean数据集，10小时的单一数据和10小时的混合数据构造为mixed数据集。然后我们用这两个数据集在官方提供的原始check point训练了π0.5和fastwam。此外，我们构造了两个测试场景，其中simple环境和收集clean数据时候的场景布局保持一致，没有开启任何随机化，random环境使用了随机seed来进行环境随机泛化。random场景中的每一个seed都被专家执行器验证过可以成功执行任务，并在验证后固定下来，在评估的时候会使用这批被验证过的seed以保证可执行性和测试的公平性。

%\textbf{Evaluating Result.}%实验结果如表。整体表现上π0.5优于fastwam。使用clean数据训练的模型在simple中表现良好，但是random环境中评估成功率会暴跌，某种程度上说明了使用单一数据训练的模型可能是学会了图像和轨迹的对应而不是真正学会了操作。使用mixed数据训练的模型在simple环境中的各个任务上的表现相比clean版本成功率各有起伏，整体上成功率稍微损失；在random环境上相比于clean版本则成功率大幅度提高。这证明了我们的数据提高多样性之后能显著的帮助模型提高泛化的效果，可以作为大规模预训练数据的来源。

\begin{table*}[!t]
\centering

\caption{\textbf{Task success rates (\%) under simple and randomized evaluation settings.}
We compare $\pi_{0.5}$ and Fast-WAM fine-tuned on the \textit{clean} or \textit{mixed} dataset.}
\label{tab:task_success_rates}

\small
\setlength{\tabcolsep}{3pt}
\renewcommand{\arraystretch}{1.0}

\begin{adjustbox}{width=\linewidth}
\begin{tabular}{@{}lcccccccccccc@{}}
\toprule

\multirow[c]{2}{*}[-0.3ex]{\textbf{Task}}
& \multicolumn{3}{c}{$\pi$0.5 (Clean)}
& \multicolumn{3}{c}{$\pi$0.5 (Mixed)}
& \multicolumn{3}{c}{FastWAM (Clean)}
& \multicolumn{3}{c}{FastWAM (Mixed)} \\

\cmidrule(lr){2-4}
\cmidrule(lr){5-7}
\cmidrule(lr){8-10}
\cmidrule(lr){11-13}

& \textbf{Simple} & \textbf{Random} & \textbf{Average}
& \textbf{Simple} & \textbf{Random} & \textbf{Average}
& \textbf{Simple} & \textbf{Random} & \textbf{Average}
& \textbf{Simple} & \textbf{Random} & \textbf{Average} \\

\midrule

Spatially Guided Placement
& 81\% & 5\% & 43.0\% & 64\% & 47\% & 55.5\% & 34\% & 1\% & 17.5\% & 25\% & 12\% & 18.5\% \\

Cuboid Placement
& 88\% & 11\% & 49.5\% & 65\% & 59\% & 62.0\% & 41\% & 2\% & 21.5\% & 28\% & 18\% & 23.0\% \\

Block Stacking
& 30\% & 4\% & 17.0\% & 39\% & 22\% & 30.5\% & 11\% & 0\% & 5.5\% & 15\% & 5\% & 10.0\% \\

Unstack and Place
& 69\% & 7\% & 38.0\% & 31\% & 18\% & 24.5\% & 26\% & 0\% & 13.0\% & 12\% & 4\% & 8.0\% \\

Drawer Opening
& 35\% & 5\% & 20.0\% & 46\% & 30\% & 38.0\% & 13\% & 1\% & 7.0\% & 18\% & 9\% & 13.5\% \\

Sequential Object Placement
& 56\% & 4\% & 30.0\% & 43\% & 37\% & 40.0\% & 21\% & 0\% & 10.5\% & 16\% & 7\% & 11.5\% \\

Bottle Righting
& 64\% & 8\% & 36.0\% & 54\% & 45\% & 49.5\% & 27\% & 1\% & 14.0\% & 22\% & 10\% & 16.0\% \\

Bowl Nesting
& 68\% & 13\% & 40.5\% & 64\% & 56\% & 60.0\% & 29\% & 2\% & 15.5\% & 26\% & 15\% & 20.5\% \\

Microwave Door Closing
& 78\% & 12\% & 45.0\% & 67\% & 61\% & 64.0\% & 37\% & 3\% & 20.0\% & 29\% & 18\% & 23.5\% \\

Sponge Wiping
& 43\% & 3\% & 23.0\% & 35\% & 27\% & 31.0\% & 16\% & 0\% & 8.0\% & 13\% & 4\% & 8.5\% \\

\midrule

\textit{Average}
& 61.2\% & 7.2\% & 34.2\%
& 50.8\% & 40.2\% & 45.5\%
& 25.5\% & 1.0\% & 13.3\%
& 20.4\% & 10.2\% & 15.3\% \\

\bottomrule
\end{tabular}
\end{adjustbox}
\end{table*}

% \textbf{Experimental Setup.}为了评估我们的数据效果，我们选取了VLA和WAM中最具代表性的两个模型：π0.5和fastwam，分别在书房（t1-t5)和厨房（t6-t10）收集的数据上进行了训练，并且记录在两种不同泛化程度下的场景中的评估表现。每个场景数据集都由5个不同的任务组成。任务设计、数据收集以及训练数据的详细信息见附录。
\textbf{Experimental Setup.}
We select $\pi_{0.5}$ and FastWAM as representative VLA and WAM models, respectively, train them on data collected in a study and a kitchen, and evaluate their performance in two settings with different generalization requirements.
Each scene dataset contains five tasks, with more details provided in the appendix.
% 此外，我们构造了两个测试场景，其中simple环境和收集clean数据时候的场景布局保持一致，没有开启任何随机化，random环境使用了随机seed来进行环境随机泛化。random场景中的每一个seed都被专家执行器验证过可以成功执行任务，并在验证后固定下来，在评估的时候会使用这批被验证过的seed以保证可执行性和测试的公平性。
We define two evaluation settings.
The \textbf{simple} setting uses the same scene layouts as those used to collect the \textit{clean} dataset, with no randomization.
The \textbf{random} setting uses scene randomization controlled by random seeds.
% Each seed used in the \textit{random} setting is validated through successful task execution by the expert executor and then held fixed.
% The same validated seed set is reused across evaluations to ensure task feasibility and a fair comparison.

% \textbf{Evaluating Result.}实验结果如表。整体表现上π0.5优于fastwam。使用clean数据训练的模型在simple中表现良好，但是random环境中评估成功率会暴跌，某种程度上说明了使用单一数据训练的模型可能是学会了图像和轨迹的对应而不是真正学会了操作。
% 将 tab:results 替换为实际结果表的标签。
\textbf{Evaluation Results.}
\Cref{tab:task_success_rates} shows models trained on the \textit{clean} dataset perform well in the \textit{simple} setting, but their success rates drop sharply in the \textit{random} setting.
This suggests that training on fixed-layout data may encourage reliance on scene-specific image--trajectory associations rather than manipulation strategies that generalize across layouts.
% 使用mixed数据训练的模型在simple环境中的各个任务上的表现相比clean版本成功率各有起伏，整体上成功率稍微损失；在random环境上相比于clean版本则成功率大幅度提高。这证明了我们的数据提高多样性之后能显著地帮助模型提高泛化的效果，可以作为大规模预训练数据的来源。
Models trained on the \textit{mixed} dataset achieve higher average success rates than their \textit{clean}-trained counterparts, with particularly pronounced gains in the \textit{random} setting.
These results show that increasing the diversity of our generated data can substantially improve generalization to randomized scenes, supporting its potential use as a source of large-scale training data.

Representative evaluation rollouts are provided in Appendix~\ref{app:eval}.

\section{Conclusion}

We presented \method, a framework for scalable embodied data generation through RSI. By integrating asset, scene, and task generation, \method~supports diverse tasks, complex embodiments, and physical interactions. 
Its coupled environment and task generation loop enables mutual refinement, improving overall generation success rates. 
Experiments demonstrate the quality, diversity, and generation efficiency of the resulting environments and demonstrations. These results suggest that jointly refining simulation environments and executable task plans offers a promising path toward scalable training data and evaluation infrastructure for generalist robot models.

\bibliography{paper}
\bibliographystyle{assets/plainnat}

\clearpage
\beginappendix
\renewcommand{\thefigure}{S\arabic{figure}}
\renewcommand{\thetable}{S\arabic{table}}
\setcounter{figure}{0}
\setcounter{table}{0}

\section{Abstract}

In this appendix, we provide additional information:

\begin{itemize}
    \item Section~B details scene generation, including the asset library, and scene-evaluation metrics.

    \item Section~C provides details of robotic demonstration generation, including task definitions and success criteria, robot skills, agent tools, evaluation protocols, ablation settings, robot configurations, and demonstrations across embodiments.

    \item Section~D describes the hierarchical randomization and feedback-driven data-generation procedure used for policy training and evaluation.

    \item Section~E provides additional robot evaluation details, including task generation prompts, data-collection statistics, training hyperparameters, and representative evaluation rollouts.
\end{itemize}

\section{Details of Environment Generation}
\label{app:4.1details}

\subsection{Used Assets Library}
\label{app:assets}
% 为保留各系统的预期行为，各基线沿用其原生资产来源和生成流程；所有方法接收相同的输入规格，并在支持的情况下使用匹配的随机种子。
To preserve the intended behavior of each system, each baseline retains its native asset source and generation pipeline, while all methods receive identical input specifications and matched random seeds whenever supported. 

% 我们的方法使用自建资产库作为场景组合的资产索引。该库包含可用于仿真的资产，具有标准化的几何表示和语义元数据，使流程能够根据场景语义及后续物理需求检索对象。给定输入规格后，流程逐步选择合适的资产，确定其功能角色与空间关系，并通过所提出的场景生成策略将其实例化。
Our method utilizes a self-constructed asset library as the asset index for scene composition. This library contains simulation-ready assets with standardized geometric representations and semantic metadata, enabling the pipeline to retrieve objects based on both scene semantics and downstream physical requirements. Given an input specification, the pipeline progressively selects suitable assets, determines their functional roles and spatial relationships, and instantiates them using the proposed scene generation strategy. 

\subsection{Evaluation Metrics}
\label{app:metrics}
% 我们使用五项感知指标和四项基于几何的指标评估生成的环境。
We evaluate generated environments using five perceptual metrics and four geometry-based metrics.

\noindent\textbf{Perceptual Metrics.}
% 对于每个场景，GPT-5.5 接收四个渲染视角，并为每项指标给出 1 至 10 分的评分。内容／功能（CF）衡量场景是否包含目标房间类型所需的对象和功能组件。空间布局（SL）评估对象的位置、朝向、间距、功能分区和通行空间。物理合理性（PP）评估视觉上感知到的落地、支撑、对齐情况，以及明显的悬浮或穿插现象。文本一致性（TF）衡量对象类别、数量、属性、材质、空间关系和房间类型与输入规格的一致程度。视觉质量（VQ）评估材质、纹理、光照、清晰度、尺度一致性和渲染伪影。
For each scene, GPT-5.5 \citep{openai_gpt55} receives four rendered views and assigns a score from 1 to 10 for each metric. \textbf{Content / Function (CF)} measures whether the scene contains the objects and functional components required by the intended room type. \textbf{Spatial Layout (SL)} evaluates object placement, orientation, spacing, functional zoning, and circulation space. \textbf{Physical Plausibility (PP)} assesses visually perceived grounding, support, alignment, and obvious floating or interpenetration artifacts. \textbf{Text Fidelity (TF)} measures consistency with the input specification in terms of object categories, quantities, attributes, materials, spatial relations, and room type. \textbf{Visual Quality (VQ)} evaluates materials, textures, lighting, clarity, scale consistency, and rendering artifacts.

\noindent\textbf{Geometry-based Metrics.}
% 碰撞（COL）统计发生实际网格相交的对象数量。可导航性（NAV）是在机器人宽度约束下，最大连续可通行区域占有效地面面积的比例。越界（OOB）统计几何形状超出有效房间或建筑边界的对象数量。开口净空／开放度（OPN）是门窗两侧中保持无遮挡的侧面所占比例。
\textbf{Collision (COL)} is the number of objects involved in actual mesh intersections. \textbf{Navigability (NAV)} is the fraction of valid floor area contained in the largest continuously traversable region under robot-width constraints. \textbf{Out of Bound (OOB)} counts objects whose geometry extends beyond valid room or building boundaries. \textbf{Opening Clearance / Openness (OPN)} is the fraction of door and window sides that remain unobstructed.

% 机器人示范生成细节。
\section{Robotic Demonstration Generation Details}
\label{app:demonstration}

% 任务定义与成功判据。
\subsection{Task Definitions and Success Criteria}
% 主评测包含 16 项任务。
% 任务表将任务分为 Easy 与 Long horizon 两组。
% 场景表汇总实际使用的场景，复用场景只列一次；任务表给出任务指令、场景对应、分组与完成条件。
The main evaluation comprises 16 tasks: ten fixed-base
manipulation tasks and six mobile manipulation tasks.
The task table groups them as Easy and Long horizon.
\Cref{tab:scene_prompts} summarizes the scenes used
in this evaluation, with shared scenes listed once.
\Cref{tab:task_definitions} provides the task instructions,
scene assignments, group labels, and completion conditions.

% 共同成功条件：多阶段任务同时检查必要中间事件与终态；
% 回忆任务还检查任务指定的观察、遮挡或中间操作阶段，再检查最终结果。
% 任务判据在反馈修订过程中保持不变。
Success is determined by predefined task-specific completion
conditions. For multi-stage tasks, a successful episode must
satisfy both the required intermediate-event conditions and
the final-state conditions. For recall tasks, any specified observation,
occlusion, or intermediate manipulation stages must be completed before
the final task outcome is evaluated. The task-completion
conditions remain fixed during feedback-driven refinement.
Task-specific thresholds for joint opening, lifting height,
and pose matching are specified in the corresponding
evaluation configurations.

% 任务智能体将参数化原子技能组织为有向无环图。
% 表中列出仿真执行层支持的机器人技能；技能节点指定执行资源、目标参数，并在适用时指定目标实体。
\subsection{Robot Skills}
The task agent composes parameterized atomic skills into a task DAG.
\Cref{tab:robot_skills} summarizes the robot skills supported by the simulation execution layer.
Each skill node specifies an execution resource and goal parameters,
together with a target entity when applicable.
\begin{table}[t]
\centering
\caption{\textbf{Robot Skills.} We summarize the parameterized atomic skills supported by the simulation execution layer.}
\label{tab:robot_skills}
\footnotesize
\setlength{\tabcolsep}{3pt}
\renewcommand{\arraystretch}{1.15}
\begin{tabular}{@{}p{0.28\linewidth}p{\dimexpr0.72\linewidth-2\tabcolsep\relax}@{}}
\toprule
\textbf{Name} & \textbf{Description} \tabularnewline
\midrule
\texttt{navigate} & Move the mobile base to a working location before local manipulation. \tabularnewline
\texttt{pick} & Execute pre-grasp, approach, gripper closure, and lift for a target object. \tabularnewline
\texttt{manualpick} & Execute a grasp with explicitly adjusted grasp pose parameters. \tabularnewline
\texttt{dexpick} & Execute a grasp using a selected predefined grasp pose. \tabularnewline
\texttt{dynamicpick} & Plan and execute a grasp for a moving target. \tabularnewline
\texttt{place} & Transport a held object to a target, descend, release it, and retreat. \tabularnewline
\texttt{dexplace} & Place a held object within a specified geometric target region. \tabularnewline
\texttt{open} & Plan and execute opening motion for an articulated object such as a drawer or door. \tabularnewline
\texttt{close} & Plan and execute closing motion for an articulated object such as a drawer or door. \tabularnewline
\texttt{artpreplan} & Execute planned approach poses for an articulated-object interaction. \tabularnewline
\texttt{rotate} & Plan and execute rotation of an articulated joint to a target state. \tabularnewline
\texttt{press} & Approach and press a designated button or contact region. \tabularnewline
\texttt{move} & Move an object toward another object along a planar trajectory. \tabularnewline
\texttt{rotate\_obj} & Reorient a grasped object while maintaining its attachment. \tabularnewline
\texttt{approach\_rotate} & Move a held object toward a target while adjusting its orientation. \tabularnewline
\texttt{flip} & Flip a grasped object and release it. \tabularnewline
\texttt{goto\_pose} & Plan arm motion to a specified end-effector pose. \tabularnewline
\texttt{gripper\_action} & Open or close the gripper independently of arm motion. \tabularnewline
\texttt{arm\_motion} & Move the arm to a joint or relative end-effector target, including a home pose. \tabularnewline
\texttt{joint\_ctrl} & Directly control selected robot joints. \tabularnewline
\texttt{observe\_hold} & Hold the robot pose during a specified observation interval. \tabularnewline
\texttt{wait} & Hold the end-effector pose for a specified number of simulation steps. \tabularnewline
\texttt{pour\_water} & Tilt a held container to transfer fluid and check the simulated outcome. \tabularnewline
\bottomrule
\end{tabular}
\end{table}

% 轻量级 harness 提供五个任务级工具。
% 智能体借助这些工具对应场景实体、构建并执行技能计划、根据执行反馈修订失败计划，并提交成功回合以供最终验证。
\subsection{Agent Tools}
The five task-level tools exposed by the harness are summarized in \Cref{tab:agent_tools} in the main text.

% 智能体配置与评估协议。
\subsection{Agent Configuration and Evaluation Protocol}
% 所有规划实验使用相同的模型与推理强度。
% 每项任务在每个生成场景中开展 10 次独立的智能体会话，初始场景状态相同。
% 每个会话生成包含初始站位、技能计划和成功判据的候选。
% 在固定场景评测中，启用反馈修订时可在最多 30 轮预算内调整计划与初始站位；
% 任务成功判据、场景资产和物体布局保持不变。
We use \texttt{gpt-5.6-sol} with \texttt{high} reasoning effort as the task generation agent for all generation experiments. For each task and generated scene, we run 10 independent agent sessions from the same initial scene state. Each session attempts to produce a candidate comprising an initial robot placement, a skill plan, and executable task-success criteria. In the fixed-scene evaluation, feedback-driven refinement may revise the plan and initial robot placement within a maximum budget of 30 rounds. The task-success criteria, scene assets, and object layout remain unchanged.

% 最终计划与站位在留出评估前冻结，并在 10 个共享留出种子上评估。
% 留出种子不参与候选生成或修订，每个评估回合开始前重置场景。
% 评估期间不进行进一步的智能体修订或场景配置编辑。
The final plan and placement are frozen before evaluation under 10 shared held-out simulation seeds that are excluded from generation and refinement. The scene is reset before each evaluation trial, and the frozen candidate is executed without further agent-driven refinement or scene-configuration edits.

% 不同种子控制任务配置中启用的物体与机器人初始位姿采样，也可能改变启用随机化的光照等观测条件；因此，同一任务计划在不同试验中可能产生不同的导航和操作轨迹。
Across seeds, task-enabled randomization can vary initial object and robot poses and, when configured, observation conditions such as lighting. These variations can produce different navigation and manipulation trajectories for the same frozen task plan.

% 评估指标与消融设置。
\subsection{Evaluation Metrics and Ablation Settings}
% 每个候选的成功率为其在 10 个留出种子上的成功比例。
% ASR 先对每项任务的全部候选求平均，再跨任务求平均；Top-5 ASR 对每项任务中成功率最高的五个候选求平均，再跨任务求平均。
% Token 统计每次智能体会话累计的 VLM 输出 token；未生成可执行计划的会话计入 token 开销，成功率记为零。
For each candidate, we compute the fraction of successful trials over the ten held-out seeds. ASR averages candidate success rates within each task and then across tasks. Top-5 ASR averages the five highest candidate success rates within each task and then across tasks. Token is the mean cumulative VLM output tokens per agent session. Sessions that do not produce an executable plan contribute zero to the success rates and remain in the token accounting.

% 两项智能体消融使用相同的初始场景与留出评估协议。
% w/o Feedback 关闭根据执行反馈修改计划的环节；w/o Placement Validation 不使用运动学与碰撞约束验证初始站位，但保留执行反馈修订与正常导航。
The agent ablations use the same initial scenes and held-out evaluation protocol. \textit{w/o Feedback} disables plan revision based on execution feedback. \textit{w/o Placement Validation} selects the initial robot placement from three rendered views and object world coordinates without kinematic or collision-based placement validation; execution-feedback-driven revision and normal navigation remain enabled. Both ablations retain the original task-success criteria.

% 机器人配置与示范录制。
\subsection{Robot Configurations and Demonstration Recording}

% 定量评测按任务采用相应的机器人配置。固定底座任务在执行期间保持底座位姿不变；移动操作任务允许底座导航至操作位置。
% 机械臂轨迹由 CuRobo 基于场景碰撞几何规划；对于同一任务，不同场景生成方法使用相同的机器人配置与底层控制接口。
The quantitative benchmark uses the robot configuration specified for each task. Fixed-base tasks keep the base pose fixed during execution, whereas mobile manipulation tasks allow base navigation to a suitable working location. Arm trajectories are planned with CuRobo using scene collision geometry. Within each task, all scene-generation methods use the same robot configuration and low-level control interfaces.

% 主定量评测回合在 Isaac Sim 6.0.1 中运行，
% 物理与渲染步长均为 1/30 秒。
% 执行期间记录时间对齐的观测、状态、动作及阶段标注。
% 任务成功日志与技能执行结果用于核查必要中间事件和终态条件。
% 仅通过任务成功判据与轨迹完整性检查的回合纳入示范数据集；
% 失败回合保留用于诊断，但不纳入示范数据集。
All quantitative-benchmark episodes are simulated in \textit{Isaac Sim 6.0.1}, with both the physics and rendering time steps set to $30 Hz$. During execution, we record temporally aligned visual observations, robot proprioceptive states, object states, and control actions, together with execution-stage annotations. Task-success logs and skill-execution outcomes are used to verify required intermediate events and final-state conditions. Only episodes that satisfy the task-success criteria and pass trajectory-integrity checks are included in the demonstration dataset, while failed episodes are retained for diagnosis but excluded from the dataset.

% The main preamble must load longtable and booktabs.
\setlength{\LTcapwidth}{\textwidth}
\begin{longtable}{@{}>{\scriptsize\raggedright\arraybackslash}p{0.23\linewidth}>{\scriptsize\raggedright\arraybackslash}p{\dimexpr0.77\linewidth-2\tabcolsep\relax}@{}}
\caption{\textbf{Scene Prompts.} We list the full scene generation prompts used in the demonstration-generation evaluation. Prompts shared by multiple tasks appear once.}\label{tab:scene_prompts}\\
\toprule
\textbf{Scene} & \textbf{Prompt} \tabularnewline
\midrule
\endfirsthead
\toprule
\textbf{Scene} & \textbf{Prompt} \tabularnewline
\midrule
\endhead
Kitchen with Cups & A modern minimalist rectangular kitchen is arranged in a top-down layout. The floor is covered with large light grayish-white square tiles, and the center of the room remains largely open to provide a wide, unobstructed area for movement and operation. Along the left wall stands a large silver-gray double-door refrigerator, while a smaller dark gray storage cabinet or mini fridge is placed in the lower-left corner. On the upper-left wall, there are white wall cabinets as well as a set of light wood-paneled cabinets. Along the upper wall, a continuous light gray kitchen countertop is installed, with a black glass induction cooktop embedded slightly to the right of center. The cooktop features multiple circular heating zones. On the left side of the countertop, a small number of utensils and a blue compartment organizer are placed. On the upper-right wall, there is a gray open wall shelf divided into multiple storage compartments. In the lower-right area, a white freestanding kitchen appliance or storage cabinet is placed, with a front panel featuring a round knob, several buttons, and a black control module. At the bottom center of the room is a white sink cabinet with a rectangular sink embedded in the countertop, and cabinet doors and drawers below. All large furniture and appliances are arranged along the edges of the room so as not to occupy the center, creating a spacious pathway suitable for robot navigation and manipulation. The overall color palette consists of white, light gray, dark gray, and light wood tones, with soft and even indoor lighting.  In the middle of the room, a tabletop workspace contains three cups, three cubes, and one tray. The cups are colored red, blue, and green, while the cubes are colored yellow, white, and black. Each cup is positioned directly in front of its assigned cube, forming the initial cup-to-cube spatial relations: the red cup is directly in front of the yellow cube, the blue cup is directly in front of the white cube, and the green cup is directly in front of the black cube. The tray is placed in front of the cups. The drawers are positioned near the edges of the workspace and are spatially separated from the cup-cube pairs. \tabularnewline
\midrule
Study with Drawer & Create a modern minimalist rectangular study room or workspace in a top-down layout. The floor is covered with light natural wood flooring, with the wood grain running continuously along the long side of the room. Along the top wall, place a large white work desk with a wide tabletop and slightly rounded edges. On the desk, arrange a small number of office supplies and electronic devices in a neat but slightly random manner, including a dark gray rectangular device, a small white device, a slender red stationery item, a round cup or mug, a gray-black upright organizer, and a brownish-red cup, with appropriate spacing between all items.Along the left wall, place a light beige open bookshelf or storage cabinet with horizontal shelves. In the center-left area of the room, place a low dark reddish-brown rectangular storage box or cover-like object with rounded corners, a top surface featuring dark texture and visible scratches, and a light gold or light wood decorative band running across the middle. In the center-right area of the room, place a large light gray circular rug with a darker gray border and a slightly mottled texture.Maintain sufficient spacing between the work desk, bookshelf, rectangular storage box, and circular rug, so that the center of the room forms a continuous, unobstructed open area suitable for robot navigation and manipulation. Use soft, even indoor lighting, and adopt a color palette dominated by light wood, white, beige, gray, and dark brown tones.A tabletop workspace contains 3 cups, 3 cubes, 1 tray. The cups are red, blue, and green, and the cubes are yellow, white, and black. Each cup is positioned directly in front of its assigned cube, forming the initial cup-to-cube spatial relations: the red cup is directly in front of the yellow cube, the blue cup is directly in front of the white cube, and the green cup is directly in front of the black cube. The tray is positioned in front of the cups. The drawers are positioned near the edges of the workspace and are spatially separated from the cup-cube pairs. \tabularnewline
\midrule
Living Room with Basket & Create a modern minimalist rectangular living room in a top-down layout. Use light natural wood flooring with a consistent wood-grain direction. Arrange most large furniture along the edges of the room and preserve a large, continuous, unobstructed open area in the center for robot navigation and manipulation.Place a light gray three-seat sofa along the bottom side of the room. The sofa should have three clearly separated seat cushions and a simple modern design.Place a rectangular light-wood table slightly left of the center of the room. Use the table as the main tabletop manipulation workspace. Scatter several small everyday objects across the tabletop with clear spacing between them, including a blue rectangular package or book, several small circular objects, a yellow handle-shaped object, a small frame-like object, and a few small red, white, and dark-colored items. Keep the tabletop objects sparse and clearly separated so that each object is individually accessible for robotic manipulation.Along the left wall, place a long and narrow light-wood storage cabinet or sideboard. Place a few small dark decorative objects or containers on top of it. In the lower-left area, place a low dark-brown rectangular storage box or small piece of furniture with a lighter vertical decorative strip.Add additional dark or brown wooden furniture along the upper and right edges of the room, while keeping these objects away from the central navigation area.Use a warm modern color palette dominated by light wood, white, light gray, dark brown, and small black accents. Use soft and even indoor lighting. Keep the overall scene clean, spacious, and suitable for robotics simulation, tabletop manipulation, object rearrangement, and indoor navigation tasks.In the middle of the room, there is a tabletop workspace containing a wooden basket with a handle and a shallow tray. \tabularnewline
\midrule
Bedroom with Button & Create a modern minimalist rectangular bedroom in a top-down layout. Cover the floor with light natural wood flooring, with the wood grain running continuously along the long side of the room. Keep the central area of the room largely open and unobstructed for robot navigation and manipulation.  Place a large dark gray double bed in the upper-left area of the room. The bed should have a rectangular shape with slightly rounded edges and a dark gray to charcoal-colored mattress and frame.  Along the right wall, place a large brown wooden three-door wardrobe. Each wardrobe door should have recessed rectangular panel details and a long curved white or silver handle. Use a light gray frame around the wardrobe.  At the bottom center of the room, place a low light-wood bench, storage cabinet, or bed-end furniture piece. Place a dark gray or black low cabinet in the lower-left area, and a dark brown wooden storage cabinet or bedside cabinet in the lower-right area.  In the upper-right area, place a small orange-brown square ottoman or padded stool. Add a small dark wooden cabinet or furniture piece nearby along the upper wall.  Arrange all large furniture close to the room boundaries and preserve a wide, continuous open area in the center. Use a warm modern color palette dominated by light wood, dark gray, brown, beige, and small orange-brown accents. Use soft and even indoor lighting. Ensure the scene is clean, spacious, and suitable for robotics simulation, indoor navigation, object retrieval, and wardrobe interaction tasks.In the middle of the room, there is a table with a blue circular button placed on top of it. \tabularnewline
\midrule
Kitchen Counter with Cabinet & The scene contains a long kitchen-style counter measuring approximately 2.0 m in length and 0.65 m in depth. A microwave is placed on one end of the counter at a height reachable by the robot. The microwave has a single graspable hinged door that can be opened and closed. A food container is placed on the countertop near the microwave. At the opposite end of the counter, place a small storage cabinet with a single graspable hinged door. Two graspable storage boxes are placed on the countertop near the cabinet. A storage basket is positioned on the counter a short distance away from the cabinet. The microwave area and cabinet area should be sufficiently separated so that the robot must reposition its mobile base when moving between them. A continuous unobstructed navigation corridor must remain along the entire front side of the counter. \tabularnewline
\midrule
Bedroom with Pose References & Create a modern minimalist rectangular bedroom in a top-down layout. Cover the floor with light natural wood flooring, with the wood grain running continuously along the long side of the room. Keep the central area of the room largely open and unobstructed for robot navigation and manipulation. \par Place a large dark gray double bed in the upper-left area of the room. The bed should have a rectangular shape with slightly rounded edges and a dark gray to charcoal-colored mattress and frame. \par Along the right wall, place a large brown wooden three-door wardrobe. Each wardrobe door should have recessed rectangular panel details and a long curved white or silver handle. Use a light gray frame around the wardrobe. \par At the bottom center of the room, place a low light-wood bench, storage cabinet, or bed-end furniture piece. Place a dark gray or black low cabinet in the lower-left area, and a dark brown wooden storage cabinet or bedside cabinet in the lower-right area. \par In the upper-right area, place a small orange-brown square ottoman or padded stool. Add a small dark wooden cabinet or furniture piece nearby along the upper wall. \par Arrange all large furniture close to the room boundaries and preserve a wide, continuous open area in the center. Use a warm modern color palette dominated by light wood, dark gray, brown, beige, and small orange-brown accents. Use soft and even indoor lighting. Ensure the scene is clean, spacious, and suitable for robotics simulation, indoor navigation, object retrieval, and wardrobe interaction tasks. \par In the middle of the room, there is a table. On the tabletop, three reference objects --- a red cube, a yellow cube, and a blue cylinder --- are arranged in a row as the reference region, each at a distinct position and orientation. An opaque rectangular memory cover with a graspable handle on its top surface is placed beside the reference row; it is large enough to be moved over the reference region and lifted vertically to reveal the objects underneath. On the other side of the table, an empty tray serves as the work region, and the three matching work objects --- identical copies of the red cube, the yellow cube, and the blue cylinder --- are placed next to the tray. \tabularnewline
\midrule
Cup Pairing Workspace & A tabletop workspace contains 3 cups, 3 cubes, one tray, a book, and a notebook. The cups have colors red, blue, and green. The cubes have colors yellow, white, and black. Each cup is positioned directly in front of its assigned cube, forming the initial cup-to-cube spatial relation: the red cup is directly in front of the yellow cube, the blue cup is directly in front of the white cube, and the green cup is directly in front of the black cube. The tray is positioned in front of the cups. The drawer are positioned near the edges of the workspace and are spatially separated from the cup-cube pairs. \tabularnewline
\midrule
Cube Counting Workspace & A tabletop workspace contains 3 cups, one tray, a book, a notebook, and multiple cubes. The cups have colors red, blue, and green. The three cup-pair cubes have colors yellow, white, and black. Each cup is positioned directly in front of its assigned cube, forming the initial cup-to-cube spatial relation: the red cup is directly in front of the yellow cube, the blue cup is directly in front of the white cube, and the green cup is directly in front of the black cube. The tray is positioned in front of the cups, and it contains three cubes: two red cubes and one blue cube. An opaque rectangular cover with a graspable top handle is placed beside the tray, large enough to be moved over the tray and hide the cubes inside. A basket is placed at one corner of the workspace, and three matching spare cubes --- two red and one blue --- are placed next to the basket. The drawer are positioned near the edges of the workspace and are spatially separated from the cup-cube pairs. \tabularnewline
\midrule
Block Order Workspace & The scene contains a large rectangular table measuring approximately 2.0 m in width, 1.0 m in depth, and 1.18 m in height.All designated entry approach zones and circulation clearances must remain completely unobstructed. In particular, if a western-side entry clearance is reserved for the door, this area must remain entirely free of cabinets, carts, lamps, storage units, or any other fixtures. Optional storage fixtures may only be placed within the designated fixture zones, outside both the door-swing envelope and the continuous primary circulation path.On the tabletop, place an opaque brown rectangular memory cover measuring approximately 0.30 m long, 0.10 m wide, and 0.10 m high. The cover should function as a bottom-open shell, allowing it to be lifted vertically to reveal objects underneath.A graspable handle is mounted at the center of the top surface of the cover. The handle measures approximately 0.10 m long, 0.03 m wide, and 0.10 m high, and should be modeled as part of, or rigidly attached to, the cover. The memory cover should be represented as a graspable RigidObject.Place eight blocks on the tabletop. The blocks are divided into four color groups, with two blocks of the same color in each group. Four of the blocks --- one of each color --- are arranged into a rear row as the reference blocks, and the other four blocks --- one of each color, matching the reference set --- are placed in a tray at the front edge of the table as the work set. In front of the rear row, four equally spaced empty slots form the front row.Position the rectangular memory cover directly behind the rear row of blocks. Its long axis should be parallel to the direction of the block arrangement, so that the cover is aligned consistently with the two-row block layout. \tabularnewline
\midrule
Button Sequence Workspace & The scene contains a large rectangular table measuring approximately 2.0 m in width, 1.0 m in depth, and 1.18 m in height.All designated entry approach zones and circulation clearances must remain completely unobstructed. In particular, if a western-side entry clearance is reserved for the door, this area must remain entirely free of cabinets, carts, lamps, storage units, or any other fixtures. Optional storage fixtures may only be placed within the designated fixture zones, outside both the door-swing envelope and the continuous primary circulation path.On the tabletop, three circular push buttons --- red, yellow, and blue --- are mounted in a row near the center of the table. A single green cube is placed directly next to the red button. A tray is placed at the front edge of the table. \tabularnewline
\midrule
Mobile Cup-Cube Workspace & A tabletop workspace contains 3 cups, 3 cubes, one tray, a book, and a notebook. The cups have colors red, blue, and green. The cubes have colors yellow, white, and black. Each cup is positioned directly in front of its assigned cube, forming the initial cup-to-cube spatial relation: the red cup is directly in front of the yellow cube, the blue cup is directly in front of the white cube, and the green cup is directly in front of the black cube. The three cup-cube pairs are distributed along the long side of the workspace with sufficient spacing between them. The tray is positioned in front of the cups. The drawer is positioned near one edge of the workspace and is spatially separated from the cup-cube pairs. A continuous unobstructed navigation corridor is maintained along the front side of the workspace, allowing a mobile robot to reposition between the drawer, cup-cube pairs, and tray. \tabularnewline
\midrule
Mobile Block Memory Table & The scene contains a large rectangular table measuring approximately 2.0 m in width, 1.0 m in depth, and 1.18 m in height. All designated entry approach zones and circulation clearances must remain completely unobstructed. In particular, if a western-side entry clearance is reserved for the door, this area must remain entirely free of cabinets, carts, lamps, storage units, or any other fixtures. Optional storage fixtures may only be placed within the designated fixture zones, outside both the door-swing envelope and the continuous primary circulation path. On the tabletop, place an opaque brown rectangular memory cover measuring approximately 0.30 m long, 0.10 m wide, and 0.10 m high. The cover should function as a bottom-open shell, allowing it to be lifted vertically to reveal objects underneath. A graspable handle is mounted at the center of the top surface of the cover. The handle measures approximately 0.10 m long, 0.03 m wide, and 0.10 m high, and should be modeled as part of, or rigidly attached to, the cover. The memory cover should be represented as a graspable RigidObject. Place eight blocks on the tabletop. The blocks are divided into four color groups, with two blocks of the same color in each group. Four blocks form a reference set and four blocks form a matching work set. Place the reference set toward one end of the table and the matching work set toward the opposite end, while preserving sufficient free tabletop space for rearrangement. Position the rectangular memory cover directly behind the reference blocks. Maintain a continuous unobstructed navigation corridor along the front side of the table so that the mobile robot can move between the reference and work areas. \tabularnewline
\bottomrule
\end{longtable}

% Required package in the main preamble: \usepackage{booktabs}.
% Standard tabular takes ONE argument: the column specification below.
% Uses the ICLR single-column text width without scaling the table.
\begin{table}[t]
    \centering
    \caption{\textbf{Task Prompts and Scene Assignments.} We list the task instruction, generated scene, and success condition for ten fixed-base and six mobile manipulation tasks, grouped into easy and long-horizon tasks.}
    \label{tab:task_definitions}

    \begingroup
    \scriptsize
    \setlength{\tabcolsep}{4pt}
    \renewcommand{\arraystretch}{1.05}
    \begin{tabular}{@{}%
        p{0.18\linewidth}%
        p{\dimexpr0.255\linewidth-2\tabcolsep\relax}%
        p{\dimexpr0.2825\linewidth-2\tabcolsep\relax}%
        p{\dimexpr0.2825\linewidth-2\tabcolsep\relax}%
        @{}}
        \toprule
        \raggedright \textbf{Task} & \raggedright \textbf{Scene} & \raggedright \textbf{Task prompt} & \raggedright \textbf{Completion condition} \tabularnewline
        \midrule

        \multicolumn{4}{@{}l}{\textbf{Fixed-base manipulation}\enspace\textit{/ Easy}} \\
        \addlinespace[2pt]
        \raggedright Open Drawer
            & \raggedright Study with Drawer
            & \raggedright Pull the drawer fully open.
            & \raggedright The drawer reaches its full-opening threshold. \tabularnewline
        \raggedright Cup to Tray
            & \raggedright Kitchen with Cups
            & \raggedright Pick the green cup and place it on the tray.
            & \raggedright The correct cup is released in the valid tray region and supported. \tabularnewline
        \raggedright Click Button
            & \raggedright Bedroom with Button
            & \raggedright Press the blue circular button.
            & \raggedright Robot interaction activates the correct button. \tabularnewline
        \raggedright Basket to Tray
            & \raggedright Living Room with Basket
            & \raggedright Pick up the basket and place it on the tray.
            & \raggedright The basket is released in the valid tray region and supported. \tabularnewline
        \raggedright Open Cabinet
            & \raggedright Kitchen Counter with Cabinet
            & \raggedright Open the cabinet door.
            & \raggedright The cabinet door reaches its hinge-opening threshold. \tabularnewline

        \addlinespace[4pt]
        \multicolumn{4}{@{}l}{\textbf{Fixed-base manipulation}\enspace\textit{/ Long horizon}} \\
        \addlinespace[2pt]
        \raggedright Cup Pairing Recall
            & \raggedright Cup Pairing Workspace
            & \raggedright Remember cup--block pairs; move all three cups to the tray, then restore them.
            & \raggedright All cups visit the tray; original pairings and relative positions are restored. \tabularnewline
        \raggedright Block Order Recall
            & \raggedright Block Order Workspace
            & \raggedright Observe and cover the reference row; reconstruct its order with matching work blocks.
            & \raggedright The covered reference's left-to-right order is reproduced in the work region. \tabularnewline
        \raggedright Cube Count Recall
            & \raggedright Cube Counting Workspace
            & \raggedright Count the tray's cubes, cover the tray, and transfer matching cubes to the basket.
            & \raggedright The basket receives exactly the remembered number of matching cubes. \tabularnewline
        \raggedright Button History Recall
            & \raggedright Button Sequence Workspace
            & \raggedright Press the cube-adjacent button, then another; move the cube to the tray; press the unused button.
            & \raggedright The ordered actions are completed; the last button differs from both earlier buttons. \tabularnewline
        \raggedright Object Gesture Recall
            & \raggedright Bedroom with Pose References
            & \raggedright Observe reference poses, cover the reference region, and arrange matching work objects.
            & \raggedright Corresponding positions and orientations match within the pose tolerances. \tabularnewline

        \midrule
        \multicolumn{4}{@{}l}{\textbf{Mobile manipulation}\enspace\textit{/ Easy}} \\
        \addlinespace[2pt]
        \raggedright Open Drawer+Nav
            & \raggedright Mobile Cup-Cube Workspace
            & \raggedright Navigate to the drawer and pull it fully open.
            & \raggedright Required navigation completes and the drawer reaches its full-opening threshold. \tabularnewline
        \raggedright Cup to Tray+Nav
            & \raggedright Mobile Cup-Cube Workspace
            & \raggedright Navigate to the red cup and transfer it to the tray.
            & \raggedright Required navigation completes; the red cup is released in the valid tray region. \tabularnewline
        \raggedright Lift Cover+Nav
            & \raggedright Mobile Block Memory Table
            & \raggedright Navigate to the brown cover, grasp its top handle, and lift vertically.
            & \raggedright The cover remains held and rises by at least the lift threshold. \tabularnewline
        \addlinespace[4pt]
        \multicolumn{4}{@{}l}{\textbf{Mobile manipulation}\enspace\textit{/ Long horizon}} \\
        \addlinespace[2pt]
        \raggedright Book Stow \& Deliver
            & \raggedright Mobile Cup-Cube Workspace
            & \raggedright Move to the drawer, pull it fully open, place the notebook inside, and close it. Then move to the tray and place the book on it.
            & \raggedright The notebook remains in the closed drawer and the book is released on the tray after the required navigation. \tabularnewline
        \raggedright Cup Pairing Nav
            & \raggedright Mobile Cup-Cube Workspace
            & \raggedright Recall cup--block pairs; navigate to move all cups to the tray, then restore them.
            & \raggedright All cups visit the tray; navigation completes and the original pairings are restored. \tabularnewline
        \raggedright Block Order Nav
            & \raggedright Mobile Block Memory Table
            & \raggedright Observe the reference order; navigate to the opposite work area and arrange matching blocks.
            & \raggedright Matching blocks reproduce the reference's left-to-right order in the work area. \tabularnewline
        \bottomrule
    \end{tabular}

    \endgroup
\end{table}

\section{Task Randomization for Policy Training and Evaluation}
\label{app: Task Randomization}

%为了收集更加多样化的数据来提高模型的泛化能力和对不同环境的适应性，我们设计了分层、可配置的随机化机制，组织数据生成过程中的资产组合、空间布局和观测条件变化。在资产层面，从经过筛选的候选池中采样桌面干扰物的类别、实例及数量，构建不同资产组合；在布局层面，在固定资产组合下采样任务物体与干扰物的位置和朝向，并通过物体间距及任务保留区域约束控制空间占用；在观测层面，复用物体布局，同时对机器人基座初始位姿、环境光照、背景纹理和相机位姿施加配置允许的扰动。各随机化维度具有独立的启用选项、候选池或参数范围，并依据任务要求设置扰动幅度。通过“big group—little group—episode”的组织方式，系统分别控制资产组合覆盖、布局覆盖和布局内样本数量，并记录随机种子与抽样结果，以支持生成过程的追踪和复现。图x是一个我们开启了桌面干扰随机化和光照纹理随机化的一个4.3.3中的的任务数据采集过程中的可视化展示。

%值得注意的是，桌面干扰物随机化包括种类，数量，布局三个要素，可以进行非常多的组合。为了进行充分的泛化和尽可能的多样性，我们没有人工设计random的随机布局，而是通过设置候选随机采样的方式来
%需要详细讲一下我们的开启随机化后如何还能保证成功率的，也就是如何不断组合和进行多样化尝试的

To broaden training-data coverage and improve adaptability to environmental changes, we design a hierarchical, configurable randomization and data-generation mechanism with three levels: asset composition, spatial layout, and observation and initial conditions. The asset level samples distractor instances and counts from a filtered pool. The layout level samples the positions and orientations of designated task objects and distractors, subject to object-clearance, support-region, and reserved manipulation-space constraints. The observation and initial-condition level perturbs lighting, background textures, camera poses, and the initial robot-base pose. Each dimension's activation, sampling range, and perturbation magnitude are configured according to task requirements.

\begin{figure}[t]
    \centering
    \includegraphics[width=\linewidth]{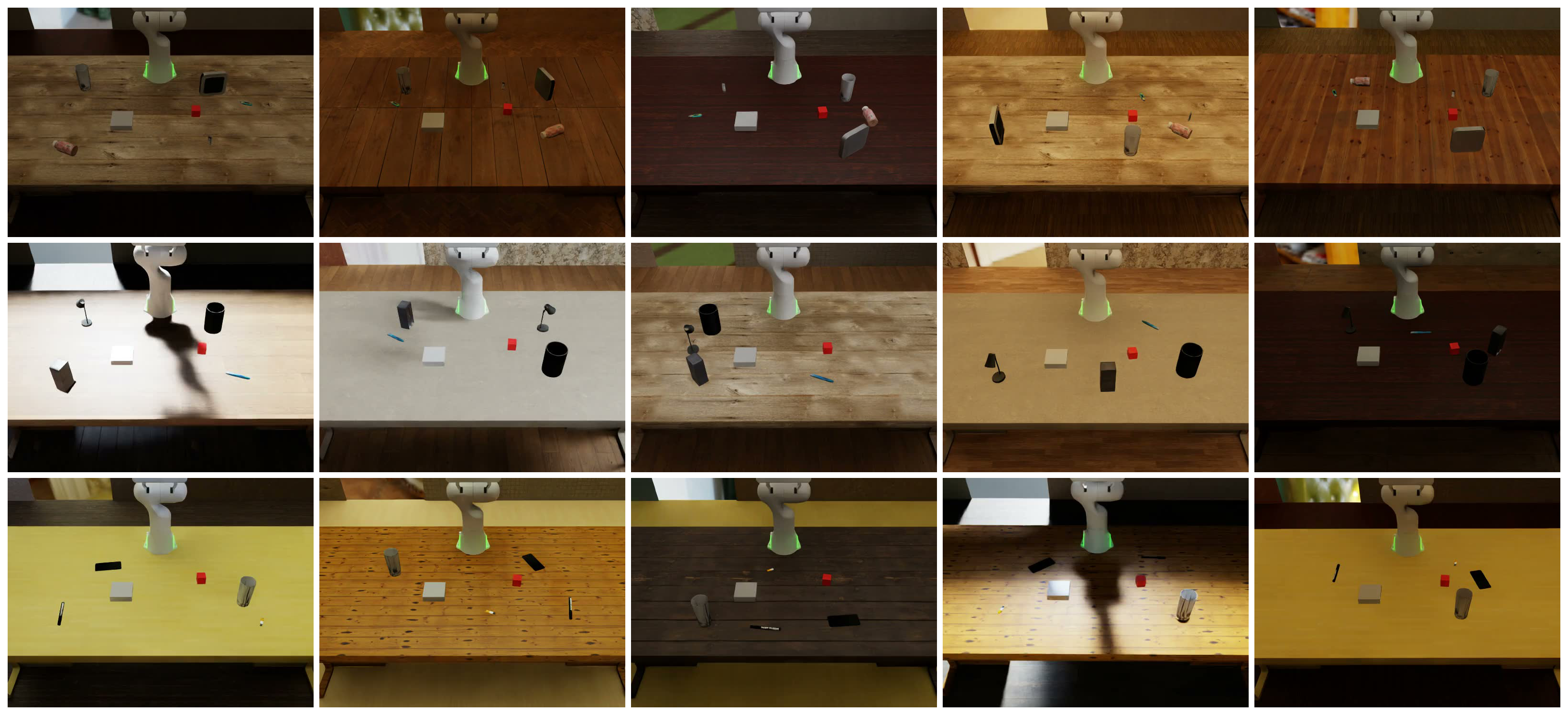}
    \caption{Data of Block stacking with randomization}
    \label{fig:random}
\end{figure}

We organize these variations through a big-group--small-group--episode hierarchy. Each big group fixes an asset composition and contains small groups with different initial object layouts. Episodes within a small group share asset identities and object layouts while resampling enabled observation conditions and robot-base perturbations. This structure separately controls asset-composition coverage, layout coverage, and episodes per layout. Combinations of distractor categories, counts, and spatial arrangements are sampled automatically, without manually enumerating layouts. \Cref{fig:random} illustrates collection examples for the task in \Cref{sec:4.3} with distractor, lighting, and texture randomization enabled.

Since geometric plausibility does not guarantee task completion, we incorporate execution feedback into grouped collection. The system replans and executes each randomized scene, counting only trajectories that pass task verification and are successfully saved toward the success quota. Once a small group reaches its quota, a new layout is sampled under the same asset composition until the big group contains enough complete small groups. Failures trigger progressively broader resampling: an individual failure preserves the object layout while episode conditions are resampled and execution is replanned; consecutive episode failures reaching a threshold trigger layout replacement; and consecutive small-group failures or exhaustion of the candidate big group's budget trigger asset-composition replacement. Rejected candidates do not count toward the target number of completed groups, and the system continues sampling new combinations to meet the collection target.

%我们收集设计的10个任务收集数据的情况如表datainfo所示，每个任务的详细描述和构造时使用的prompt如task prompt所示，两个模型的训练的超参数如表eval所示，实时推理过程展示如图teneval所示。

\section{Robot Evaluation Experiment Details}
\label{app:eval}

\begin{figure}[p]
    \centering
    \includegraphics[width=0.90\linewidth]{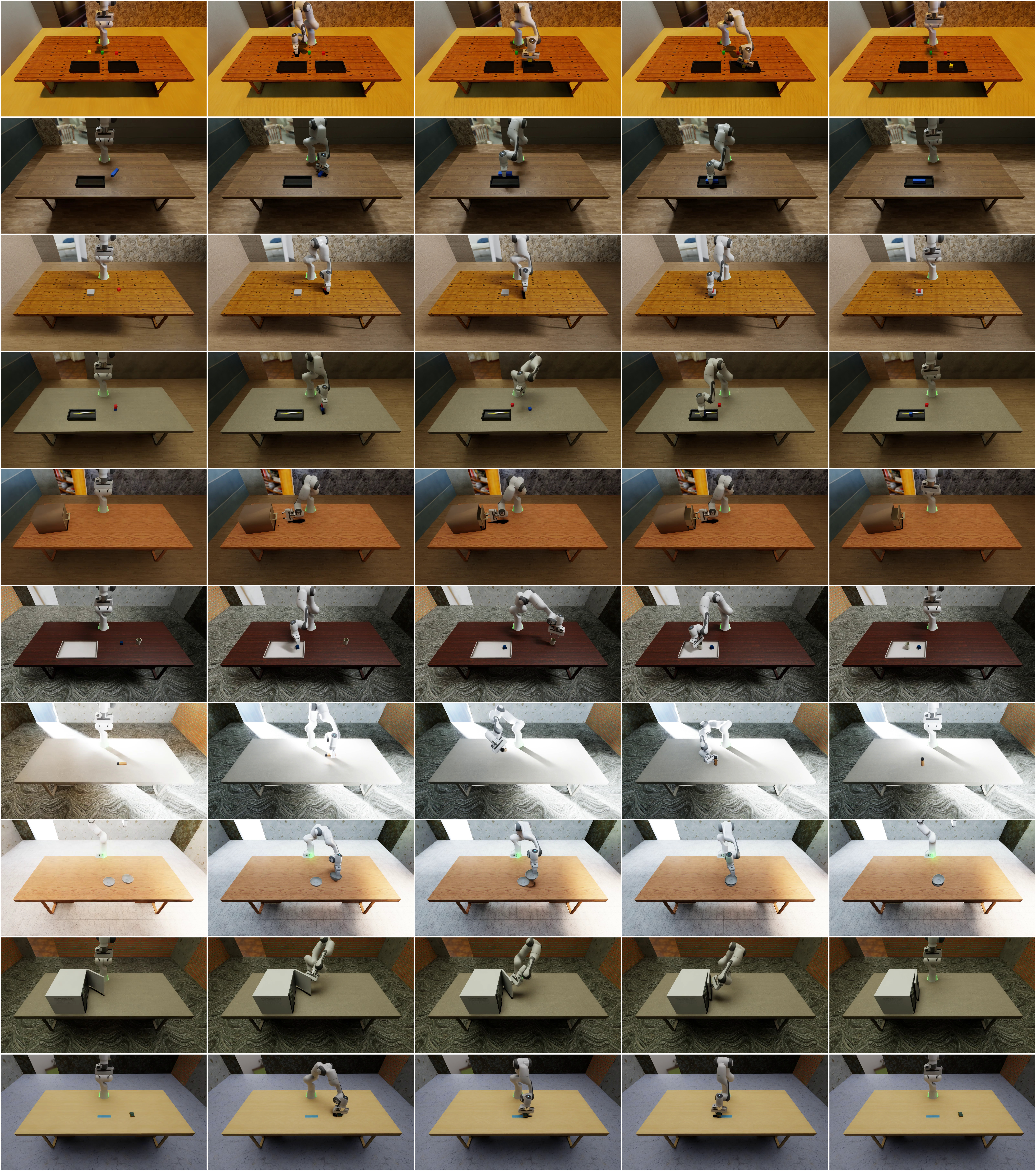}
    \caption{\textbf{Representative rollouts of the ten evaluation tasks.}
    Each row corresponds to one task, while the columns show representative
    execution stages in temporal order from left to right. For clearer
    visualization of task execution, tabletop distractors are disabled
    in these illustrative rollouts.}
    \label{fig:teneval}
\end{figure}

The task-generation prompts for the ten evaluation tasks are provided in
\Cref{tab:policy_task_prompts}, preceded by their shared requirements.
The data collection statistics are summarized in
\Cref{tab:datainfo}.
The training hyperparameters of the two evaluated models are reported in
\Cref{tab:eval_para}.
\Cref{fig:teneval} shows representative execution stages for all ten evaluation tasks, with the five study-room tasks in the upper rows and the five kitchen tasks in the lower rows. The columns follow temporal order from the initial state through intermediate interactions to task completion.

\noindent\textbf{Shared Task-generation Requirements.}
The following requirements accompany each task prompt in \Cref{tab:policy_task_prompts}.

Plan an independent task using the final assets and object IDs of the existing scene. Export an executable InterData/SimBox YAML configuration, reset settings, randomization ranges, a short policy instruction, and success conditions. Use the fixed Franka FR3 single arm in the scene and keep its mounting position unchanged. The \texttt{left} field in the project YAML denotes this arm's control channel; do not assign actions to a nonexistent right arm. Interpret directions in the reference frame specified by each task. In particular, image-relative left, middle, and right refer to the main camera view and must not be replaced by object names or world coordinates.

Reuse existing skills such as \texttt{pick}, \texttt{place}, \texttt{open}, \texttt{close}, and \texttt{move}, with parameters grounded in the registered interfaces and asset annotations rather than copied absolute coordinates. Object reorientation may be expressed through grasp and placement pose constraints. Select rotational or prismatic interaction procedures according to the asset, rather than lifting an entire cabinet or microwave with an ordinary \texttt{pick}. Check the gripper opening, approach direction, grasp annotations, lifting clearance, and support surface at the final asset scale. Avoid other objects during transport, and release and retreat before checking placement stability. Include home and observation steps only when needed, without unnecessary transfers or prolonged waits.

Reset each task to a valid initial state that does not already satisfy its goal. Configure task-appropriate randomization of object positions, orientations, and joint states, checking reachability, collision, and support relations. Keep the instruction's target, reference frame, and action order consistent with each episode's layout and trajectory. Execute a complete episode using the existing scripts and acceptance procedure, and verify task success, actions, images, states, and metadata. Preserve physical collision, support, and release behavior; do not complete tasks by teleporting objects during execution, fixing manipulated objects in place, or globally disabling collision. Only accepted tasks proceed to batch collection. Images, successful YAML parsing, or lifting an object alone do not establish task completion.

\begingroup
\footnotesize
\setlength{\tabcolsep}{4pt}
\renewcommand{\arraystretch}{1.12}
\begin{longtable}{@{}p{0.19\linewidth}p{\dimexpr0.81\linewidth-2\tabcolsep\relax}@{}}
\caption{\textbf{Task-generation Prompts for Policy Training and Evaluation.} We present the revised task-generation prompts for the five study-room tasks (A1--A5) and five kitchen tasks (B1--B5). Each entry specifies the initial state, action sequence, success conditions, and policy instruction.}
\label{tab:policy_task_prompts}\\
\toprule
\textbf{Task} & \textbf{Task-generation prompt} \\
\midrule
\endfirsthead
\toprule
\textbf{Task} & \textbf{Task-generation prompt} \\
\midrule
\endhead
\endfoot
\bottomrule
\endlastfoot

A1: Spatially Guided Placement &
\textbf{Initial state.} Use the existing study-room scene. Reset three differently colored small blocks to separate tabletop positions with clear left, middle, and right relations in the main camera view. The target tray is on the right side of the image, and the position instruction identifies the block to manipulate.\par
\textbf{Actions.} Pick the designated block, approach from a suitable direction, close the gripper securely, and lift high enough to clear nearby objects. Place it inside the target tray, release, retreat, and return the arm to its initial pose. Do not displace either of the other blocks.\par
\textbf{Success.} The block at the instructed image-relative position is fully inside the tray and stable after release, while the other blocks remain in place. Generate left, middle, and right target instructions and align each with its trajectory; color must not be the sole selection cue.\par
\textbf{Policy instruction.} ``Put the leftmost block in the image into the tray on the right, keeping the other two blocks in place.'' Generate the corresponding instructions for the middle and rightmost blocks. \\
\midrule

A2: Cuboid Placement &
\textbf{Initial state.} Use the existing study-room scene with the robot mounting position unchanged. A long box lies flat on the table, and an empty tray is within reach nearby. Their long axes initially form a clear angle.\par
\textbf{Actions.} Pick the box, adjust its horizontal orientation, and place it at the center of the tray. Choose a grasp compatible with the gripper opening. After lifting the box, align it while keeping it flat, and preserve the target orientation through placement. Leave clearance for the gripper and end effector during approach and release to avoid the table and tray.\par
\textbf{Success.} The box is fully inside the tray, with its long axis parallel to the tray's long axis and orientation error no greater than $15^\circ$. It remains stable after release and retreat, and the arm returns to its initial pose. Do not add slot insertion, flipping, or precision assembly.\par
\textbf{Policy instruction.} ``Place the long box at the center of the tray with its long side parallel to the tray's long side.'' \\
\midrule

A3: Block Stacking &
\textbf{Initial state.} Use the existing study-room scene. A wide base block and a small block lie separately on the table. The base block's top surface is clearly larger than the small block, with sufficient grasping and placement clearance.\par
\textbf{Actions.} Pick only the small block and place it at the center of the wide base block. Keep the base block in place. Lift high enough to clear the base block, approach its support surface from above, and release without striking the base block or pushing the upper block out of alignment. Build only a two-block stack.\par
\textbf{Success.} The small block is physically supported by the base block, does not overhang its edges, and remains stable for at least 1 second after the gripper retreats. The arm returns to its initial pose. Height alone is insufficient, and the task must not pass while the gripper still supports the block.\par
\textbf{Policy instruction.} ``Stack the small block at the center of the wide base block.'' \\
\midrule

A4: Unstack and Place &
\textbf{Initial state.} Use the existing study-room scene. Reset two blocks into a stack so that the upper block obstructs the main grasping space of the lower block. Place an empty tray nearby and reserve a temporary placement area beside the stack.\par
\textbf{Actions.} Pick the upper block, place it in the temporary area, pick the lower block, and place it in the tray. Confirm that the upper block has been released and settled before grasping the lower block. Do not topple the stack or grasp both blocks together. The temporary placement must not obstruct the approach to the lower block. Release, retreat, and return to the initial arm pose.\par
\textbf{Success.} The upper block remains in the temporary area, the lower block is fully inside the tray, and both are stable. Preserve evidence that the upper block was removed before the lower block was grasped; final positions alone do not verify the required order.\par
\textbf{Policy instruction.} ``Move the upper block aside first, then put the lower block into the tray.'' \\
\midrule

A5: Drawer Opening &
\textbf{Initial state.} Use the existing study-room scene. Fix an empty tabletop drawer module within reach, with its handle facing the robot. Reset the drawer to a closed or nearly closed state. Check its actual sliding direction, joint zero, limits, and handle keypoints, and leave clearance for opening and retreat.\par
\textbf{Actions.} Use an \texttt{open} skill appropriate for a prismatic drawer. Approach and grasp the handle securely, pull the drawer approximately 7\,cm along its actual sliding direction, release, retreat safely, and return to the initial arm pose. Keep the target opening within the actual joint limits. Do not lift the entire module with an ordinary \texttt{pick} or add manipulation inside the drawer.\par
\textbf{Success.} The joint reaches the target opening and remains open after release and retreat. The cabinet remains in place, without obvious interpenetration or sustained oscillation. Opening alone is required; subsequent closing is not part of the task.\par
\textbf{Policy instruction.} ``Pull open the tabletop drawer.'' \\
\midrule

B1: Sequential Object Placement &
\textbf{Initial state.} Use the existing kitchen scene. A small cup and a small block begin outside the tray. Reserve two non-overlapping placement regions inside the tray, with left and right defined relative to the robot's initial heading.\par
\textbf{Actions.} Pick the block and place it on the left, then pick the cup and place it on the right. Use stable grasps and avoid tilted insertion that may cause slipping. Keep the cup upright and avoid disturbing the placed block during the second transfer. Keep the tray in place, then release, retreat, and return to the initial arm pose.\par
\textbf{Success.} Both objects are physically supported by the tray, with their horizontal projections inside its valid interior, and satisfy the specified left--right relation without crowding each other. The cup remains upright and the gripper has released. The cup may extend above the shallow tray rim. For a reversed left--right variant, update the instruction, target regions, and action configuration together.\par
\textbf{Policy instruction.} ``Put the block on the left side of the tray and the small cup on the right.'' \\
\midrule

B2: Bottle Righting &
\textbf{Initial state.} Use the existing kitchen scene. A seasoning bottle lies horizontally on the table with clearance for grasping, reorientation, and placement. Reset it to a non-upright state. Do not fix the bottle in place or directly change its pose during execution.\par
\textbf{Actions.} Pick the bottle, lift it clear of the table, orient it with the cap upward, and place it upright on the table. Select a stable grasp according to the bottle dimensions and set the placement height from the upright bottle base and tabletop. Open the gripper fully and retreat without knocking the bottle over, then return to the initial arm pose.\par
\textbf{Success.} The cap faces upward, the bottle base is physically supported by the table, and the bottle remains upright and stable after release and retreat. The task is only to right the bottle, without moving it to a designated storage region or adding a separate return step. Reasonable position changes needed for righting are allowed.\par
\textbf{Policy instruction.} ``Stand the fallen seasoning bottle upright.'' \\
\midrule

B3: Bowl Nesting &
\textbf{Initial state.} Use the existing kitchen scene. Two identical empty bowls sit separately on the table, both facing upward. The left bowl is the upper bowl to be manipulated, and the right bowl is the lower bowl that remains in place. Check rim-grasp clearance, actual collision geometry, and whether the bowls can be stacked.\par
\textbf{Actions.} Pick the left bowl, move it above the right bowl, and place it into a natural stack. Use a grasp on one side of the rim that fits the gripper opening, rather than spanning the entire bowl opening. Keep the opening facing upward during transport, lower the bowl into place, release, retreat, and return to the initial arm pose. Stack only two bowls. Do not lift the lower bowl or modify or disable the original asset collision geometry to force nesting.\par
\textbf{Success.} The upper bowl is physically supported by the lower bowl, which remains supported by the table and in its original position. Both openings face upward and the stack remains stable after release and retreat. Verify actual stacking, rather than only nearby bowl centers or increased upper-bowl height.\par
\textbf{Policy instruction.} ``Stack the empty bowl on the left onto the empty bowl on the right.'' \\
\midrule

B4: Microwave Door Closing &
\textbf{Initial state.} Use the existing kitchen scene. Fix an empty microwave within reach, with its front facing the robot. Reset its side-hinged door to approximately $45$--$60^\circ$ open. Check the actual hinge axis, closed zero position, joint limits, and door contact point.\par
\textbf{Actions.} Adapt the \texttt{close} skill to the current asset using the skill organization of the InterData microwave-closing task. Approach the contact point, close the door along its actual hinge arc, and release or retreat. Do not pick up the appliance, require a handle absent from the model, press buttons, or start heating. Check clearance between the wrist, door, and surrounding objects throughout execution.\par
\textbf{Success.} The door angle is within $5^\circ$ of its actual closed zero and within the asset's valid closed interval. It remains stable for at least 1 second after retreat, while the appliance body stays in place. The initial open state must not already satisfy success.\par
\textbf{Policy instruction.} ``Close the microwave door.'' \\
\midrule

B5: Sponge Wiping &
\textbf{Initial state.} Use the existing kitchen scene. A rigid sponge lies at its storage location, and a short colored strip, approximately 15--20\,cm long, lies on a nearby clear tabletop area. The path has no raised border or other obstacles.\par
\textbf{Actions.} Pick the sponge, approach the strip's start, use \texttt{move} to wipe in one direction to its end, lift, and place the sponge back at its storage location. Grasp across the sponge's short axis, not its long axis, allowing the bottom surface to contact the table. Maintain tool orientation and continuous, smooth contact during wiping, without pressing deeply into the surface. Do not add repeated strokes or require deformable behavior. Finally, release and retreat.\par
\textbf{Success.} The sponge covers the prescribed path in actual contact with the table and reaches its end, then returns to storage and is released. Passing above the surface, reaching only the endpoint, or hiding the colored strip does not count as completion. Retain evidence of contact and path coverage; if the corresponding checks are missing, implement and validate them before collection.\par
\textbf{Policy instruction.} ``Wipe once along the colored strip with the sponge, then put it back.'' \\

\end{longtable}
\endgroup

\begin{table}[H]
    \centering
    \caption{Task statistics for trajectory collection.}
    \label{tab:datainfo}
    \small
    \setlength{\tabcolsep}{5pt}
    \renewcommand{\arraystretch}{1.05}

    \begin{tabular}{lccc}
        \toprule
        \textbf{Task} &
        \textbf{Atomic Actions} &
        \textbf{Avg. Duration (s)} &
        \textbf{Success Rate (\%)} \\
        \midrule
        Spatially Guided Placement  & 3 & 13.40 & 88.55 \\
        Cuboid Placement            & 3 & 14.80 & 97.32 \\
        Block Stacking              & 4 & 13.37 & 97.09 \\
        Unstack and Place           & 6 & 25.99 & 85.25 \\
        Drawer Opening              & 3 & 19.60 & 79.38 \\
        Sequential Object Placement & 6 & 31.98 & 80.32 \\
        Bottle Righting             & 4 & 22.74 & 85.61 \\
        Bowl Nesting                & 4 & 18.75 & 70.56 \\
        Microwave Door Closing      & 6 & 16.64 & 90.43 \\
        Sponge Wiping               & 5 & 49.81 & 75.54 \\
        \bottomrule
    \end{tabular}
\end{table}

\begin{table}[H]
    \centering
    \caption{Training hyperparameters. Learning-rate schedules use 10k warmup steps for $\pi_{0.5}$ and 5k for FastWAM. FastWAM retains a 100k-step cosine schedule.}
    \label{tab:eval_para}

    \begin{tabular}{lcc}
        \toprule
        \textbf{Hyperparameters} & $\pi_{0.5}$ & \textbf{FastWAM} \\
        \midrule
        Batch Size (Total)      & 256                & 64 \\
        Learning Rate           & $5 \times 10^{-5}$ & $1 \times 10^{-4}$ \\
        Learning Rate Schedule  & Constant           & Cosine Decay \\
        Training Steps (Max.)   & 20k                & 70k \\
        \bottomrule
    \end{tabular}
\end{table}

% \subsection{Physical Asset Generation.}
% \label{app:physical_asset_generation}
% Physical asset generation extends the common asset pipeline through three stages, establishing simulation readiness, physical validity, and task validity. The construction stage produces simulator-ready assets for different physical modalities. Articulated assets use LLM to iteratively write and revise construction programs from validation feedback ~\citep{zhou2026articraft}. Deformable assets use part-aware geometry with regional material properties~\citep{chen2026diaggen}, while fluid assets couple geometry with solver-specific parameters. Physics validation then tests whether assets exhibit the intended behavior in simulation, retaining revisions only when the failed metric improves. Finally, task validation checks robot-interaction feasibility using CuRobo~\citep{sundaralingam2023curobo} for collision-free motion planning. Failed assets are repaired on cloned instances, with only validated changes committed. Assets passing all stages are released to the scene generator.

%数据生产的部分，数据量的部分，任务展示的部分，训练超参数的展示

\end{document}